\documentclass{article}
\usepackage{authblk}
\newif\ifshowtmp \showtmptrue
\showtmpfalse % comment here and show the struct of paper

\newif\ifnofigure \nofiguretrue
\nofigurefalse

\PassOptionsToPackage{numbers}{natbib}
\ifnofigure
	\usepackage{comment}
	\usepackage[final]{neurips_2019}
	\excludecomment{figure}
	
	\excludecomment{table}
	
\else
	\usepackage[final]{neurips_2019}
\fi

\usepackage[utf8]{inputenc} % allow utf-8 input
\usepackage[T1]{fontenc}    % use 8-bit T1 fonts
\usepackage{url}            % simple URL typesetting
\usepackage{booktabs}       % professional-quality tables
\usepackage{amsfonts}       % blackboard math symbols
\usepackage{nicefrac}       % compact symbols for 1/2, etc.
\usepackage{microtype}      % microtypography

\label{User-Defined-Begin}
\usepackage{custom_common}
\label{Terminology}
\def\pmethod/{TRGPPO}
\newcommand{\ProposeMethod}[1][ ]{Trust Region-Guided PPO#1}

\def\Feasiblechange/{Feasible variation range}
\def\feasiblechange/{feasible variation range}
\def\explorationvariation/{exploration variation}
\newcommand{\clippingmechanism}[1][]{clipping mechanism#1}
\newcommand{\clippingratio}[1][ ]{clipping range#1}

\def\old{_{\rm old}}
\def\newp{_{\rm new}}
\def\newb{_{t+1}}
\def\oldb{_{t}}

\newcommand{\adv}[1][t]{A_{#1}}

\newcommand{\scomma}[1][]{s_{#1},}
\newcommand{\seps}[1][]{|s_{#1}}

\newcommand{\advbandit}{c }

\newcommand{\opta}{_{\rm opt}}
\newcommand{\subopta}{_{\rm subopt}}
\newcommand{\samplea}{ \hat{a}\oldb }

\newcommand{\PPO}{{\rm PPO}}

\newcommand{\setting}[2]{${\cal{C}}_{{\rm #1}}=(\{#2,s_t,a_t,\adv\}_{t=1}^T,\pi\old)$}

\newcommand{\settingshort}[1]{${\cal{C}}_{{\rm #1}}$}

\newcommand{\piopt}[1]{ \pi\newp^{\rm #1} }
\newcommand{\Piopt}[1]{ \Pi\newp^{\rm #1} }
\def\numrandomseed/{10}

\newcommand{\ratio}[1]{ \frac{\pi(a_{#1}|s_{#1})}{ \pi\old(a_{#1}|s_{#1}) } }
\newcommand{\ratiobandit}[1]{ \frac{\pi(a_{#1})}{ \pi\old(a_{#1}) } }
\newcommand{\explore}[2]{\Delta^{ \rm #1 }_{\pi_0,#2}}

\newcommand{\Lfeasiblechange}[2]{{\cal L}^{#2}_{\pi \old}(s,a_{#1})}
\newcommand{\Ufeasiblechange}[2]{{\cal U}^{#2}_{\pi \old}(s,a_{#1})}

\label{User-Defined-End}
\title{Trust Region-Guided Proximal Policy Optimization}

\author[ ]{Yuhui Wang}
\author[ ]{Hao He}
\author[ ]{Xiaoyang Tan}%\textdagger
\author[ ]{Yaozhong Gan}%\textdagger

\affil[ ]{{\tore{College of Computer Science and Technology,Nanjing University of Aeronautics and Astronautics}}}
\affil[ ]{MIIT Key Laboratory of Pattern Analysis and Machine Intelligence}
\affil[ ]{Collaborative Innovation Center of Novel Software Technology and Industrialization}
\affil[ ]{\textit {\{y.wang, hugo, x.tan, yzgancn\}@nuaa.edu.cn}}

\begin{document}

\maketitle

\begin{abstract}
Proximal policy optimization (PPO) is one of the most popular deep reinforcement learning (RL) methods, achieving state-of-the-art performance across a wide range of challenging tasks. However, as a model-free RL method, the success of PPO relies heavily on the effectiveness of its exploratory policy search. 
In this paper, we give an in-depth analysis on the exploration behavior of PPO, and show that PPO is prone to suffer from the risk of lack of exploration especially under the case of bad initialization, which may lead to the failure of training or being trapped in bad local optima. 
To address these issues, we proposed a novel policy optimization method, named \ProposeMethod[] (\pmethod/), 
which adaptively adjusts the clipping range within the trust region. We formally show that this method not only 
improves the exploration ability within the trust region but enjoys a better performance bound compared to
the original PPO as well. Extensive experiments verify the advantage of the proposed method.
\end{abstract}

\section{Introduction}
\label{Introduction}

Deep model-free reinforcement learning has achieved great successes in recent years, notably in video games \citep{mnih2015human}, board games \citep{silver2017mastering}, robotics \citep{levine2016end}, and challenging control tasks \citep{schulman2016high, duan2016benchmarking}.
Among others, policy gradient (PG) methods are commonly used model-free policy search algorithms \citep{peters2008reinforcement}. However, the first-order optimizer is not very accurate for curved areas. One can get overconfidence and make bad moves that ruin the progress of the training. Trust region policy optimization (TRPO) \cite{schulman2015trust} and proximal policy optimization (PPO) \cite{schulman2017proximal} are two representative methods to address this issue. To ensure stable learning, both methods impose a constraint on the difference between the new policy and the old one, but with different policy metrics. 

In particular, TRPO uses a divergence between the policy distributions (total variation divergence or KL divergence), whereas PPO uses a probability ratio between the two policies\footnote{{There is also a variant of PPO which uses KL divergence penalty. In this paper we refer to the one clipping probability ratio as PPO by default, which performs better in practice.}}.
The divergence metric is proven to be theoretically-justified as optimizing the policy within the divergence constraint (named trust region) leads to guaranteed monotonic performance improvement. Nevertheless, the complicated second-order optimization involved in TRPO makes it computationally inefficient and difficult to scale up for large scale problems. PPO significantly reduces the complexity by adopting a clipping mechanism which allows it to use a first-order optimization. PPO is proven to be very effective in dealing with a wide range of challenging tasks while being simple to implement and tune.

However, how the underlying metric adopted for policy constraints influence the behavior of the algorithm is not well understood. It is normal to expect that the different metrics will yield RL algorithms with different exploration behaviors. In this paper, we give an in-depth analysis on the exploration behavior of PPO, and show that the ratio-based metric of PPO tends to continuously weaken the likelihood of choosing an action in the future if that action is not preferred by the current policy. As a result, PPO is prone to suffer from the risk of lack of exploration especially under the case of bad initialization, which may lead to the failure of training or being trapped in bad local optima. 

To address these issues, we propose an enhanced PPO method, named \ProposeMethod[] (\pmethod/), which is theoretically justified by the improved exploration ability and better performance bound compared to the original PPO. In particular, \pmethod/ constructs a connection between the ratio-based metric and trust region-based one, such that the resulted ratio clipping mechanism allows the constraints imposed on the less preferred actions to be relaxed. This effectively encourages the policy to explore more on the potential valuable actions, no matter whether they were preferred by the previous policies or not. Meanwhile, the ranges of the new ratio-based constraints are kept within the trust region; thus it would not harm the stability of learning. Extensive results on several benchmark tasks show that the proposed method significantly improves both the policy performance and the sample efficiency. 
Source code is available at \url{https://github.com/wangyuhuix/TRGPPO}.

\section{Related Work}

Many researchers have tried to improve proximal policy learning from different perspectives. 
%\citeauthor{Hmlinen2018PPOCMAPP} \ttt{claims} that actions with a negative advantage value may cause instability to policy update of PPO, but if only positive advantage value is used in policy gradient estimation, the exploration variance can shrink prematurely, and they try to improve exploration behavior with evolution strategies.
%\citeauthor{Hmlinen2018PPOCMAPP} proposed a method to improve exploration behavior with evolution strategies \cite{Hmlinen2018PPOCMAPP}.
\citeauthor{Chen2018AnAC} also presented a so-called ``adaptive \clippingmechanism[]" for PPO \cite{Chen2018AnAC}.
Their method adaptively adjusts the scale of policy gradient according to the significance of state-action. They did \emph{not} make any alteration on the \clippingmechanism[] of PPO, while our method adopts a newly adaptive \clippingmechanism[]. 
\citeauthor{fakoor2019p3o} used proximal learning with penalty on KL divergence to utilize the off-policy data, which could effectively reduce the sample complexity \cite{fakoor2019p3o}.
%Second, their method make the adaptation according to
% the significance of state-action, while our method is adapt to probability of the action.
%\citeauthor{ilyas2018deep} performed a fine-grained examination to evaluate the performance of PPO empirically. They found that the optimization tricks used in PPO are important to its success \cite{ilyas2018deep}.
In our previous work, we also introduced trust region-based clipping to improve boundness on policy of PPO \cite{wang2019truly}.  While in this work, we use the trust region-based criterion to guide the clipping range adjustment, which requires additional computation but is more flexible and interpretable.

% to be exact, the advantage value on that state-action,
%Recently, \citeauthor{ilyas2018deep} performed a fine-grained examination to evaluate TRPO and PPO performance empirically. We also analyse these two methods in experiment and in theoretical, but going deeper.
%\citep{Hmlinen2018PPOCMAPP} shows that PPO's exploration variance can shrink prematurely and they use a CMA-ES algorithm to improve exploration behavior. 
%\todo{PPO-CMA}
%Third, they lack theoretical analytical comparison with PPO, while we provide better performance guarantee compared to PPO.

Several methods have been proposed to improve exploration in recent research. \citeauthor{osband2016deep} tried to conduct consistent exploration using posterior sampling method \cite{osband2016deep}. 
%To improve exploration with internally consistent strategy, 
\citeauthor{fortunato2018noisy} presented a method named NoisyNet to improve exploration by generating perturbations of the network weights \cite{fortunato2018noisy}. % presented a method named NoisyNet to 
Another popular algorithm is the soft actor-critic method (SAC) \citep{haarnoja2018soft}, which maximizes expected reward and entropy simultaneously.

\section{Preliminaries}
A Markov Decision Processes (MDP) is described by the tuple $(\mathcal{S},\mathcal{A},{\cal T},c,\rho_1,\gamma)$. $\mathcal{S}$ and $\mathcal{A}$ are the state space and action space; ${\cal T}: \mathcal{S} \times \mathcal{A} \times \mathcal{S} \rightarrow {\mathbb{R}}$ is the transition probability distribution; $c:\mathcal{S} \times \mathcal{A} \rightarrow \mathbb{R} $ is the reward function; $\rho_1$ is the distribution of the initial state $s_1$, and $\gamma \in (0,1)$ is the discount factor. The return is the accumulated discounted reward from timestep $t$ onwards, $R_t^\gamma=\sum_{k=0}^\infty \gamma^k c(s_{t+k},a_{t+k})$. 
The performance of a policy $\pi$ is defined as
$
%\begin{equation}
	\eta (\pi ) = {\mathbb{E}_{s\sim{\rho ^\pi },a\sim\pi }}\left[ {c(s,a)} \right]%(\cdot|s)
%\end{equation}
$
where ${\rho ^{{\pi  }}}(s) = (1-\gamma)\mathop \sum_{t = 1}^{\infty} {\gamma ^{t - 1}}{\rho_t^{{\pi }}}(s)$, $\rho_t^{\pi}$ is the density function of state at time $t$.
Policy gradients methods \citep{NIPS1999_1713} update the policy by the following surrogate performance objective,
%\begin{equation}\label{eq_policy_gradient}
$
%\begin{aligned}
L_{\pi\old}(\pi )={{\mathbb{E}_{s\sim{\rho ^{\pi\old}},a\sim\pi\old}}\left[ { \ratio{} {A^{\pi\old}}(s,a)} \right] + \eta(\pi\old)}, %(\cdot|s)
%\end{aligned}
$
%\end{equation}
where $ {\pi (a|s)}/{\pi\old(a|s)}$ is the \emph{probability ratio} between the new policy $\pi$ and the old policy $\pi\old$,
$A^{\pi}(s,a)=\mathbb{E}[R_t^\gamma|s_t=s,a_t=a; \pi ]-\mathbb{E}[R_t^\gamma|s_t=s; \pi ]$ is the advantage value function of policy $\pi$.
%\ttt{We will omit the sample distribution in the paper and ...}
%\tore{\cite{kakade2002approximately} and \cite{schulman2015trust}} {stated} that excessively optimizing the policy by \eqref{eq_policy_gradient} without limit may {lead to a worse policy}.
%\citep{kakade2002approximately, schulman2015trust}.
%, and let $D_{{\rm{KL}}}^{{\rm{max}}}\left( {\pi\old,\pi } \right) \triangleq {\max _{s\in\mathcal{S}}} D_{{\rm{KL}}}^{{s}}\left( {\pi\old,\pi } \right)$,
Let $D_{{\rm{KL}}}^{{s}}\left( {\pi\old,\pi } \right) \triangleq D_{{\rm{KL}}}^{}\left( {\pi\old(\cdot|s)||\pi (\cdot|s)} \right)$,
Schulman et al. \cite{schulman2015trust} derived the following performance bound:
\begin{theorem}\label{thm_lowerbound}
	Define that
		$C = \mathop {\max }\limits_{s,a} \left| {{A^{\pi\old}}\left( {s,a} \right)} \right|{4\gamma } \mathord{\left/
			 {\vphantom {{4\gamma } {{{(1 - \gamma )}^2}}}} \right.
			 \kern-\nulldelimiterspace} {{{(1 - \gamma )}^2}}
		$, 
%	\begin{flalign}
$
	{M_{\pi\old}}(\pi ) = {L_{\pi\old}}(\pi ) - C\max_{s \in {\cal S}}D_{\rm{KL}}^{\rm{s}}\left( {\pi\old,\pi } \right)
$. 
%	\label{eq_lower_bound}
%	\end{flalign}
{We have}
%	\begin{align}
$
	\eta(\pi)\geq{M_{\pi\old}}(\pi ),
	\eta(\pi\old)={M_{\pi\old}}(\pi\old ).
$
%	\label{eq_eta_lower_bound}
%	\end{align}
\end{theorem}
This theorem implies that maximizing ${M_{\pi\old}}(\pi )$ guarantee non-decreasing of the performance of the new policy $\pi$.
%if the new policy $\pi$ strays too far from the old one $\pi\old$, the performance of the new policy $\pi$ is not guaranteed.
%Maximizing $M_{\pi\old}(\pi)$ leads to a guaranteed performance improvement of policy $\pi$. 
To take larger steps in a robust way, TRPO optimizes $L_{\pi\old}(\pi)$ with the constraint $\max_{s \in {\cal S}}D_{\rm{KL}}^{\rm{s}}\left( {\pi\old,\pi } \right) \le \delta$, which is called the \emph{trust region}.
%\begin{subequations}\label{eq_TRPO}
%\begin{align}
%\mathop {\max }\limits_\pi  & L_{\pi\old}(\pi)
%\\
%\text{s.t.}& \max_{s \in {\cal S}}D_{\rm{KL}}^{\rm{s}}\left( {\pi\old,\pi } \right) \le \delta  \label{eq_trust_region}%
%\end{align}
%\end{subequations}
%{Constraint \eqref{eq_trust_region}} is called the \emph{trust region constraint}, which is a constraint on the KL divergence between the old policy and the new one.

\section{The Exploration Behavior of PPO}\label{sec_problem}

In this section will first give a brief review of PPO and then show that how PPO suffers from an exploration issue when the initial policy is sufficiently far from the optimal one. 
%To give a clear illustration, we discuss under the bandit problem \citep{agrawal1995continuum}, which is a special case of MDP with no state transitions. \todo{more complex case}

PPO imposes the policy constraint through a clipped surrogate objective function:
%restricting the probability ratio $\pi(a\seps)/\pi\old(a\seps)$ through a clipping function, 
%\todo{write as a sampled based}
\begin{equation}
	L_{\pi\old}^{\rm{CLIP}}(\pi ) \hiderel{=} \mathbb{E}
%	_{s\sim\rho^{\pi\old},a\sim\pi\old} 
	\left[ \min \left( \ratio{} A^{\pi\old}(\scomma a), \mathop{clip} \left( {\ratio{},{l_{\scomma {a}}},{u_{\scomma {a}}}} \right) A^{\pi\old}(\scomma a) \right) \right]
	\label{eq_approx_PPO}
\end{equation}
where $l_{\scomma a}\in(0,1)$ and $u_{\scomma a}\in(1,+\infty)$ are called the lower and upper \emph{\clippingratio[]} on state-action $(s,a)$.
%\ttt{Such constant setting could over-restrict variation on $\pi(a_t|s_t)$ when $\pi\old(a_t|s_t)$ is smaller, compared to , which would cause the \emph{lack of exploration} problem.}
%\ttt{In such a constant setting, update of policy $\pi$ is less restricted at $(s_t,a_t)$ if $\pi\old(a_t|s_t)$ is larger.
%This could cause $\pi(a_t|s_t)$ to be progressively larger if $\pi\old(a_t|s_t)$ is larger, which means that the policy is encouraged to be deterministic.}
%\ttt{We found that such a constant setting could make policy become less random, even the current policy is a locally optimal one. 
%Such encouragement of deterministic could make the policy lack of exploration and easily get trapped in local optima.
%}
The probability ratio $\pi(a \seps)/\pi\old( a \seps)$ will be clipped once it is out of $(l_{\scomma a}, u_{\scomma a})$. 
%As a result, the new policy $\pi$ will stop enlarging its difference with the old policy $\pi\old$.
Therefore, such clipping mechanism could be considered as a constraint on policy with ratio-based metric, i.e., $l_{\scomma a}\leq \pi(a\seps)/\pi\old(a \seps)\leq u_{\scomma a}$, which can be rewritten as, $-\pi\old(a \seps)(1-l_{\scomma a })\leq \pi(a\seps)-\pi\old(a \seps)\leq \pi\old(a \seps)(u_{\scomma a}-1)$. We call $(\Lfeasiblechange{}{l},\Ufeasiblechange{}{u}) \triangleq \left(-\pi\old(a \seps)(1-l_{\scomma a}), \pi\old(a \seps)(u_{\scomma a}-1)\right)$ the \emph{\feasiblechange/} of policy $\pi$ w.r.t. $\pi\old$ on state-action $(s,a)$ with the clipping range setting $(l,u)$, which is a measurement on the allowable change of policy $\pi$ on state-action $(s,a)$.

Note that the original PPO adopts a constant setting of \clippingratio[], i.e., $l_{\scomma a}={1-\epsilon}, u_{\scomma a}={1+\epsilon}$ for any $(\scomma a)$ \citep{schulman2017proximal}.
The corresponding \feasiblechange/ is $(\Lfeasiblechange{}{1-\epsilon},\Ufeasiblechange{}{1+\epsilon})=(-\pi\old(a \seps)\epsilon, \pi\old(a \seps)\epsilon )$. 
As can be seen, given an optimal action $a\opta$ and a sub-optimal one $a\subopta$ on state $s$, if $\pi\old(a\opta \seps)<\pi\old(a\subopta \seps)$, then $|(\Lfeasiblechange{\opta}{1-\epsilon},\Ufeasiblechange{\opta}{1+\epsilon})|<|(\Lfeasiblechange{\subopta}{1-\epsilon},\Ufeasiblechange{\subopta}{1+\epsilon})|$.
This means that the allowable change of the likelihood on optimal action, i.e., $\pi(a{\opta}|s)$, is smaller than that of $\pi(a{\subopta}|s)$.
%This means that the optimal action is less preferred than the sub-optimal one by the old policy, then the \feasiblechange/ of probability on the optimal action would be more limited than that on the sub-optimal one. 
Note that $\pi(a{\opta}|s)$ and $\pi(a{\subopta}|s)$ are in a zero-sum competition, such unequal restriction may continuously weaken the likelihood of the optimal action and make the policy trapped in local optima. We now give a formal illustration.
%We give a detailed illustration below.
%Second, 
%the \feasiblechange/ could be quite limited when $\pi\old(a \seps)$ is sufficiently small.
%This means that it would require a large number of \ttt{steps} to approach the optimal policy if the policy is initialized from a bad one.

%We will first give a example
 
%The reward function is defined as
%% $c$ 
%%\begin{equation*}
%%c_{\rm }(a)=
%%\begin{cases}
%%0.5 & 1\leq a \leq 2\\
%%1 & 2.5\leq a \leq 4\\
%%0 & \text{otherwise}
%%\end{cases}
%%\end{equation*}
%\begin{center}
%\begin{tabular}{| c | c | c | c | }
%\hline
%Action & $1 \leqslant a \leqslant 2$ & $2.5\leqslant a \leqslant 4$ & otherwise \\ 
%\hline
%Reward & 0.5 & 1 & 0   \\
%\hline
%\end{tabular}
%\end{center}

%As \Cref{alg_simple} describes, the policy is optimized through the surrogate objective function of PPO until the policy converges to the optimal one.

\begin{algorithm}[!h]
\footnotesize
\caption{Simplified Policy Iteration with PPO }
%\footnotetext{The change on $\pi(a_t)$ will also cause a change on other $a\neq a_t$ }
\begin{algorithmic}[1]\label{alg_simple}
\STATE Initialize a policy $\pi_0$, $t\leftarrow0$.
\REPEAT 
\STATE
Sample an action $\samplea \sim \pi\oldb$. 
\STATE
Get the new policy $\pi\newb$ by optimizing the empirical surrogate objective function of PPO based on $\samplea$:

\begin{equation}\label{eq_pi_new}
\begin{scriptsize}
\hat\pi\newb(a) 
= 
\begin{cases}
\pi\oldb(a )u_{a}   & a=\samplea \text{ and } \advbandit( a)>0  \\
\pi\oldb(a )l_{ a} & a=\samplea \text{ and } \advbandit( a)<0 \\
\pi\oldb(a ) - \frac{ \pi\oldb(\samplea )u_{\samplea} - \pi\oldb(\samplea)}{ |{\cal A}|-1 }  
	& a \neq \samplea \text{ and } \advbandit( \samplea)>0 \\
\pi\oldb(a ) + \frac{ \pi\oldb(\samplea) (1-l_{\samplea}) }{ |{\cal A}|-1 }  
	& a \neq \samplea \text{ and } \advbandit( \samplea)<0 \\
\pi\oldb(a ) &  \advbandit( \samplea)=0 
\end{cases}
\end{scriptsize}
\end{equation}
\STATE $\pi_{t+1} ={\mathop {Normalize}} (\hat\pi_{t+1}) $\footnotemark. $t\leftarrow t+1$.
\UNTIL{$\pi\oldb$ converge}
%\label{alg_simple_step}
%\UNTIL{aa}
%\STATE \ttt{If $\pi\newp(a_{\rm opt})=1$, then end; else $\pi\oldb \leftarrow \pi\newp$, go to \ref{alg_simple_step}.}
\end{algorithmic}
\end{algorithm}
\footnotetext{$\hat\pi\newb$ may violate the probability rules, e.g., $\sum_a \hat\pi\newb(a)>1$. Thus we need to enforce specific normalization operation to rectify it.
To simplify the analysis, we assume that $\pi\newb = \hat \pi\newb $.
% for any $\samplea$.
}
%\footnotetext{The \ttt{size} of the $c(a)$ may also result in different new policy in practice. However, with sufficient training, the new probability would reach the threshold. 
%We simplify the statement to see how the clipping range affect the algorithm behavior.
%}

%We begin by introducing the problem and the algorithm settings.
We investigate the exploration behavior of PPO under the discrete-armed bandit problem, where there are no state transitions and the action space is discrete.
%\footnote{{The exploration behavior of PPO in continuous action space case is similar to that in the discrete one.}}.
The objective function of PPO in this problem is 
$
	L_{\pi\old}^{\rm{CLIP}}(\pi ) \hiderel{=} \mathbb{E}
	\left[ \min \left( \ratiobandit{} \advbandit( a), \mathop{clip} \left( {\ratiobandit{},{l_{ {a}}},{u_{ {a}}}} \right) \advbandit( a) \right) \right]
$.
%defined as $\eta(\pi)={\mathbb E}_{a\sim\pi}\left[ c(a) \right]$, and the advantage function is defined as $\advbandit(a)=c(a)-\eta(\pi\old)$. 
\emph{
Let ${\cal A}^+\triangleq\{ a \in {\cal A} | \advbandit(a)>0 \}$, ${\cal A}^-\triangleq\{ a \in {\cal A} | \advbandit(a)<0 \}$ denote the actions which have positive and negative reward respectively, and ${\cal A}\subopta={\cal A}^+/\{a\opta\} $ denote the set of the sub-optimal actions. Let $a\opta=\argmax_a{c(a)}$ and $a\subopta \in {\cal A}\subopta$ denote the optimal \footnote{Assume that there is only one optimal action.} and a sub-optimal action.
}
Let us consider a simplified online policy iteration algorithm with PPO. As presented in \Cref{alg_simple},
the algorithm iteratively sample an action $\samplea$ based on the old policy $\pi\old$ at each step and obtains a new policy $\pi\newp$. 
%We give an example of the settings below.

%Now we investigate how the exploration behavior is affected by the setting of clipping range. 
%Intuitively, the policy should explore more on the optimal action $a\opta$. More exactly, it should allocate more probability on the optimal action\footnote{\ttt{Although the algorithm does not know which action is the optimal one, we could judge the algorithm behavior from a god-like perspective by observing its decision on the optimal action.}}. 
%Therefore, we measure the exploration behavior by judging whether probability $\pi\newp(a\opta)$ is promoted, which is formally described by $\pi\newp(a\opta)-\pi\old(a\opta)$. Note that new policy $\pi\newp$ depends on the stochastic action sampled from $\pi\old$ (see \cref{eq_pi_new}), thus the promotion measure is taken to be the expected one, i.e., $\Delta(a|\pi\old) \triangleq \mathbb{E}_{\pi\newp}\left[ \pi\newp(a) - \pi\old(a) | \pi\old \right]$, which is named \emph{exploration variation on action $a$}. Larger and positive $\Delta(a|\pi\old)$ means action $a$ will be explored more in the next step.
We measure the exploration ability by the expected distance between the learned policy $\pi_t$ and the optimal policy $\pi^*$ after $t$-step learning, i.e., $\explore{}{t} \triangleq \mathbb{E}_{\pi_t} \left[ \| \pi_t - \pi^*  \|_{\infty} | \pi_0 \right] $, where $\pi^*(a\opta)=1$, $\pi^*(a)=0$ for $a\neq a\opta$, $\pi_0$ is the initial policy, $\pi_t$ is a stochastic element in the policy space and depends on the previous sampled actions $\{a_{t'}\}_{t'=1}^{t-1}$ (see \cref{eq_pi_new}). \emph{Note that smaller $\explore{}{t}$ means better exploration ability, as it is closer to the optimal policy.} We now derive the exact form of $\explore{}{t}$.
\begin{lemma}\label{theorem_distance_opt}
$\explore{}{t} \triangleq\mathbb{E}_{\pi_t} \left[ \| \pi_t - \pi^*  \|_{\infty} | \pi_0 \right]= 1-\mathbb{E}_{\pi_t} \left[ \pi_t(a\opta) | \pi_0 \right] $.
\end{lemma}
\begin{lemma} \label{theorem_iterative_pi}
$\mathbb{E}_{\pi\newb} \left[ \pi\newb(a) | \pi_0 \right] = \mathbb{E}_{\pi\oldb} \left[ \mathbb{E}_{\pi\newb} \left[ \pi\newb(a) |\pi\oldb \right] |\pi_0\right]$.
\end{lemma}
We provide all the proofs in {Appendix} \ref{app-sec_proof}.
\Cref{theorem_distance_opt} implies that we can obtain the exploration ability $\explore{}{t}$ by computing the expected likelihood of the optimal action $a\opta$, i.e., $\mathbb{E}_{\pi_t} \left[ \pi_t(a\opta) | \pi_0 \right]$.
And \Cref{theorem_iterative_pi} shows an iterative way to compute the exploration ability.
By \cref{eq_pi_new}, 
%for PPO with general clipping range $(l_a,u_a)$, 
for action $a$ which satisfies $\advbandit(a)>0$, we have

\begin{equation}\label{eq_pinew_general}
\begin{tiny}
\begin{aligned}
 \mathbb{E}_{\pi\newb}\left[\pi\newb(a) | \pi\oldb\right]
% \\
= 
\pi\oldb(a) +
\left[
\pi\oldb^2(a)( u_a - 1 )
- \sum_{a^+ \in {\cal A}^+/\{ a \} }  \frac{\pi\oldb^2(a^+)}{ |{\cal A}|-1 } (u_{a^+}-1)
+ \sum_{a^-\in {\cal A}^- }   \frac{ \pi\oldb^2(a^-) }{ |{\cal A}|-1 } (1-l_{a^-})
\right]
%- \frac{1}{ |{\cal A}|-1 } \sum_{a\in {\cal A}/\{a\opta\}} \mathop{sign}\left( \advbandit(a)\right) \pi\oldb^2(a) 
\end{aligned}
\end{tiny}
\end{equation}
% ---------------- for PPO
%Particularly, for the original PPO with hyperparameter $\epsilon$, we have
%\begin{equation}
%\begin{aligned}
% &\mathbb{E}_{\pi\newb^\PPO}\left[\pi\newb^\PPO(a) | \pi\oldb\right]
%=  &  \pi\oldb(a)
%+
%%\underbrace{
%\left[ 
%\pi\oldb^2(a) 
%- \sum_{a^+ \in {\cal A}^+/\{a\} }  \frac{\pi\oldb^2(a^+)}{ |{\cal A}|-1 }
%+ \sum_{a^- \in {\cal A}^- }   \frac{ \pi\oldb^2(a^-) }{ |{\cal A}|-1 }
%\right] \epsilon
%%}_{\text{variation term at $t$}}
%\end{aligned}
%\end{equation}
% --------------- End for PPO
%\tore{set remove right?}
%Note that the resulted $\pi\newb$ depends on the stochastic sampled action (as \cref{eq_pi_new} implies), thus $\mathbb{E}_{\pi\newb} \left[  \pi\newb(a_{\rm opt}) \right]$ means the expected value of $\pi\newb(a_{\rm opt})$. 
This equation provides a explicit form of the case when the likelihood of action $a$  would decrease. 
That is, if the second term in RHS of \cref{eq_pinew_general} is negative, then the likelihood on action $a$ would decrease.
%Note that $\pi_{t}$ also relies on the last policy $\pi_{t-1}$, and so on. 
This means that the initialization of policy $\pi_0$ profoundly \ttt{affects} the future policy $\pi_t$. 
Now we show that if the policy $\pi_0$ initializes from a bad one, $\pi(a\opta)$ may continuously be decreased. 
Formally, for PPO, we have the following theorem:
%If the \emph{variation term at $t$} is negative, then the new probability $\pi\newb^\PPO(a)$ is expected to decrease which would make the new \emph{variation term at $t+1$} be negative.
%\begin{corollary}
%For PPO, if $\pi_\oldb( a\opta )<\pi_\oldb( a\subopta )$, then $\Delta(a\opta)<\Delta(a\subopta)$.
%\end{corollary}
\begin{theorem}
%Let $\{ \pi_t \}_{t=0}^T$ denote a policy sequence of PPO in \Cref{alg_simple}. 
Given initial policy $\pi_0$,
if 
$
%\begin{equation}\label{eq_exploration_condition}
\pi_0^2(a\opta) \cdot |{\cal A}| < \sum_{a\subopta \in {\cal A}\subopta }  {\pi_0^2(a\subopta)} - \sum_{a^- \in {\cal A}^- } { \pi_0^2(a^-) }, 
%\end{equation}
$ 
then we have 

(\romannumeral1) $\sum_{a\subopta \in {\cal A}\subopta } \pi_0^{\rm }(a\subopta) < \sum_{a\subopta \in {\cal A}\subopta } \mathbb{E}_{\pi_{1}^\PPO} \left[  \pi_{1}^\PPO(a\subopta) | \pi_0 \right] < \cdots < \sum_{a\subopta \in {\cal A}\subopta } \mathbb{E}_{\pi_{t}^\PPO} \left[  \pi_{t}^\PPO(a \subopta) | \pi_0 \right]$;

(\romannumeral2) $\pi_0(a\opta) > \mathbb{E}_{\pi_{1}^\PPO} \left[  \pi_{1}^\PPO(a\opta) | \pi_0 \right] > \cdots > \mathbb{E}_{\pi_{t}^\PPO} \left[  \pi_{t}^\PPO(a\opta) |\pi_0 \right]$;

(\romannumeral3) $\explore{}{0} < \explore{PPO}{1} < \cdots < \explore{PPO}{t}$.
\end{theorem}
Conclusion (\romannumeral1) and (\romannumeral2) implies that if the optimal action $a\opta$ is relatively less preferred than the sub-optimal action $a\subopta$ by the initial policy, then the preference of choosing the optimal action would continue decreasing while that of the sub-optimal action would continue increasing. This is because the feasible variation of probability on the optimal action $\pi(a\opta)$ is larger than that on the sub-optimal one $\pi(a\subopta)$, increasing probability on the latter one could diminish the former one.
Conclusion (\romannumeral3) implies that the policy of PPO is expected to diverge from the optimal one (in terms of the infinity metric). We give a simple example below.
%\tore{For \Cref{example_discrete_bandit}, $\explore{PPO}{1}=-0.061$.}

\begin{example}\label{example_discrete_bandit}
Consider a three-armed bandit problem, the reward function is $c(a_{\rm opt})=1,c(a_{\rm subopt})=0.5,c(a_{\rm worst})=-50$. The initial policy is $\pi_0(a\opta)=0.2,\pi_0(a\subopta)=0.6, \pi_0(a_{\rm worst})=0.2$. 
The hyperparameter of PPO is $\epsilon=0.2$. We have \ttt{$\explore{PPO}{0}=0.8$, $\explore{PPO}{1}=0.824$,\ldots, $\explore{PPO}{6}\approx0.999$}, which means the policy diverges from the optimal one.
%, which means $(l_a,u_a)=(0.8,1.2)$ for all $a$.
\end{example}

Note that the case that the optimal action $a\opta$ is relatively less preferred by the initial policy may be avoided in discrete action space, where we can use uniform distribution as initial policy. However, such a case could hardly be avoided in the high dimensional action space, where the policy is possibly initialized far from the optimal one.
We have experimented \Cref{example_discrete_bandit} and a continuous-armed bandit problem with random initialization for multiple trials; about 30\% of the trials were trapped in the local optima. See \Cref{sec_experiment_bandit} for more detail.

%Then, we measure the sample complexity of the algorithm by the number of steps taken to approach the optimal policy.
%\begin{theorem}\label{thm_sample_complexity}
%Let $T$ denote the steps for \Cref{alg_simple} to converge to the optimal policy, i.e., $\pi(a\opta)=1$ and $\pi(a)=0$ for all $a\neq a\opta$; 
%let $T_{\rm min}$ denote the least steps required to converge to the optimal policy. 
%Assume that $\pi_0(a\opta)>0$, we have
%\begin{multicols}{2}
%\begin{equation} \label{T_min}
%T_{\rm min}=\lceil \log_{u_{a\opta}} \frac{1}{\pi_0(a_{\rm opt})} \rceil, \quad
%\end{equation}\break
%\begin{equation} \label{T_prob}
%P(T \leqslant T_{\rm min})=\pi_0(a_{\rm opt})^{T_{\rm min}}u_{a\opta}^{T_{\rm min}-1}
%\end{equation}
%\end{multicols}
%\end{theorem}
%We provide all the proofs in {Appendix} \ref{app-sec_proof}.
%For PPO, if the initial probability $\pi_0(a_{\rm opt})$ is small, it would require a large number of steps for PPO to converge to the optimal policy.
%For \Cref{example_discrete_bandit}, $T_{\rm min}^{\rm PPO}=13$ and $P(T\leqslant 13)< 10^{-12}$.
%If we adjust the clipping range $u_{a}$ to be larger when $\pi\oldb(a)$ is small, then we would have smaller sample complexity $T_{\rm min}$ and larger probability $P(T\leq T_{\rm min})$.

In summary, PPO with constant clipping range could lead to an exploration issue when the policy is initialized from a bad one.
%due to that the restriction is relatively restrict when the action is not preferred by the initial policy. 
%Both the result in Eq. \eqref{T_min}, \eqref{T_prob} and \eqref{eq_Delta_general} 
However, \cref{eq_pinew_general} inspires us a method to address this issue \textminus{} enlarging the clipping range $(l_{a},u_{a})$ when the probability of the old policy $\pi\old(a)$ is small.

\section{Method}
%In this section, we will detail our method, \pmethod, and then give an analytical comparison of \pmethod/ and PPO.
 
\subsection{\ProposeMethod}

% old
%As discussed in section \ref{sec_problem}, the problems with PPO, i.e., the uneven restrictions on different states and actions, derives from the constant setting of \clippingratio[] on all states. Adjusting the \clippingratio[] for each state action may solve this problem. However, improper enlarging the \clippingratio[] lead to large divergence of policy which may result in degradation.
% v1
%Our objective is to adaptively adjust the clipping range for each state-action pair to make 
%\emph{(i) the allowable change on each state-action pair are close}
%while \emph{(ii) the ``trust region'' constraint is retained}.
% v2
%Our proposed method, \ProposeMethod (\pmethod/), adaptively adjust the clipping range for each state-action pair and could make 
%\emph{(i) the allowable change on each state-action pair are more \ttt{closer}}
%while \emph{(ii) the ``trust region'' constraint is retained}, compared to original PPO.

%We aim to improving the exploration behavior of PPO.
%Both TRPO \cite{a} and PPO \cite{b} devotes to making policy learning stable by imposing a constraint on the difference between the new policy and the old one. 
In the previous section, we have concluded that the constant clipping range of PPO could lead to an exploration issue. 
%and we have also provided a plausible resolution to address these issues \textminus{} allocating larger clipping range $(l_{s,a},u_{s,a})$ when the preference of the action $\pi\old(a|s)$ is small. 
We consider how to adaptively adjust the clipping range to improve the exploration behavior of PPO.
The new clipping range $(l^\delta_{s,a}, u^\delta_{s,a})$, where $\delta$ is a hyperparameter, is set as follows:
%However, \ttt{inconsiderately} enlarging the clipping range may lead to unstable policy learning.
%Our proposed method, \ProposeMethod (\pmethod/), adaptively adjust the clipping range, the new clipping range $(l^{\delta}_{s,a}, u^\delta_{s,a})$ is set as following:
%The \clippingratio[s] $(l_{s,a}, u_{s,a})$ in \eq \eqref{eq_approx_PPO} to be
%\begin{multicols}{2}
%\begin{small}
\begin{gather}
l_{s,a}^{\delta} = \mathop {\min }\limits_\pi   \left\{  \frac {\pi(a|s)}{\pi\old(a|s)} : {D_{\rm KL}^{s}(\pi\old,\pi ) \leq \delta }  \right\},
%\end{gather}
%\\
%\begin{gather}
u_{s,a}^{\delta} = \mathop {\max }\limits_\pi   \left\{  \frac {\pi(a|s)}{\pi\old(a|s)} : {D_{\rm KL}^{s}(\pi\old,\pi ) \leq \delta }  \right\}
\label{eq_clippingrange_within_trust}
\end{gather}
%\end{small}
%\end{multicols}
%As these equations imply, the upper \clippingratio[s] are set to be {the largest probability ratio} that the new policy could achieve within the trust region, while lower \clippingratio[s] are set to be the smallest one. 

%\todo{can it be deleted}
{To ensure the new adaptive clipping range would not be over-strict, an additional truncation operation is attached:}
%\begin{gather}
$
l_{s,a}^{\delta, \epsilon} = \min( l_{s,a}^{\delta}, 1-\epsilon ), u_{s,a}^{\delta,\epsilon}=\max(u_{s,a}^{\delta} ,1+\epsilon).
$
%\end{gather}
This setting of clipping range setting could be motivated from the following perspectives.

First, the clipping range is related to the policy metric of constraint. Both TRPO and PPO imposes a constraint on the difference between the new policy and the old one. TRPO uses the divergence metric of the distribution, i.e., $D_{\rm KL}^s (\pi\old,\pi)=\mathbb{E}_{a} \left[ \log \frac{\pi\old(a|s)}{\pi(a|s)} \right] \leq \delta$ for all $s\in{\cal S}$, which is more theoretically-justified according to \Cref{thm_lowerbound}.
Whereas PPO uses a ratio-based metric on each action, i.e., $1-\epsilon\leq \frac{\pi(a|s)}{\pi\old(a|s)}\leq1+\epsilon$ for all $a \in {\cal A}$ and $s \in {\cal S}$. 
The divergence-based metric is averaged over the action space while the ratio-based one is an element-wise one on each action point. If the policy is restricted within a region with the ratio-based metric, then it is also constrained within a region with divergence-based one, but not vice versa. 
Thus the probability ratio-based metric constraint is somewhat more strict than the divergence-based one.
Our method connects these two underlying metrics \textminus{} adopts the probability ratio-based constraint while getting closer to the divergence metric. 

\ifnofigure
\else
%\begin{wrapfigure}{R}{0.8\textwidth}
\begin{figure}[!b]
	\setlength{\abovecaptionskip}{0mm}
	\begin{tiny}
		\centerline{
		\begin{subfigure}[]{0.25\linewidth}
			\includegraphics[width=1.0\linewidth]{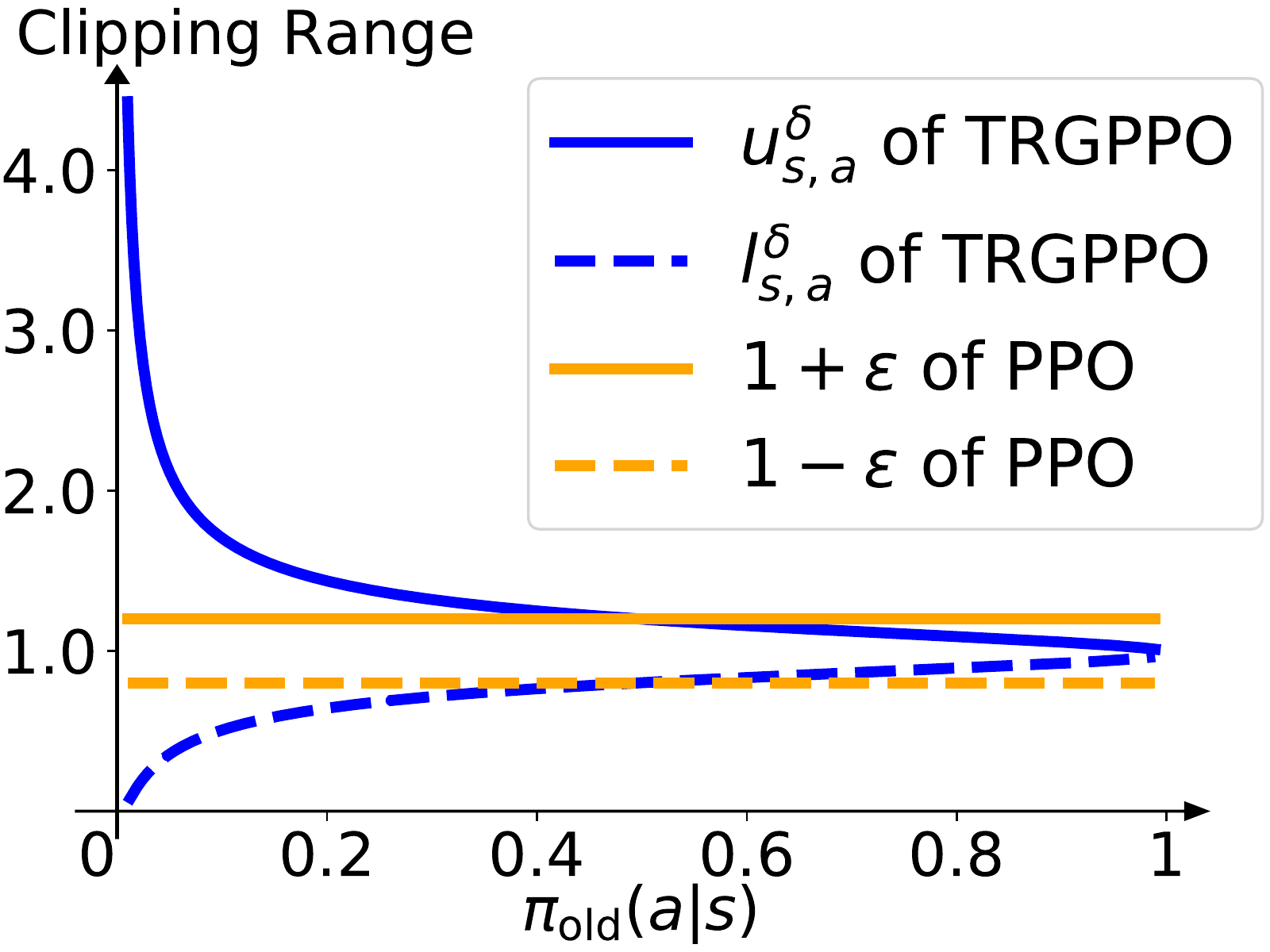} 
			\caption{}\label{fig_kl2clip_discrete_ratio}
		\end{subfigure}
		\begin{subfigure}[]{0.25\linewidth}
			\includegraphics[width=1.0\linewidth]{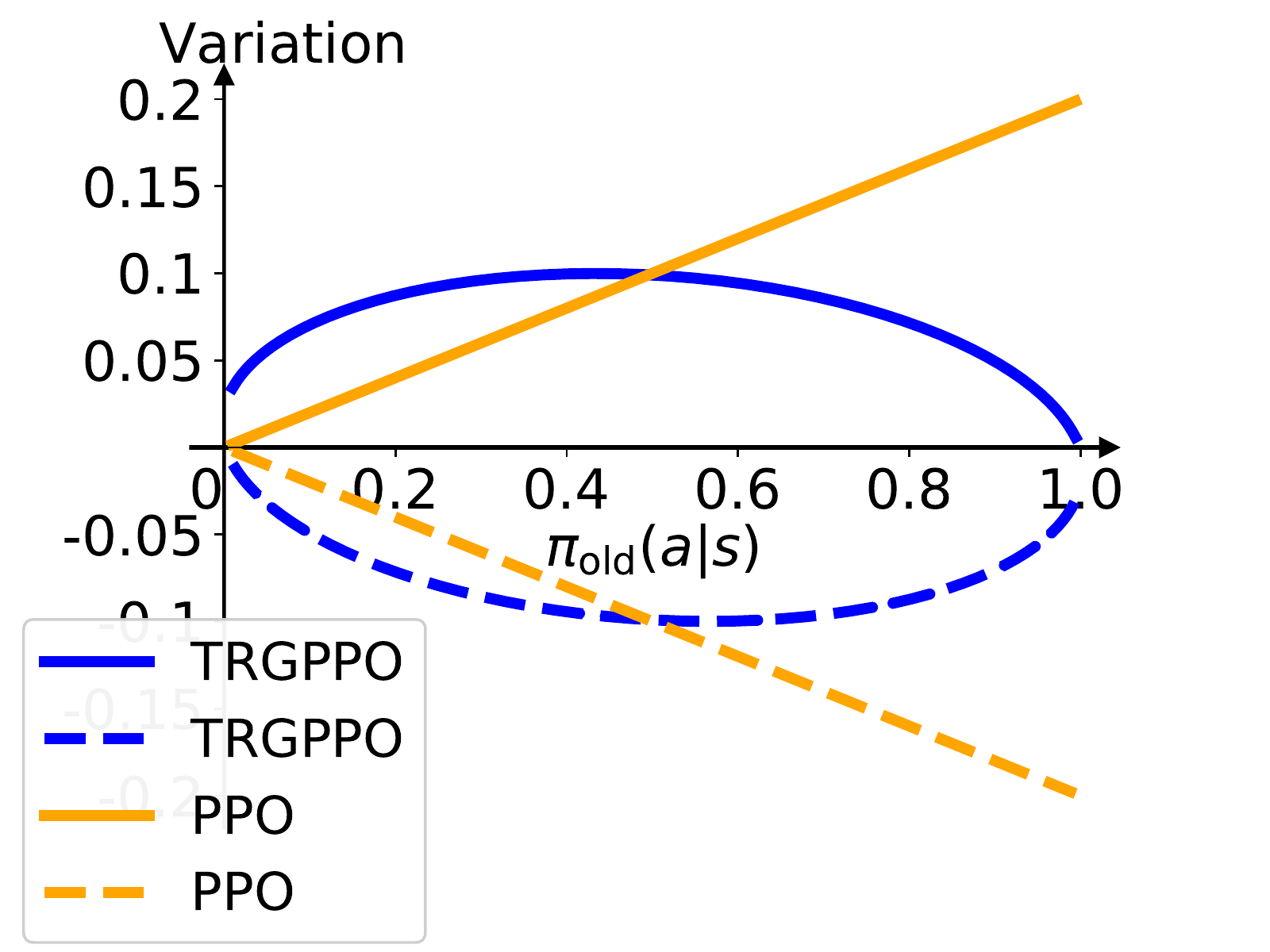} 
			\caption{}\label{fig_kl2clip_discrete_probability}
		\end{subfigure}
		\begin{subfigure}[]{0.25\linewidth}
			\includegraphics[width=1.0\linewidth]{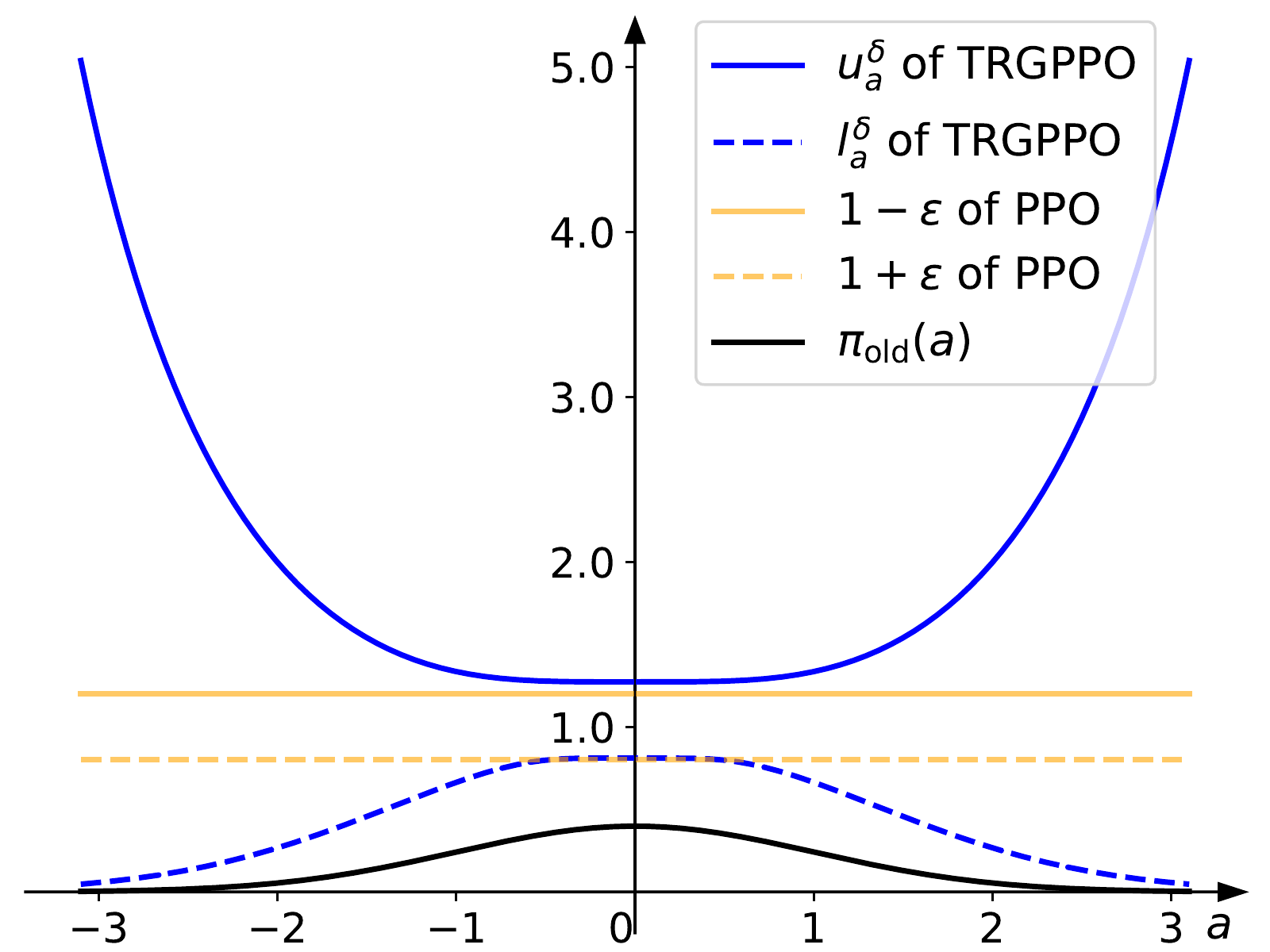} 
			\caption{}\label{fig_kl2clip_gaussian}
		\end{subfigure}
		}
	\end{tiny}
	\iftrue
		\caption{
			 (a) and (b) plot the clipping range and the \feasiblechange/ under different $\pi\old(a|s)$ for discrete action space task. 
			 (c) plots the clipping range under different $a$ for continuous action space task, where $\pi\old(a|s)={\cal N}(a|0,1)$ (black curve).
		}\label{fig_kl2clip}
	\else
	\fi
\end{figure}
%\end{wrapfigure}
\fi

Second, a different underlying metric of the policy difference may result in different algorithm behavior.
In the previous section, we have concluded that PPO's metric with constant clipping range could lead to an exploration issue, due to that it imposes a relatively strict constraint on actions which are not preferred by the old policy.
Therefore, we wish to relax such constraint by enlarging the upper clipping range while reducing the lower clipping range. 
\fig \ref{fig_kl2clip_discrete_ratio} shows the clipping range of \pmethod/ and PPO. 
For \pmethod/ (blue curve), as $\pi\old(a|s)$ gets smaller, the upper clipping range increases while the lower one decreases, which means the constraint is relatively relaxed as $\pi\old(a|s)$ gets smaller.
This mechanism could encourage the agent to explore more on the potential valuable actions which are not preferred by the old policy.
%Whereas the clipping range of PPO (orange curve) is a constant along different $\pi\old(a|s)$. 
We will theoretically show that the exploration behavior with this new clipping range is better than that of with the constant one in \Cref{sec_exploration_our}.

Last but not least, although the clipping ranges are enlarged, it will not harm the stability of learning, as the ranges are kept within the trust region. We will show that this new setting of clipping range would not enlarge the policy divergence and has better performance bound compared to PPO in \Cref{sec_lower_bound}.
%Now let us first describe the method for computation of the adaptive \clippingratio[] of \pmethod/.

Our \pmethod/ adopts the same algorithm procedure as PPO, except that it needs an additional computation of adaptive clipping range.
We now present methods on how to compute the adaptive clipping range defined in \eqref{eq_clippingrange_within_trust} efficiently.
For discrete action space, {by using the KKT conditions}, {the} problem \eqref{eq_clippingrange_within_trust} is transformed into {solving} the following equation {w.r.t} $X$.
\begin{equation}\label{eq_adapative_clippingrange_l}
g( \pi\old(a|s), X ) \triangleq
\left( {1 - \pi\old({a}|{s})} \right)\log \frac{{1 - \pi\old({a}|{s})}}{{1 - \pi\old({a}|{s})X}} 
%{\quad\quad\quad\quad}\\{\quad\quad\quad\quad\quad\quad\quad\quad\quad\quad}
- \pi\old({a}|{s})\log {X} = \delta
\end{equation}
which has two solutions, one is for $l_{s,a}^\delta$ which is within $(0,1)$, and another one is for $u_{s,a}^\delta$ which is within $(1,+\infty)$.
%While problem \eqref{eq_opt_ratioinkl_max} is transformed into solving the same equation but the solution is $u_{s,a}^\delta>1$.
%This equation has two solutions, in which $0<x<1$ is the one for $l_{s,a}^\delta$, while $x>1$ is the one for $u_{s,a}^\delta$. 
%We use MINPACK's HYBRD and HYBRJ routines \citep{powell1970hybrid} as the solver. 
%We use a DNN to output the optimal clipping range for each state-action directly. 
We use MINPACK's HYBRD and HYBRJ routines \citep{powell1970hybrid} as the solver. 
To accelerate this computation procedure, we adopt two additional measures.
%Thus we don't need to prepare large numbers of training data.
%We train a DNN which input these constants and approximately output the solution . 
%We sample finite constants of the optimization problem and obtain solutions from the solver. Then these data are used to train the DNN. 
First, we train a Deep Neural Network (DNN) which input $\pi\old(a|s)$ and $\delta$, and approximately output the initial solution. 
Note that the solution in \eqref{eq_adapative_clippingrange_l} only depends on the probability $\pi\old(a|s)$ and the hyperparameter $\delta$, and it is not affected by the dimension of the action space. Thus it is possible to train one DNN for all discrete action space tasks in advance. 
Second, with fixed $\delta$, we discretize the probability space and save all the solutions in advance.
This clipping range computation procedure with these two acceleration measures only requires only additional 4\% wallclock computation time of the original policy learning. See Appendix \ref{app-sec_computaion_efficiency} for more detail.

While for the continuous actions space task, we make several transformations to make the problem independent of the dimension of the action space, which makes it tractable to apply the two acceleration measures above.
%This approach can significantly accelerate the procedure of computing the adaptive \clippingratio[].
See Appendix \ref{app-sec_opt_continuous} for more detail.

%In the following, we will prove that \pmethod/ has better exploration behavior and lower performance bound than PPO does.
%Although these theoretical results are discussed under discrete action space task, they also holds for the continuous action space tasks when the policy a parametrized distribution like Gaussian or Beta distribution.

\subsection{Exploration Behavior}\label{sec_exploration_our}

In this section, 
we will first give the property of the clipping range of \pmethod/, which could affect the exploration behavior (as discussed in \Cref{sec_problem}).
Then a comparison between \pmethod/ and PPO on the exploration behavior will be provided.
\begin{lemma}\label{thm_clippingrange_pi}
For \pmethod/ with hyperparameter $\delta$, we have $\frac{d u^\delta_{s,a}}{d \pi\old(a|s)} <0$,  $\frac{dl^\delta_{s,a}} {d \pi\old(a|s)}>0$. 

%(\romannumeral2) if $\pi\old(a|s)\rightarrow 0$, then $u^\delta_{s,a}\rightarrow +\infty$ and $l_{s,a}^\delta \rightarrow 0 $ \footnote{Although the clipping range could be quite relaxed, we theoretilly prove that it would not enlarge the policy divergence than PPO does.}.
\end{lemma}
This result implies that the upper clipping range becomes larger as the preference on the action by the old policy $\pi\old(a|s)$ approaches zero, while the lower clipping range is on the contrary. This means that the constraints are relaxed on the actions which are not preferred by the old policy, such that it would encourage the policy to explore more on the potential valuable actions, no matter whether they were preferred by the previous policies or not.

%Enlarging clipping range as $\pi\old(a|s)$ goes smaller means that the restriction on the action which are not preferred by the old policy is relatively relaxed. 
%Such mechanism could reduce sample complexity and encourage the policy to explore more when the policy is initialized from a bad one. We theoretically show this below.
We now give a formal comparison on the exploration behavior.
As mentioned in \Cref{sec_problem}, we measure the exploration ability by the expected distance between the learned policy $\pi_t$ and the optimal policy $\pi^*$ after $t$-step learning, i.e., $\explore{}{t} \triangleq \mathbb{E}_{\pi_t} \left[ \| \pi_t - \pi^*  \|_{\infty} | \pi_0 \right] $. Smaller $\explore{}{t}$ means the better exploration ability. The exploration ability of \pmethod/ is denoted as $\explore{\pmethod/}{t}$ while that of PPO is denoted as $\explore{PPO}{t}$. By \cref{eq_pinew_general} and \Cref{thm_clippingrange_pi}, we get the following conclusion.

\begin{theorem}
For \pmethod/ with hyperparameter $(\delta,\epsilon)$ and PPO with same $\epsilon$.
%if $\delta \leq g( \max_{a\in  {\cal A}\subopta } \pi_0(a),  1+\epsilon )$, then we have $\explore{\pmethod/}{t} \leq \explore{PPO}{t}$ for any $t$.
%(\romannumeral1)
If  $ \delta \leq g( \max_{a\in  {\cal A}\subopta } \pi\oldb(a),  1+\epsilon ) $ for all $t$, then we have $\explore{\pmethod/}{t} \leq \explore{PPO}{t}$ for any $t$.  
%Furthermore, with the previous condition holds, if $ \delta >  g(  \pi_0(a\opta),  1+\epsilon ) $, then $\explore{\pmethod/}{t} < \explore{PPO}{t}$.
\end{theorem}
This theorem implies that our \pmethod/ has better exploration ability than PPO, with proper setting of the hyperparameter $\delta$.
%Furthermore, \pmethod/ shows \ttt{overwhelming} advantage than PPO when the initial policy is much far from the optimal one,  
%which quite possibly occurs in high dimensional action space tasks.

\subsection{Policy Divergence and Lower Performance Bound}\label{sec_lower_bound}
%\ttt{
%In this section, we make a comparison on the policy divergence and the lower performance bound, whose improvement leads to guaranteed performance improvement (see \Cref{thm_lowerbound}). 
To investigate how \pmethod/ and PPO perform in practical, let us consider an empirical version of lower performance bound:
$
%\begin{equation}
{\hat{M}_{\pi\old}}(\pi ) = 
{\hat{L}_{\pi\old}}(\pi ) - C\max_t{D}_{\rm{KL}}^{s_t}\left( {\pi\old,\pi } \right),
%\end{equation}
$
where ${\hat{L}_{\pi\old}}(\pi )=\frac{1}{T}\sum_{t=1}^T{\left[{ \ratio{t} \adv }\right]}+\hat{\eta}^{\pi\old}$, $s_t\sim{\rho _{\pi\old}},a_t\sim\pi\old(\cdot|s_t)$ are the sampled states and actions, where we assume $s_i\neq s_j$ for any $i\neq j$, $\adv$ is the estimated value of $A^{\pi\old}(s_t,a_t)$, $\hat{\eta}^{\pi\old}$ is the estimated performance of old policy $\pi\old$.

Let $\Piopt{PPO}$ denote the set of all the optimal solutions of the empirical surrogate objective function of PPO, and let $\piopt{PPO} \in \Piopt{PPO}$ denote the optimal solution which achieve minimum KL divergence over all optimal solutions, i.e., $D_{\rm KL}^{s_t}(\pi\old,\piopt{PPO})\leq D_{\rm KL}^{s_t}(\pi\old,\pi)$ for any $\pi \in \Piopt{PPO}$ under all $s_t$. 
%As we have make assumption that the sampled states are different, we can find such optimal $\piopt{PPO}$. 
This problem can be formalized as $ \piopt{PPO}= \argmin_{\pi \in \Piopt{PPO}} \left( D_{\rm KL}^{s_1}(\pi\old,\pi), \ldots, D_{\rm KL}^{s_T}(\pi\old,\pi) \right) $. 
Note that $\pi(\cdot|s_t)$ is a conditional probability and the optimal solution on different states are independent from each other.
Thus the problem can be optimized by independently solving $\min_{\pi(\cdot|s_t) \in \{\pi(\cdot|s_t) : \pi \in \Piopt{PPO} \}}  D_{\rm KL}\left(\pi\old(\cdot|s_t),\pi(\cdot|s_t)\right)$ for each $s_t$. 
The final $\piopt{PPO}$ is obtained by integrating these independent optimal solutions $\piopt{PPO}(\cdot|s_t)$ on different state $s_t$.
%The corresponding optimal solution $\piopt{PPO}(\cdot|s_t)$ for different $s_t$ are assembled into a complete $\piopt{PPO}$ which achieves minimum KL divergence on each state $s_t$.
%could be decomposed by solving $\pi(\cdot|s_t)$ for each $s_t$, since the solution of $\pi(\cdot|s_t)$ are independent.
Similarly, $\piopt{\pmethod/}$ is the one of \pmethod/ which has similar definition as $\piopt{PPO}$.
Please refer to Appendix \ref{app-sec_proof} for more detail.

%, and let $\piopt{}$ denote the one with general setting of clipping range $(l_{s,a},u_{s,a})$.
%\begin{lemma}
%\begin{equation}
%\text{For $\adv<0$, we have }\quad
%\piopt{}(a|s_t) = 
%\begin{cases}
%\frac{\pi\old(a|s_t)(1-\pi\old(a_t|s_t)l_{s_t,a_t})}{1-\pi\old(a_t|s_t)} & a\neq a_t\\
%\pi\old(a_t|s_t)l_{s_t,a_t} & a=a_t
%\end{cases}
%{\quad\quad\quad}
%\end{equation}
%\begin{equation}
%\text{For $\adv>0$, we have }\quad
%\piopt{}(a|s_t) = 
%\begin{cases}
%\frac{\pi\old(a|s_t)(1- \min(\pi\old(a_t|s_t)u_{s_t,a_t},1) )}{1-\pi\old(a_t|s_t)} & a\neq a_t\\
%\min(\pi\old(a_t|s_t)u_{s_t,a_t},1)&a=a_t
%\end{cases}
%\end{equation}
%\end{lemma}

To analyse \pmethod/ and PPO in a comparable way, we introduce a variant of \pmethod/.
The hyperparameter $\delta$ of \pmethod/ in \cref{eq_clippingrange_within_trust} is set adaptively by $\epsilon$. That is,
$
%\begin{equation}
{\delta} = \max\left( \left( {1 - {p^ + }} \right)\log \frac{{1 - {p^ + }}}{{1 - {p^ + }(1 + \epsilon)}} - {p^ + }\log (1 + \epsilon), 
%\\
%{\quad\quad\quad}
\left( {1 - {p^ - }} \right)\log \frac{{1 - {p^ - }}}{{1 - {p^ - }(1 - \epsilon)}} - {p^ - }\log (1 - \epsilon) \right),
%\end{equation}
$
where $p^+=\max\limits_{t:\adv>0} \pi\old(a_t|s_t)$, $p^-=\max\limits_{t:\adv<0} \pi\old(a_t|s_t)$. One may note that this equation has a similar form to that of \cref{eq_adapative_clippingrange_l}. In fact, if \pmethod/ and PPO share a similar $\epsilon$, then they have the same KL divergence theoretically. We conclude the comparison between \pmethod/ and PPO by the following theorem.

\begin{theorem}\label{thm_comparison}
Assume that $\max_{t}D_{\rm KL}^{s_t}(\pi\old, \piopt{PPO})<+\infty$ for all $t$.
If \pmethod/ and PPO have the same hyperparameter $\epsilon$,
we have: 

(\romannumeral1) $u_{s_t,a_t}^\delta \geq 1+\epsilon$ and $l_{s_t,a_t}^\delta \leq 1-\epsilon$ for all $(s_t,a_t)$;

(\romannumeral2) $\max_t {{D}_{\rm KL}^{{s_t}}(\pi\old,\piopt{\pmethod/} )} = \max_t {{D}_{\rm KL}^{{s_t}}(\pi\old, \piopt{PPO})}$;

(\romannumeral3) 
%if there exists at least one $(s_{t},a_{t})$ that satisfies Assumption \ref{asp_pi}, then 
%\begin{equation*}
$
{{\hat{M}}_{{\pi\old}}}(\piopt{\pmethod/}) \geq {{\hat{M}}_{{\pi\old}}}( \piopt{PPO} )
$.
%${\hat{L}_{\pi\old}}(\pi )=\frac{1}{T}\sum_{t=1}^T{\left[{{r_\pi }({s_t},{a_t})\adv }\right]}$.
%\end{equation*}
Particularly, if there exists at least one $(s_{ t},a_{ t})$ such that $\pi\old(a_{ t}|s_{ t}) \neq \max\limits_{\hat t:\adv[{\hat t}] <0} \pi\old(a_{\hat t}|s_{\hat t}) $ and $\pi\old(a_{ t}|s_{ t}) \neq \max\limits_{{\hat t}:\adv[{\hat t}] >0} \pi\old(a_{\hat t}|s_{\hat t})$,
then ${{\hat{M}}_{{\pi\old}}}( \piopt{\pmethod/} ) > {{\hat{M}}_{{\pi\old}}}( \piopt{PPO} )$.
\end{theorem}
%The theorem implies that \pmethod/ could provide better empirical performance guarantee, thus performs more sample efficient.
%Conclusion (\romannumeral1) \ttt{implies} that the \clippingratio[s] in \pmethod/ are enlarged compared to PPO. 
\ttt{
Conclusion (\romannumeral1) implies that \pmethod/ could enlarge the \clippingratio[s] compared to PPO and accordingly allow larger update of the policy. 
Meanwhile, the maximum KL divergence is retained, which means \pmethod/ would not harm the stability of PPO theoretically.
Conclusion (\romannumeral3) implies that \pmethod/ has better empirical performance bound.
}

\section{Experiment}
We conducted experiments to answer the following questions:
(1) Does PPO suffer from the \emph{lack of exploration} issue?
(2) Could our \pmethod/ relief the exploration issue and improve sample efficiency compared to PPO?
(3) Does our \pmethod/ maintain the stable learning property of PPO?
To answer these questions, we first evaluate the algorithms on two simple bandit problems and then compare them on high-dimensional benchmark tasks. 
%To answer (1), we evaluate \pmethod/ and PPO on the bandit problems. To answer (2) and (3), we compare \pmethod/ and PPO on high-dimensional benchmark tasks.
%\begin{itemize}
%	\item The \clippingratio[s] of \pmethod/ are set adaptively to different state-action pair while that of PPO are fixed. How do these changes affect the  sample efficiency of the algorithm?
%	\item As our adaptive mechanism implies that smaller probability of action lead to larger \clippingratio[], which intuitively make this approach less susceptible to get trapped in local optima. Could \pmethod/ contribute to find out better solutions?
%	\item As our analysis result implies that \pmethod/ could enlarge the \clippingratio[s] while remain the maximum KL divergence \ttt{maintained}. Does the reality conform to theoretical analysis?
%\end{itemize}

\subsection{Didactic Example: Bandit Problems}\label{sec_experiment_bandit}
%In order to verify that PPO suffers from the \ttt{lack of exploration} issue and that our \pmethod/ method could relieve this problem, 
We first evaluate the algorithms on the bandit problems. 
In the continuous-armed bandit problem, the reward is $0.5$ for $a\in(1,2)$; $1$ for $a \in(2.5,5)$; and $0$ otherwise. And a Gaussian policy is used.
The discrete-armed bandit problem is defined in section \ref{sec_problem}.We use a {Gibbs} policy $\pi(a)\propto \exp(\theta_a)$, where the parameter $\theta$ is initialized randomly from $  {\cal N}(0,1)$.
%an additional discrete-armed bandit problem, in which 
%For the discrete-armed bandit problem, the action space is ${\cal A}=\{a_{\rm subopt}, a_{\rm opt}, a_{\rm worst}\}$ and the reward function $c$ is defined as $c(a_{\rm subopt})=0.5,c(a_{\rm opt})=1,c(a_{\rm worst})=-50$.
%The reward function of discrete-armed bandit problem is defined as 
%\begin{center}
%\begin{tabular}{| c | c | c | c | }
%\hline
%Action & $a_{\rm subopt}$ & $a_{\rm opt}$ & $a_{\rm worst}$ \\ 
%\hline
%Reward & 0.5 & 1 & -50   \\
%\hline
%\end{tabular}
%\end{center}
%\begin{equation}
%c(a)=
%\begin{cases}
%0.5, a=a_{\rm subopt}\\
%1, a=a_{\rm opt}\\
%-50, a=a_{\rm worst}
%\end{cases}
%\end{equation}
%For discrete-armed bandit problem, 
%For \pmethod/ we use $\delta=0.03$, while for PPO we use $\epsilon=0.2$. 
We also consider the vanilla Policy Gradient method as a comparison.
Each algorithm was run for 1000 iterations with 10 random seeds.

%For continuous-armed bandit problem, 
%we use Gaussian policy $\pi(a)={\cal N}(a|\mu,\exp(\sigma))$, where $\mu, \sigma$ are the parameters initialized with $\mu \sim {\cal N}(0,0.01)$, $\sigma=0$. 

%The learning rate is 0.01 for all algorithms and problems.
% DISCARD: Reward function
%For discrete-armed bandit problem, ${\cal A}=\{a_{\rm opt},a_{\rm subopt},a_{worst}\}$,
%%reward function is $c(a_1)=0.5, c(a_2)=1, c(a_3)=-10$. 
%the policy $\pi(a)\propto \exp(\theta_a)$, where $\theta$ is the parameter initialized with $\theta_a \sim {\cal N}(0,1)$.
%For continuous-armed bandit problem, ${\cal A}={\mathbb R}$,
%%reward function is $c(a)=0.5$, when $ 1\leq a \leq 2$; $c(a)=1$, when $ 2.5\leq a \leq 3.5$; $c(a)=0$, otherwise. 
%the policy $\pi(a)={\cal N}(a|\mu,\exp(\sigma))$, where $\mu, \sigma$ are the parameters initialized with $\mu \sim {\cal N}(0,0.01)$, $\sigma=0$.
%Th reward function for these two bandit problems are 
%\begin{equation*}
%c_{\rm d}(a)=
%\begin{cases}
%0.5, a=a_{\rm opt}\\
%1, a=a_{\rm subopt}\\
%-50, a=a_{\rm worst}
%\end{cases}
%c_{\rm c}(a)=
%\begin{cases}
%0.5, 1\leq a \leq 2\\
%1, 2.5\leq a \leq 4\\
%0, otherwise
%\end{cases}
%\end{equation*}
%respectively. \fig \ref{fig_bandit_continuous} shows the reward function of continuous-armed bandit problem (\ttt{purple} dashed curve).We compare \pmethod/ and PPO on these two tasks.

\ifnofigure
\else
%\begin{figure}[htbp]
\begin{wrapfigure}{R}{0.4\textwidth}
\centering
%\begin{minipage}[t]{0.46\textwidth}
\includegraphics[width=1\linewidth]{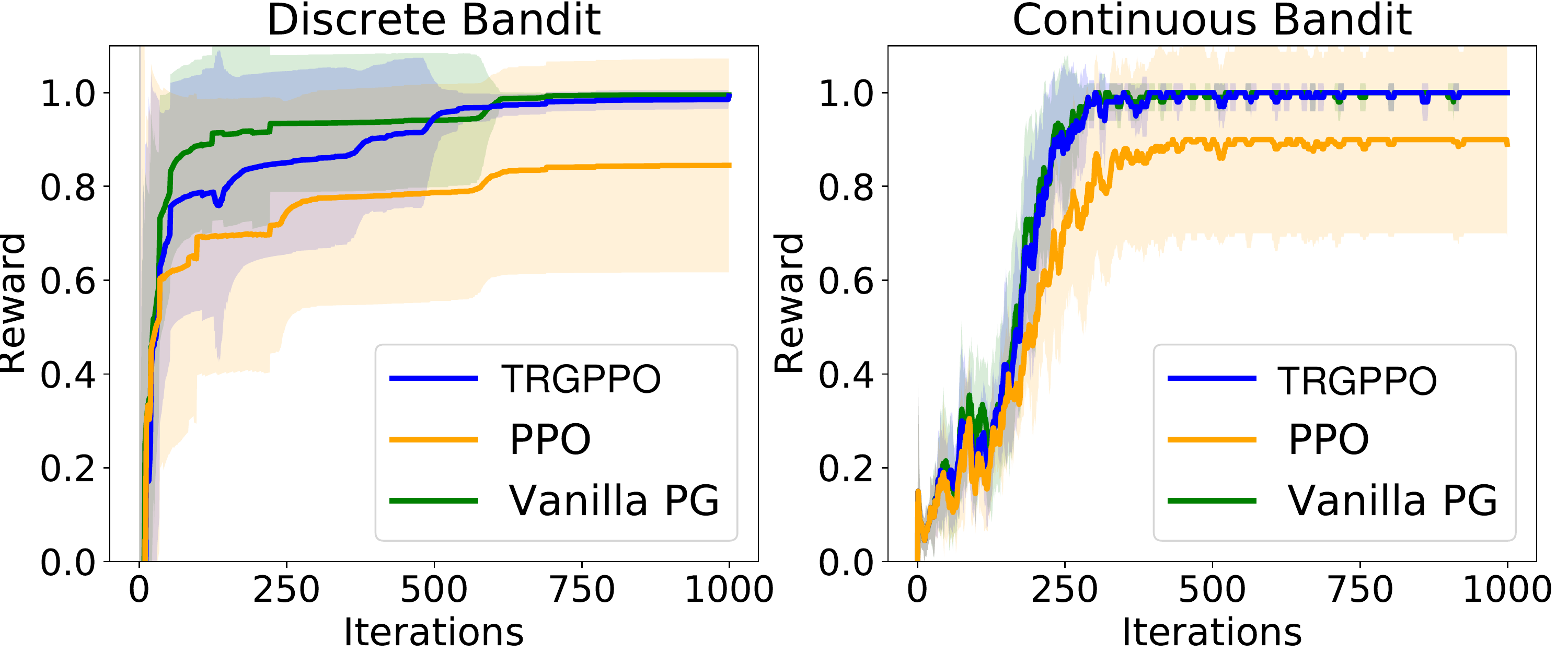}
	\caption[width=\columnwidth]{
		The performance on discrete and continuous-armed bandit problems during training process.
%		PPO (orange curve) performs susceptible to get trapped in local optima, while \pmethod/ (black curve) could always find the optimal policy.
	}\label{fig_bandit_performance}
%\end{minipage}
%\hspace{ 0.01\textwidth }
%\end{figure}
\end{wrapfigure}
\fi

%\ttt{For PPO, 40\% of the cases get trapped in local optima on the discrete-armed bandit problem and 30\% of the cases get trapped in local optima on the continuous one. However, our \pmethod/ can always find optimal solution in both discrete and continuous-armed bandit problems.}
%Table \ref{tab_bandit_result} shows the the success rate of achieving optimal solution in $10$ trials. We judge the solution is an optimal one if it has largest probability on optimal actions.
\fig \ref{fig_bandit_performance} plots the performance during the training process.  
\ttt{
PPO gets trapped in local optima at a rate of 30\% and 20\% of all the trials on discrete and continuous cases respectively, while our \pmethod/ could find the optimal solution on almost all trials.}
For continuous-armed problem, we have also tried other types of parametrized policies like Beta and Mixture Gaussian, and these policies behaves similarly as the Gaussian policy.
In discrete-armed problem, we find that when the policy is initialized with \ttt{a local optima, PPO could easily get
 trapped in that one.} 
Notably, since vanilla PG could also find the optimal one, \ttt{it could be inferred that the exploration issue mainly derives from the ratio-based clipping with constant clipping range}.
%These results are consistent with our analysis in \Cref{sec_problem}.
%On both tasks, PPO is more likely to get trapped in sub-optimal solution, while \pmethod/ could find the optimal one on the most cases.
%Vanilla PG method also fails same as PPO. This is because the vanilla PG method does not enforce constraint on the policy updating, thus \ttt{the policy may be over-optimized into a locally optimal one}.

\subsection{Evaluation on Benchmark Tasks}

We evaluate algorithms on benchmark tasks implemented in OpenAI Gym \citep{Brockman2016OpenAI}, simulated by MuJoCo \citep{Todorov2012MuJoCo} and Arcade Learning Environment \citep{bellemare2013arcade}.
%using the  
For continuous control tasks, we evaluate algorithms on {6 benchmark tasks}.
% (including a challenging high-dimensional Humanoid locomotion task).
%, defined in MuJoCo physics engine \citep{Todorov2012MuJoCo}. 
{All tasks were run with 1 million timesteps except that the Humanoid task was 20 million timesteps.}  
The trained policies are evaluated after sampling every 2048 timesteps data.
The experiments on discrete control tasks are detailed in Appendix \ref{app-sec_atari}.

\ifnofigure
\else
%	\begin{wrapfigure}{R}{0.6\textwidth}
	\begin{figure}[!tb]
		\centerline{
  			\includegraphics[width=0.33\linewidth]{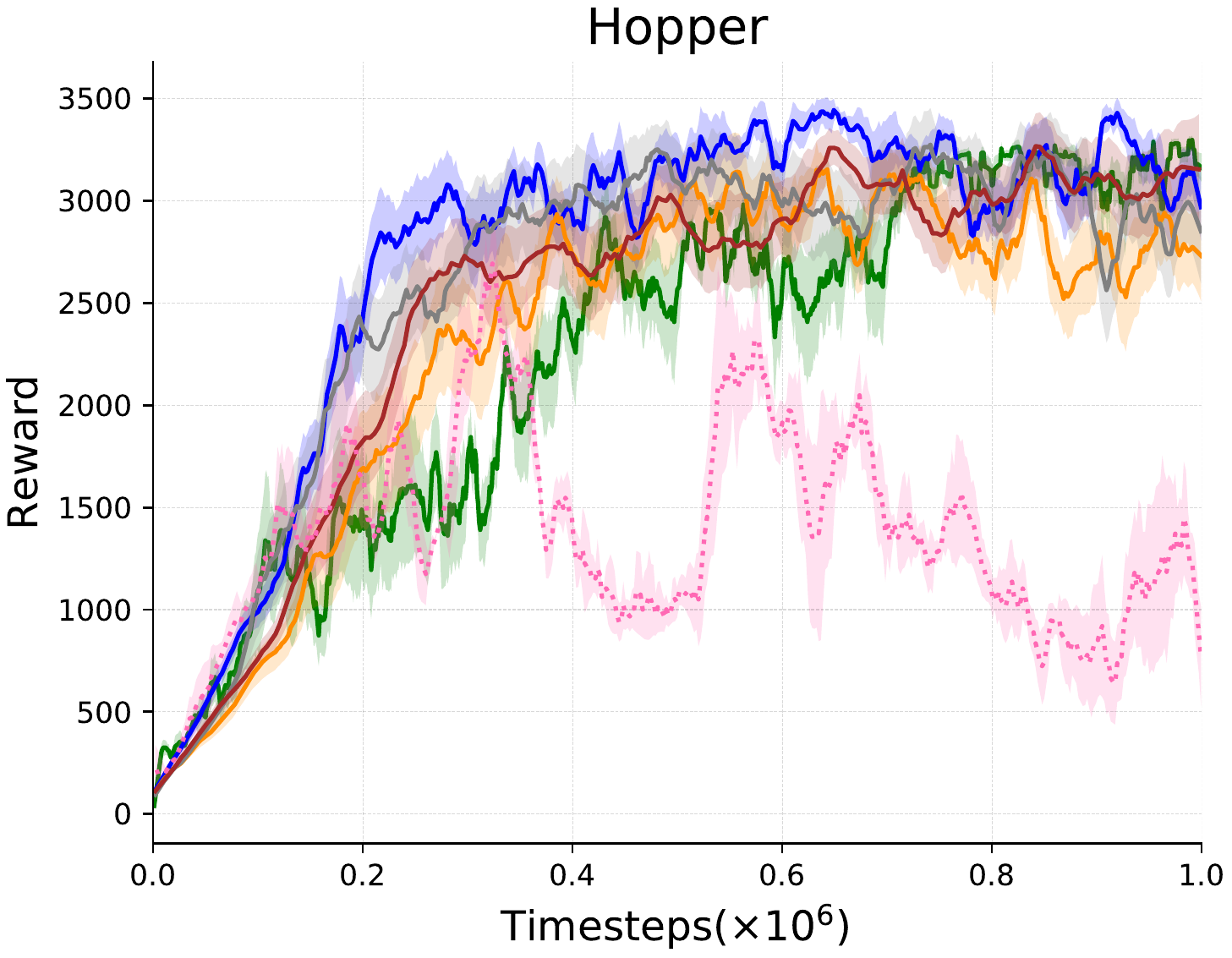}
  			\includegraphics[width=0.33\linewidth]{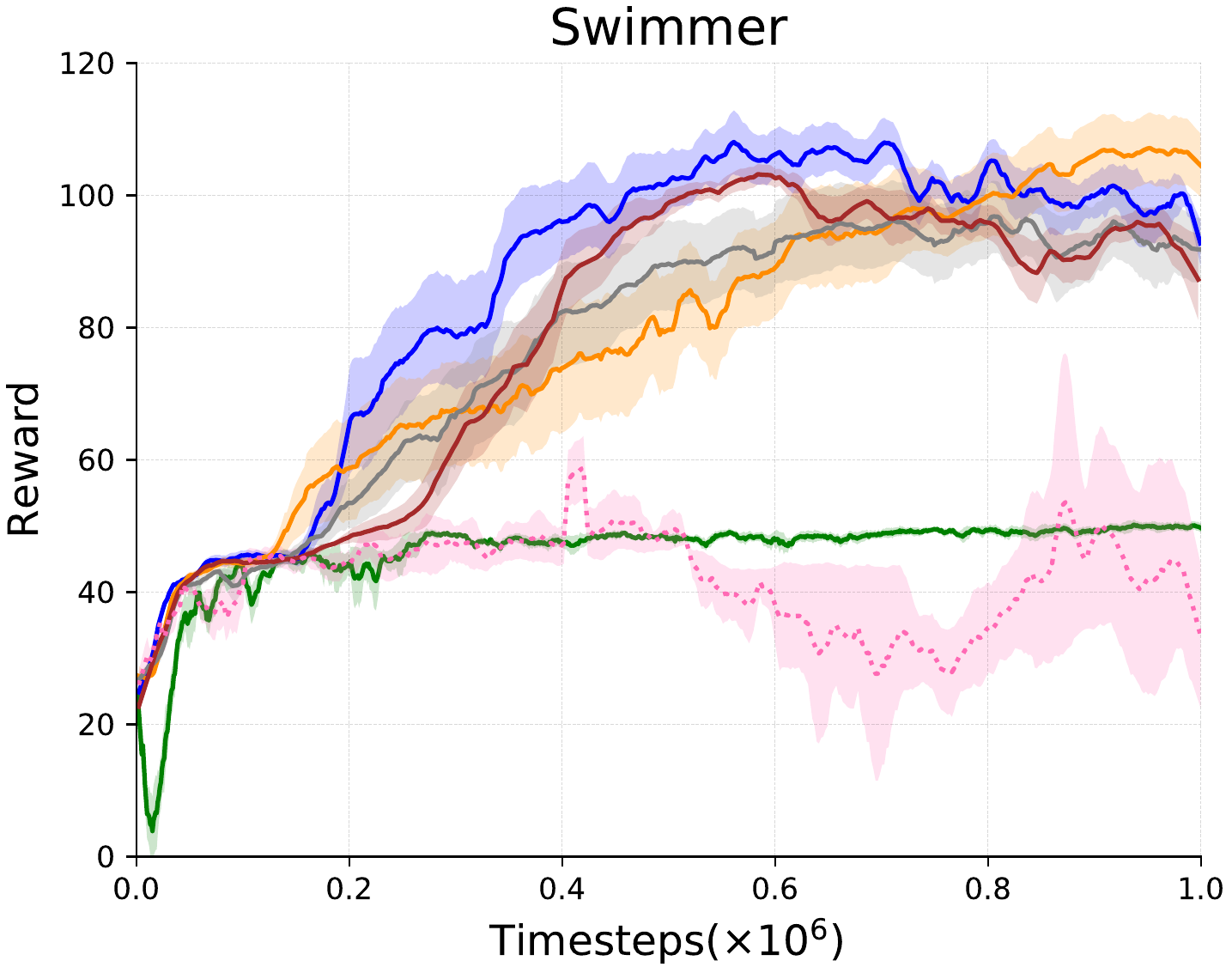}
  			\includegraphics[width=0.33\linewidth]{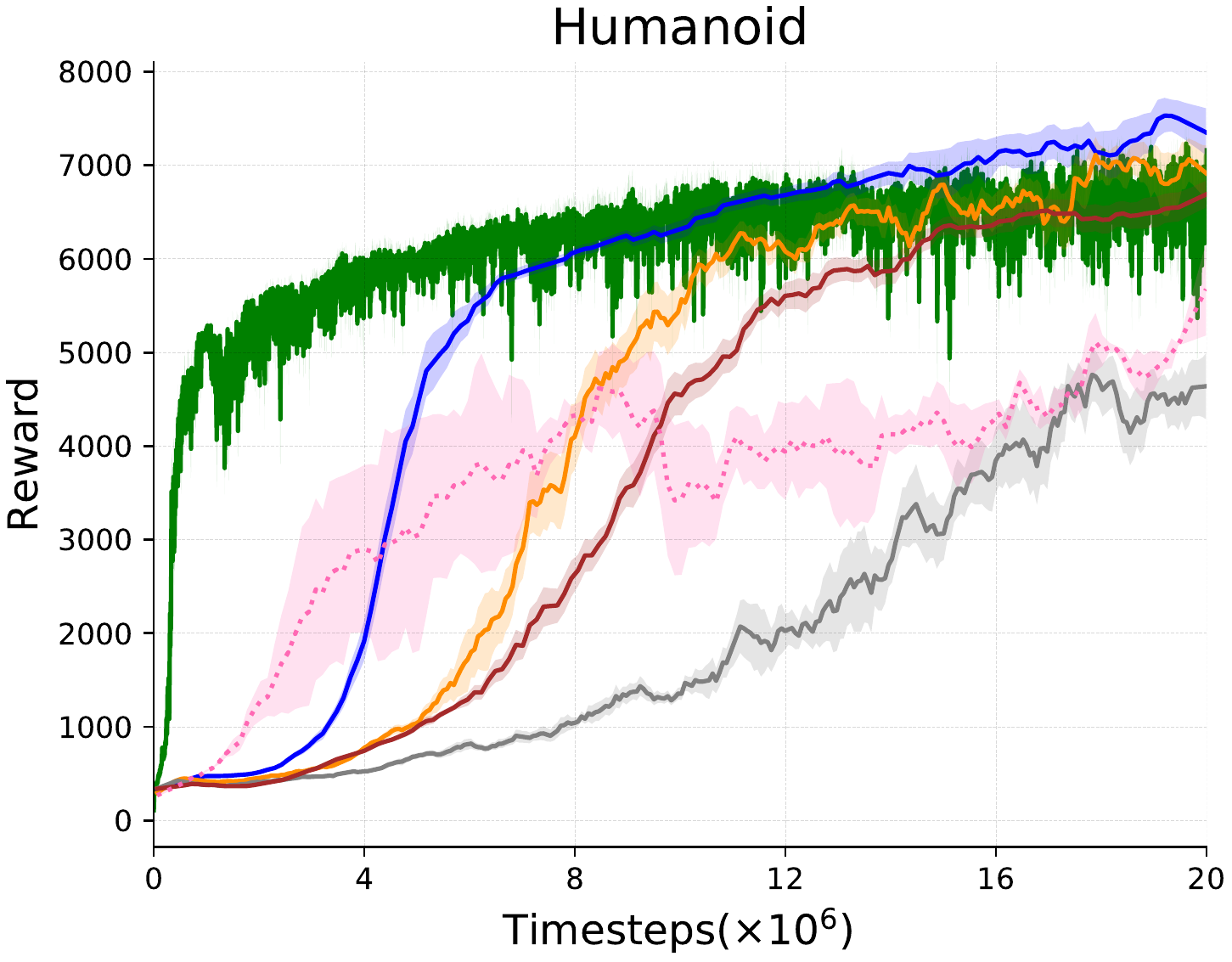}
  		}
  		\centerline{
	  		\includegraphics[width=0.33\linewidth]{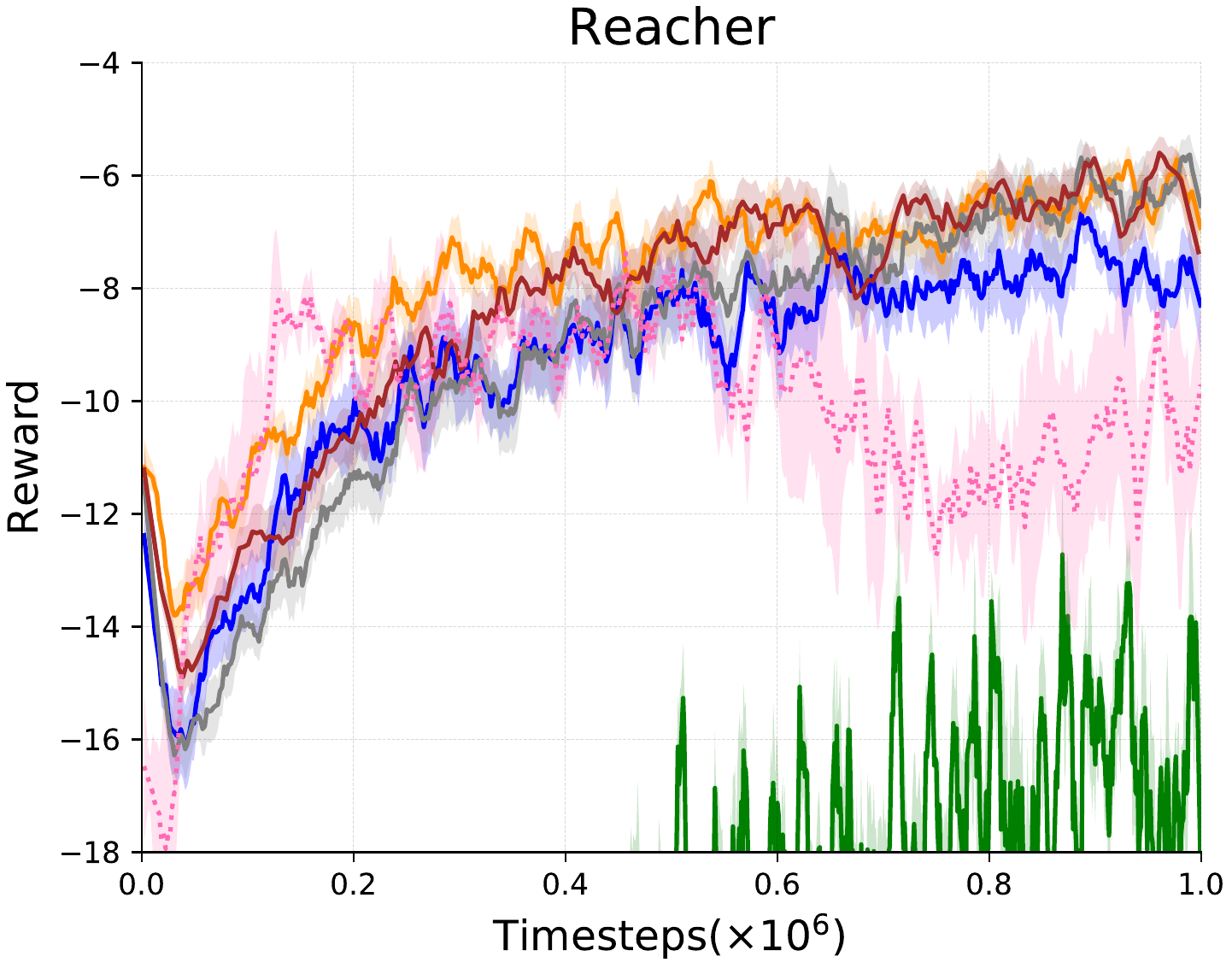}
	  		\includegraphics[width=0.33\linewidth]{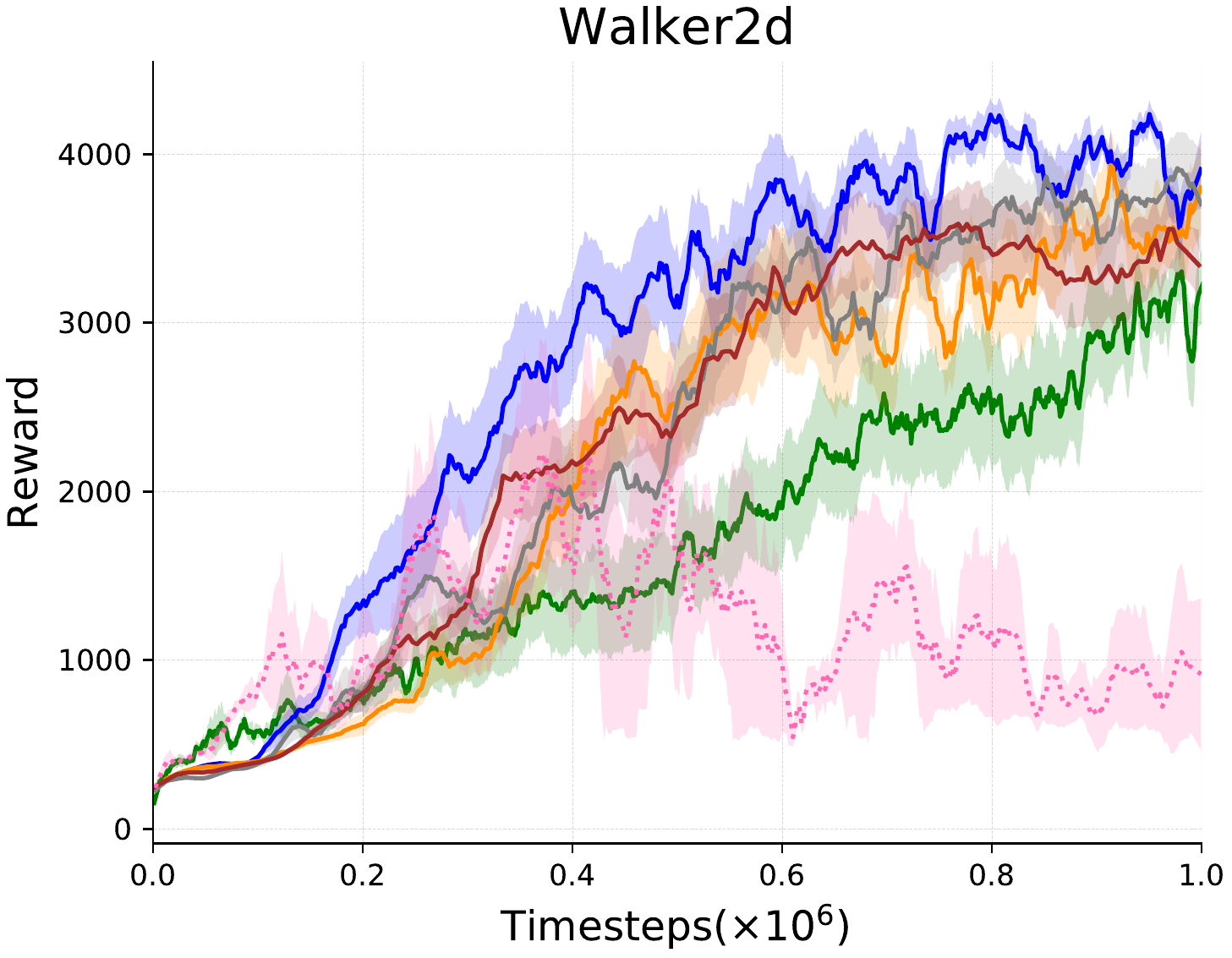}
	  		\includegraphics[width=0.33\linewidth]{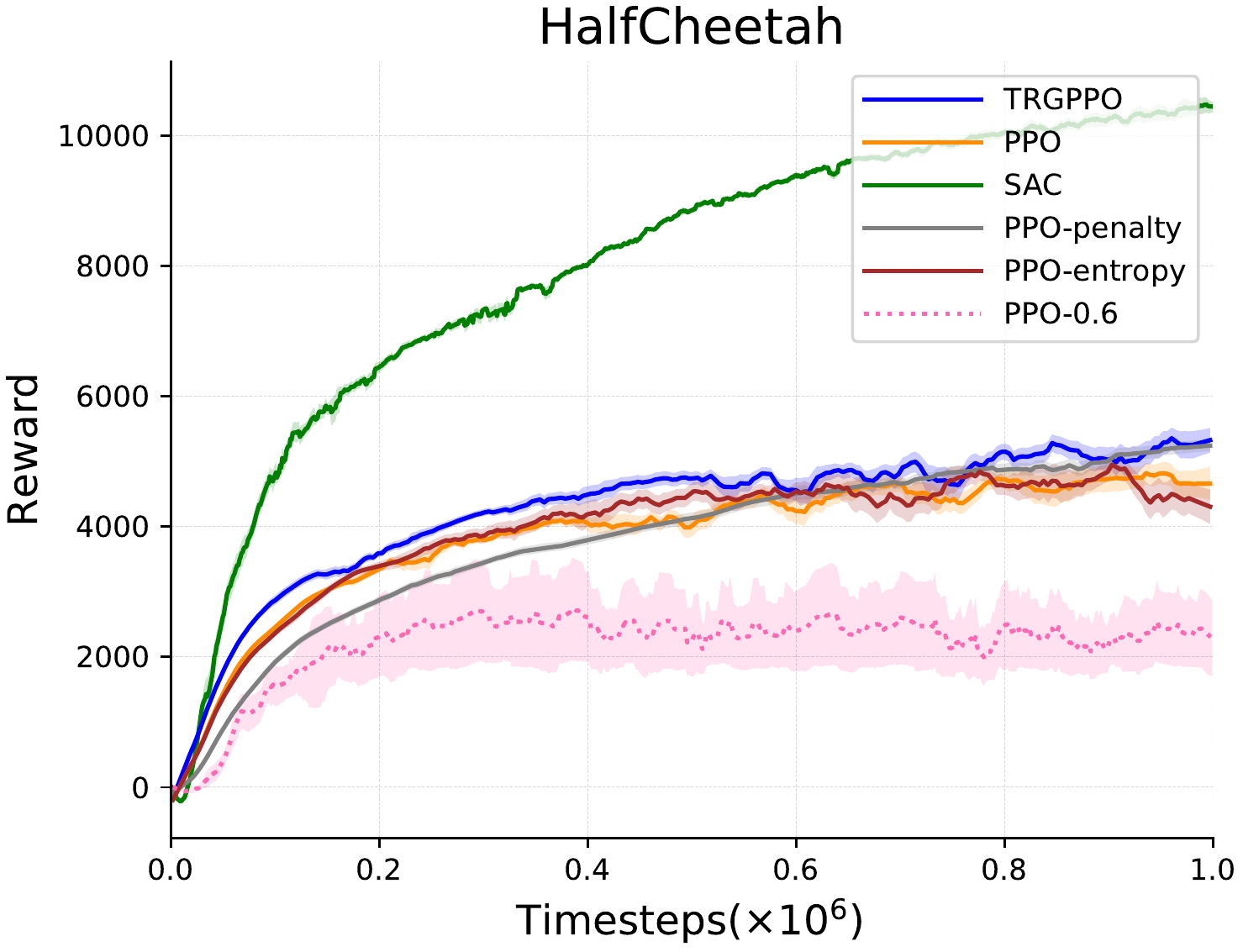}
  		}
%		\centering
%		\includegraphics[width=0.6\columnwidth]{/media/d/yunwithhugo/rew.pdf}
		\caption{
		Episode rewards during the training process; the shaded area indicate the standard deviation over \numrandomseed/ random seeds.  
		}
		\label{fig_rew}
	\end{figure}
%	\end{wrapfigure}
\fi

\begin{table}[!t]
	\footnotesize
	\setlength{\tabcolsep}{0pt}
	\centering
	\caption{
	Results of timesteps to hit a threshold within 1 million timesteps (except Humanoid with 20 million) and averaged rewards over last 40\% episodes during training process.
	}\label{tab_reward_hit}
		\begin{tabular}{
			@{}
			+m{0.6in}<{\centering}
			Ym{0.55in}<{\centering}
			Ym{0.55in}<{\centering}Ym{0.55in}<{\centering}Ym{0.55in}<{\centering}Ym{0.5in}<{\centering}
			|
			Ym{0.55in}<{\centering}Ym{0.55in}<{\centering}Ym{0.55in}<{\centering}Ym{0.5in}<{\centering}
			@{}
			}
			\toprule 
			& \multicolumn{5}{c} \scriptsize{(a) Timesteps to hit threshold ($\times 10^3$) } & \multicolumn{4}{c} \scriptsize{(b) Averaged rewards \quad\quad\quad   }   \\
			\noalign{\smallskip}
         \rowstyle{} & Threshold & \pmethod/ & PPO   & {PPO-penalty}   & SAC     & \pmethod/ & PPO   &  {PPO-penalty}  & SAC 
                   \\
			\midrule{}
			%\multirow{8}{*}{$\epsilon=0.0$}										
%			 Humanoid  &  5000  &  7072 & 9088 & \textbf{386} & {390}   & 7276.5 & \textbf{7530.7} & 7440.4 & 5812.3     \\  
%			 Reacher  &  -5  &  249 & \textbf{129} & 248 & {212}   & -1.2 & -3.2 & -2.1 & \textbf{-0.7}     \\  
%			 Swimmer  &  90  &  \textbf{225} & 490 & /\footnotemark & /   & \textbf{122.5} & 102.3 & 80 & 45.2     \\  
%			 HalfCheetah  &  2100  &  {275} & /  & \textbf{31} & /   & 3948 & 1627.5 & \textbf{11007.6} & 2113.2     \\  
%			 Hopper  &  3000  &  \textbf{165} & 409 & 178 & {523}   & \textbf{3693.4} & 3615.2 & 3334.7 & 3549.7     \\  
%			 Walker2d  &  3000  &  \textbf{215} & 587 & 554 & /   & \textbf{5059.4} & 4054.6 & 3788 & 2520.9     \\ 
    Humanoid &  5000 &            4653 &            7241 &  13096.0 &  \textbf{343.0} &  \textbf{7074.9} &         6620.9 &  3612.3 &           6535.9 \\
     Reacher &    -5 &             201 &  \textbf{178.0} &    301.0 &             265 &             -7.9 &  \textbf{-6.7} &    -6.8 &            -17.2 \\
     Swimmer &    90 &  \textbf{353.0} &             564 &    507.0 &               /\footnotemark &   \textbf{101.9} &          100.1 &    94.1 &               49 \\
 HalfCheetah &  3000 &             117 &             148 &    220.0 &   \textbf{53.0} &           4986.1 &         4600.2 &  4868.3 &  \textbf{9987.1} \\
      Hopper &  3000 &  \textbf{168.0} &             267 &    188.0 &             209 &  \textbf{3200.5} &         2848.9 &  3018.7 &           3020.7 \\
    Walker2d &  3000 &  \textbf{269.0} &             454 &    393.0 &             610 &  \textbf{3886.8} &         3276.2 &    3524 &             2570 \\
			\bottomrule
		\end{tabular}
	\end{table}

%We compare variants of \pmethod/ with PPO and state-of-art model-free policy gradient methods, ACTKR\citep{wu2017scalable}.
For our \emph{\pmethod/}, the trust region coefficient $\delta$ is adaptively set by tuning $\epsilon$ (see Appendix \ref{app-sec_set_delta_by_epsilon} for more detail). We set $\epsilon=0.2$, same as PPO.
%TODO: see for more detail....
The following algorithms were considered in the comparison.
%(a) \emph{\pmethod/}: our method by tuning $\delta$. We set $\delta=0.03$ for all tasks (except for the high-dimensional Humanoid task we use $\delta=0.003$). 
%\ttt{We also tested other $\delta$. For low-dimensional tasks($\mathop{dim}({\cal A})\leq10$), we found that the performance changed little when $\delta$ is within $(0.01,0.04)$. While for relatively high-dimensional tasks ($\mathop{dim}({\cal A})>10$), we recommend $\delta$ to be within $(0.001,0.004)$.}
%(a) 
%an alternative approach to tune hyperparameter of \pmethod/, where $\delta$ is computed automatically according to a given $\epsilon$ to make \pmethod/ and PPO has approximately KL divergence during the training process, as Theorem \ref{thm_comparison} implies. The calculation of $\delta$ given $\epsilon$ is detailed in \ttt{appendix}. We set $\epsilon=0.2$, same as PPO.
(a) \emph{PPO}: we used $\epsilon=0.2$ as recommended by \citep{schulman2017proximal}. 
(b) \emph{PPO-entropy}: PPO with an explicit entropy regularization term $\beta {\mathbb E}_s \left[ H\left(\pi\old(\cdot|s), \pi(\cdot|s) \right) \right]$, where $\beta=0.01$.
(c) \emph{PPO-$0.6$}: PPO with a larger \clippingratio[] where $\epsilon=0.6$. 
%We evaluate such setting to investigate the effect of simply enlarging \clippingratio[]. 
%(e) \emph{ACKTR}: {a popular model-free policy optimization algorithm \citep{wu2017scalable}.}
(d) \emph{PPO-penalty}:  a variant of PPO which imposes a penalty on the KL divergence and adaptively adjust the penalty coefficient \cite{schulman2017proximal}.
(e) \emph{SAC}: Soft Actor-Critic, a state-of-the-art off-policy RL algorithm \citep{haarnoja2018soft}.
Both \pmethod/ and PPO adopt exactly same implementations and hyperparameters except the \clippingratio[] based on OpenAI Baselines \citep{baselines}. This ensures that the differences are due to algorithm changes instead of implementations or hyperparameters. 
%For ACKTR, we adopt the implementations and hyperparameters in \citep{baselines}. 
For SAC, we adopt the implementations provided in \citep{haarnoja2018soft}. 
%Each algorithm was run with \numrandomseed/ random seeds.

%\todo{PPO set clipping range by $\delta$, PPO.....}
%The hyperparameters are detailed in Appendix \ref{app-sec_implementation}.

	\textbf{Sample Efficiency:}
	Table \ref{tab_reward_hit} (a) lists the timesteps required by algorithms to hit a prescribed threshold within 1 million timesteps and \Cref{fig_rew} shows episode rewards during the training process. The thresholds for all tasks were chosen according to \citep{wu2017scalable}.  
%	As can be seen in Table \ref{tab_reward_hit}, \pmethod/ reaches the threshold in quite less timesteps than PPO on almost all the tasks except Reacher. 
%	Notably, on Swimmer, Walker2d and Hopper, \pmethod/ requires almost only 3/5 timesteps of \ttt{PPO}.
	As can be seen in Table \ref{tab_reward_hit}, \pmethod/ requires about only 3/5 timesteps of \ttt{PPO} on 4 tasks except HalfCheetah and Reacher.

			\footnotetext{`/' means that the method did not reach the reward threshold within the required timesteps on all the seeds.}
	
\begin{figure}[!t]
	%	\centerline{
	%	\captionsetup[sub]{font=tiny,labelfont={bf,sf}}
	%	\begin{minipage}[]
		\centerline{
			\begin{subfigure}[]{0.33\linewidth}
				\centerline{
		  			\includegraphics[width=0.5\linewidth]{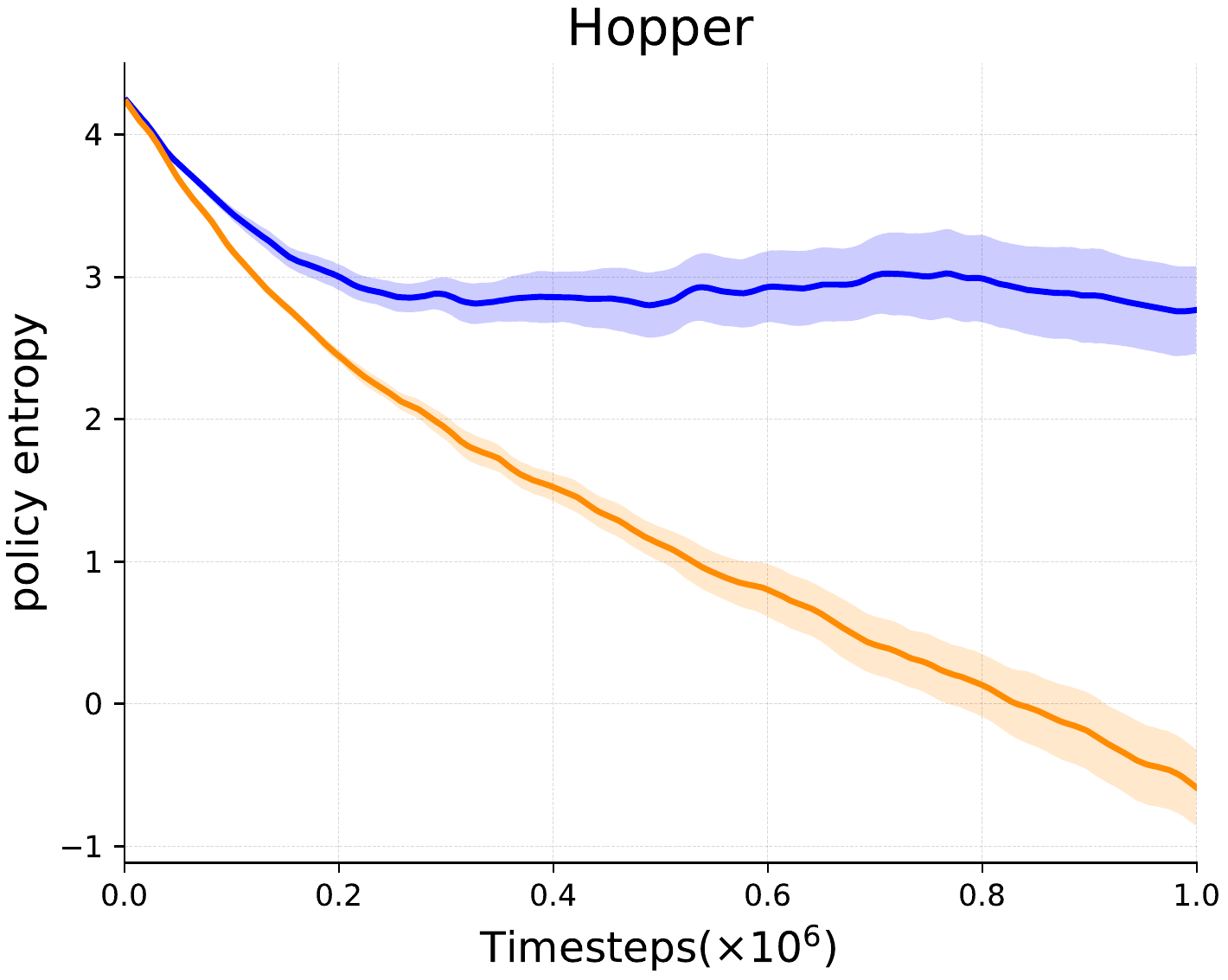}
		  			\includegraphics[width=0.5\linewidth]{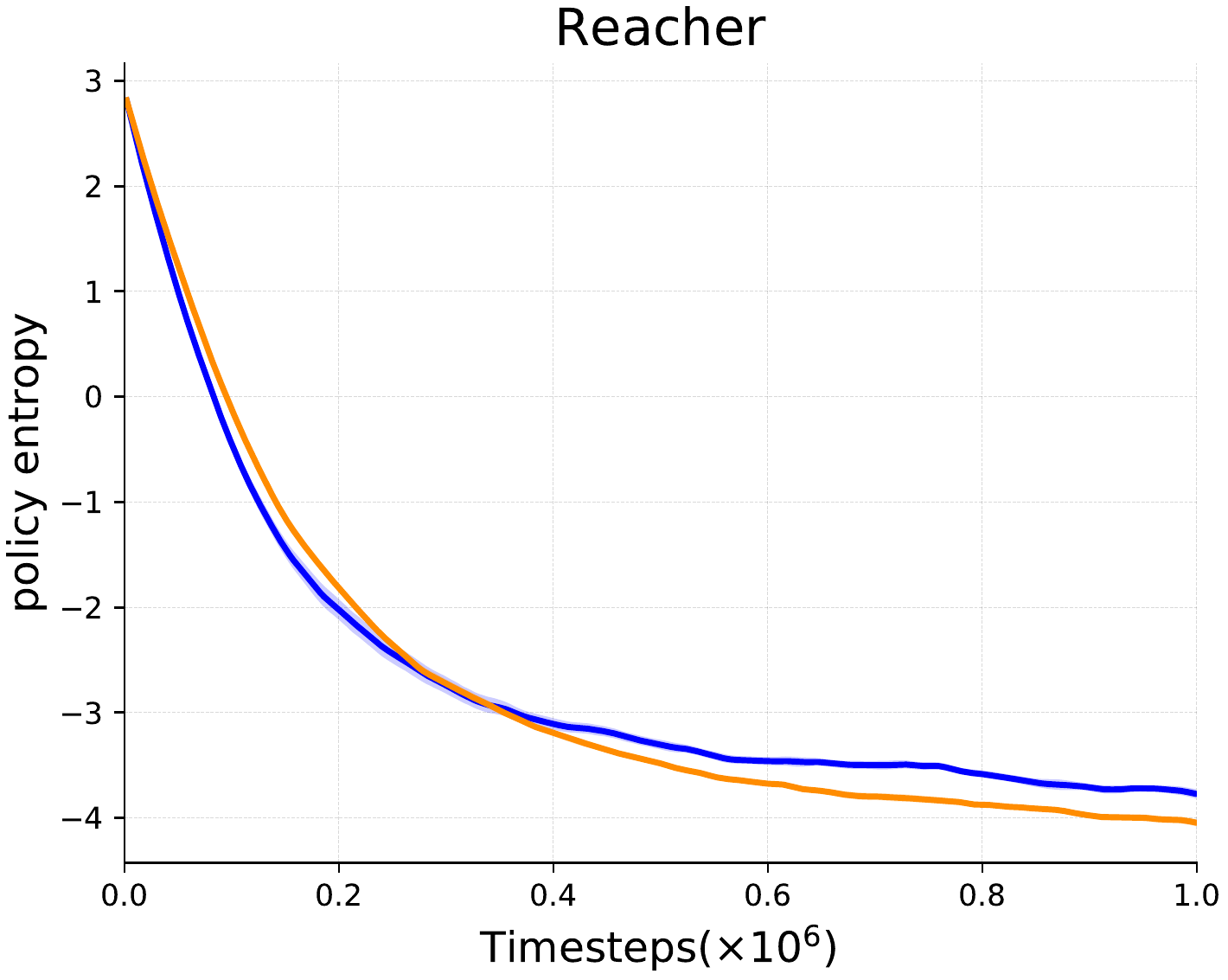}
		  		}
		  		\centerline{
			  		\includegraphics[width=0.5\linewidth]{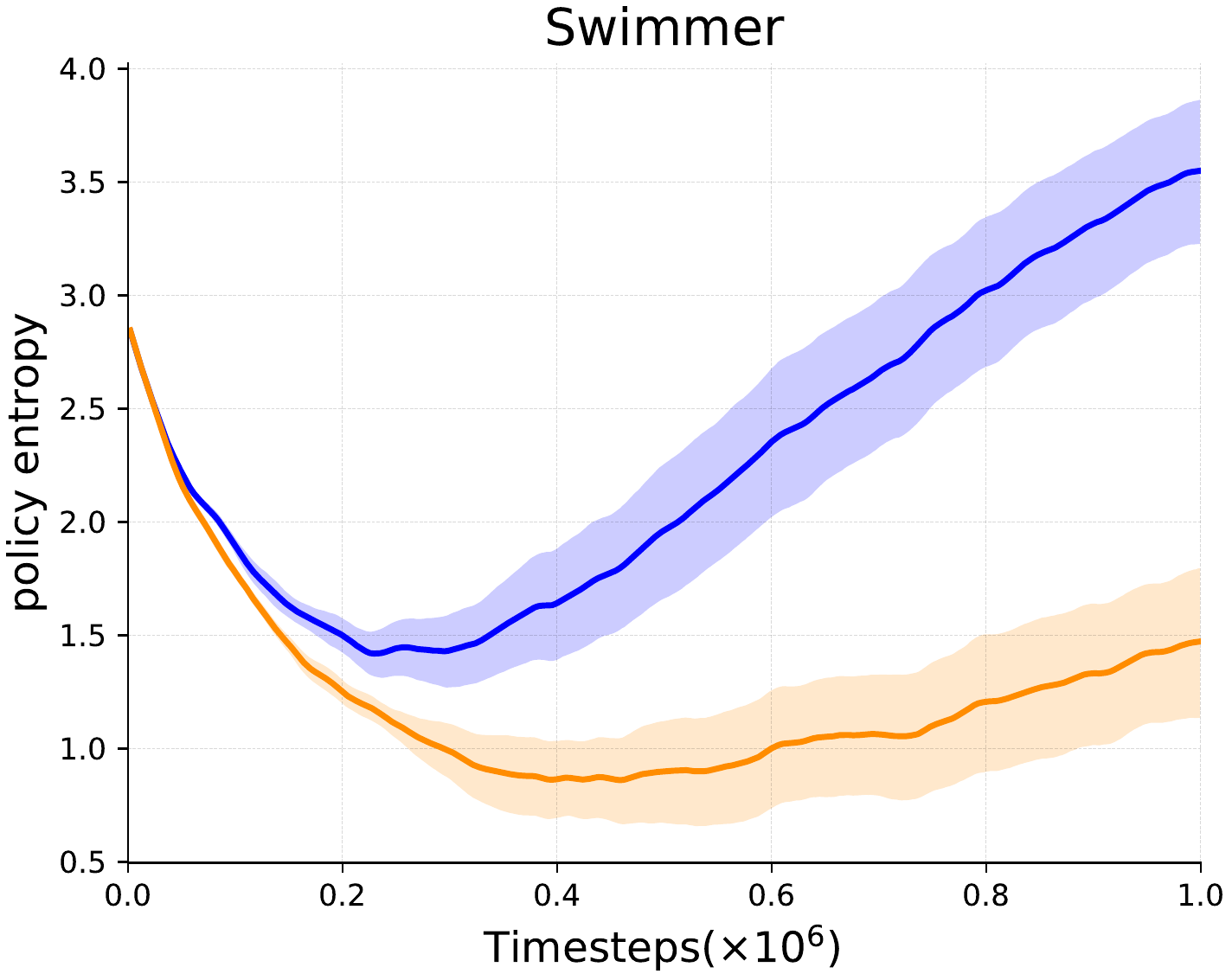}
			  		\includegraphics[width=0.5\linewidth]{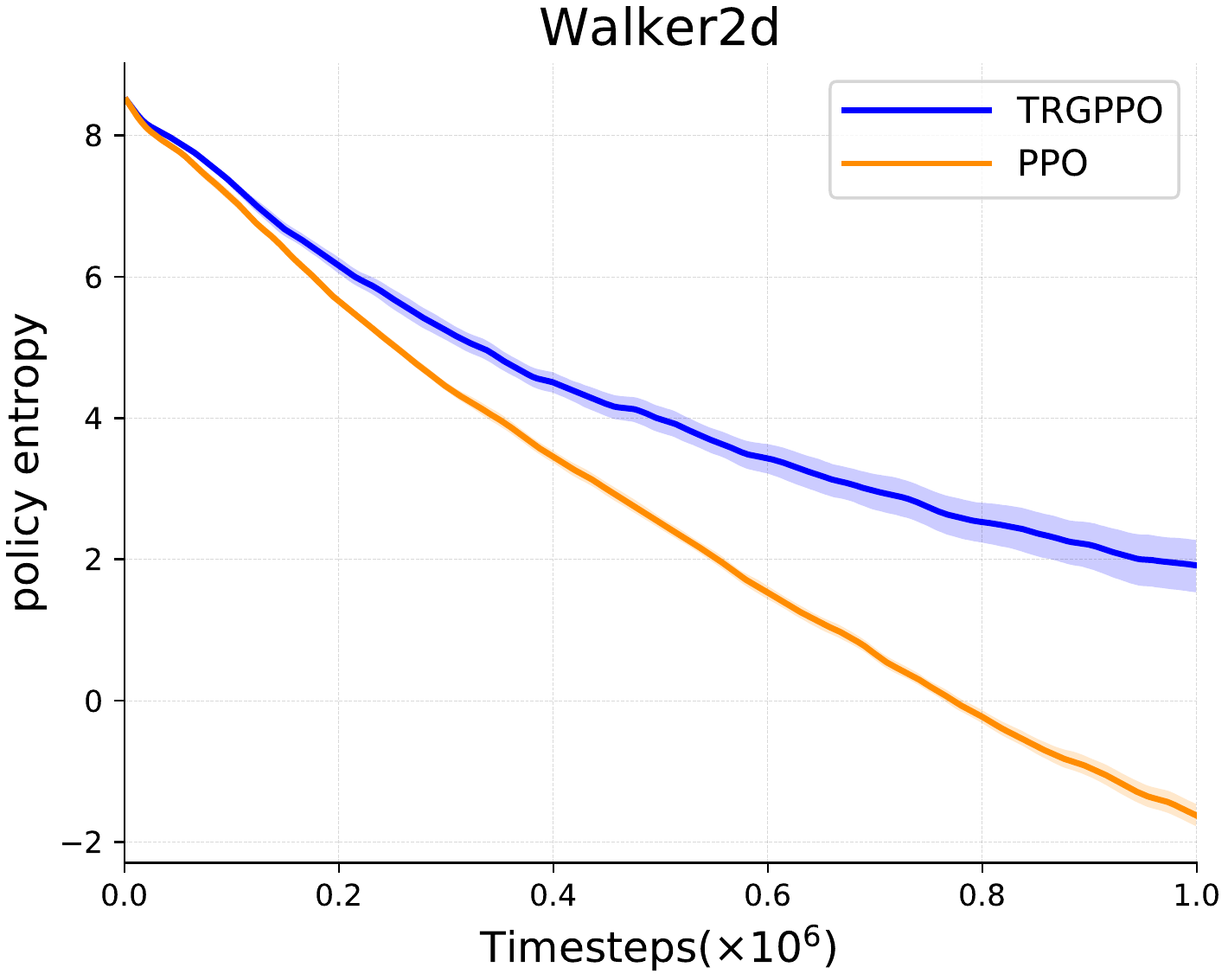}
		  		}
		  		\caption{Policy entropy}\label{fig_entropy}
	  		\end{subfigure}
			\begin{subfigure}[]{0.33\linewidth}
				\centerline{
		  			\includegraphics[width=0.5\linewidth]{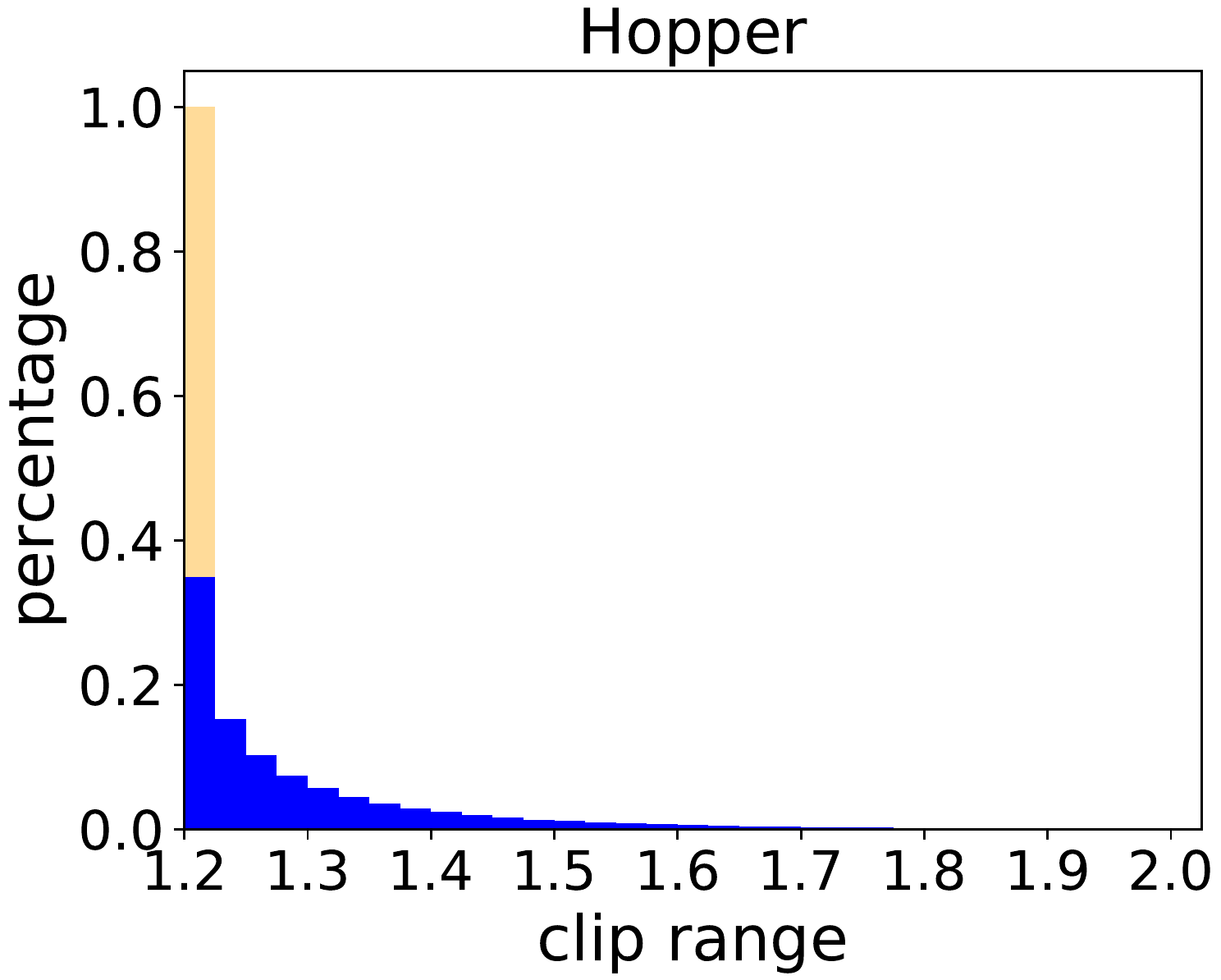}
		  			\includegraphics[width=0.5\linewidth]{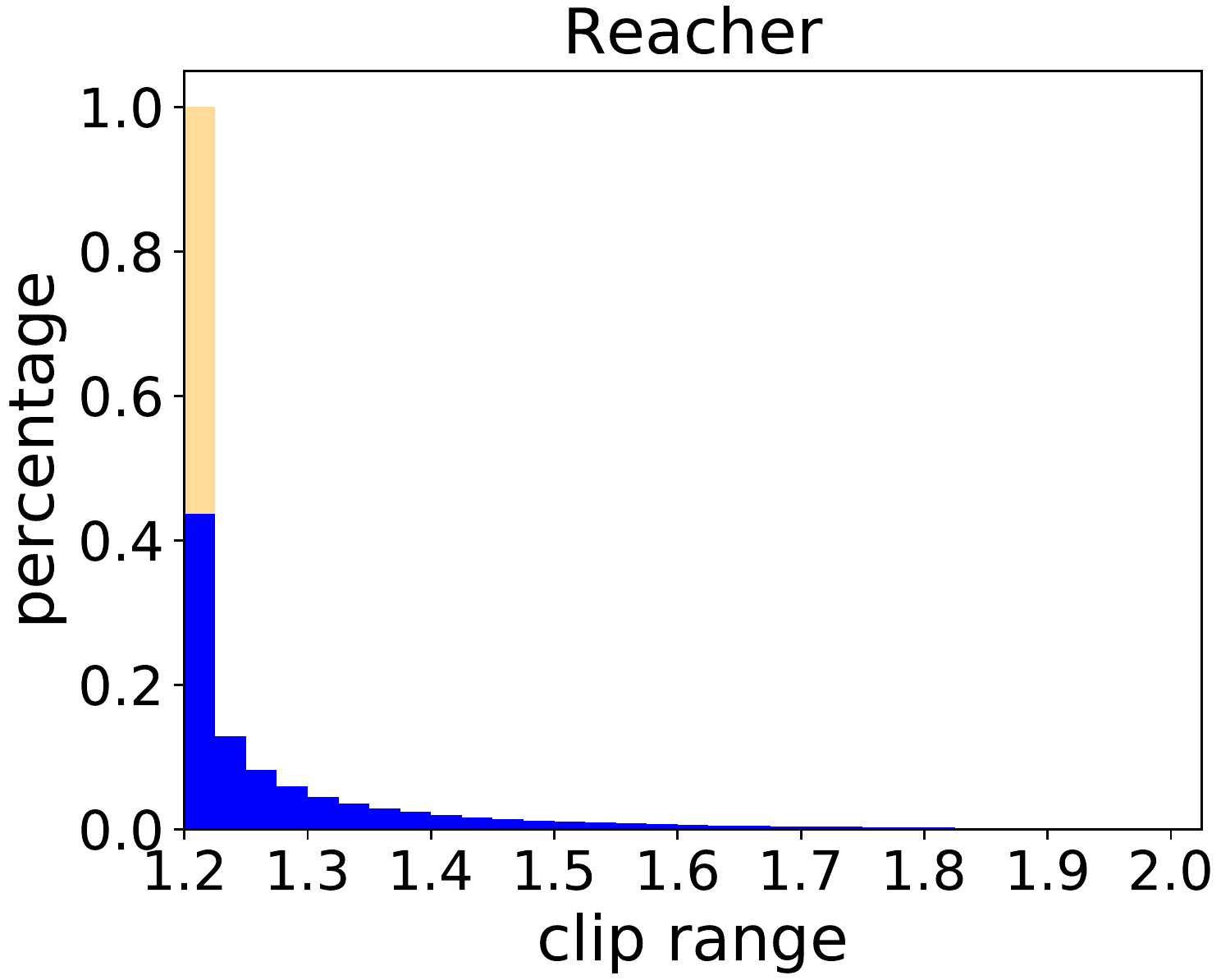}
		  		}
		  		\centerline{
			  		\includegraphics[width=0.5\linewidth]{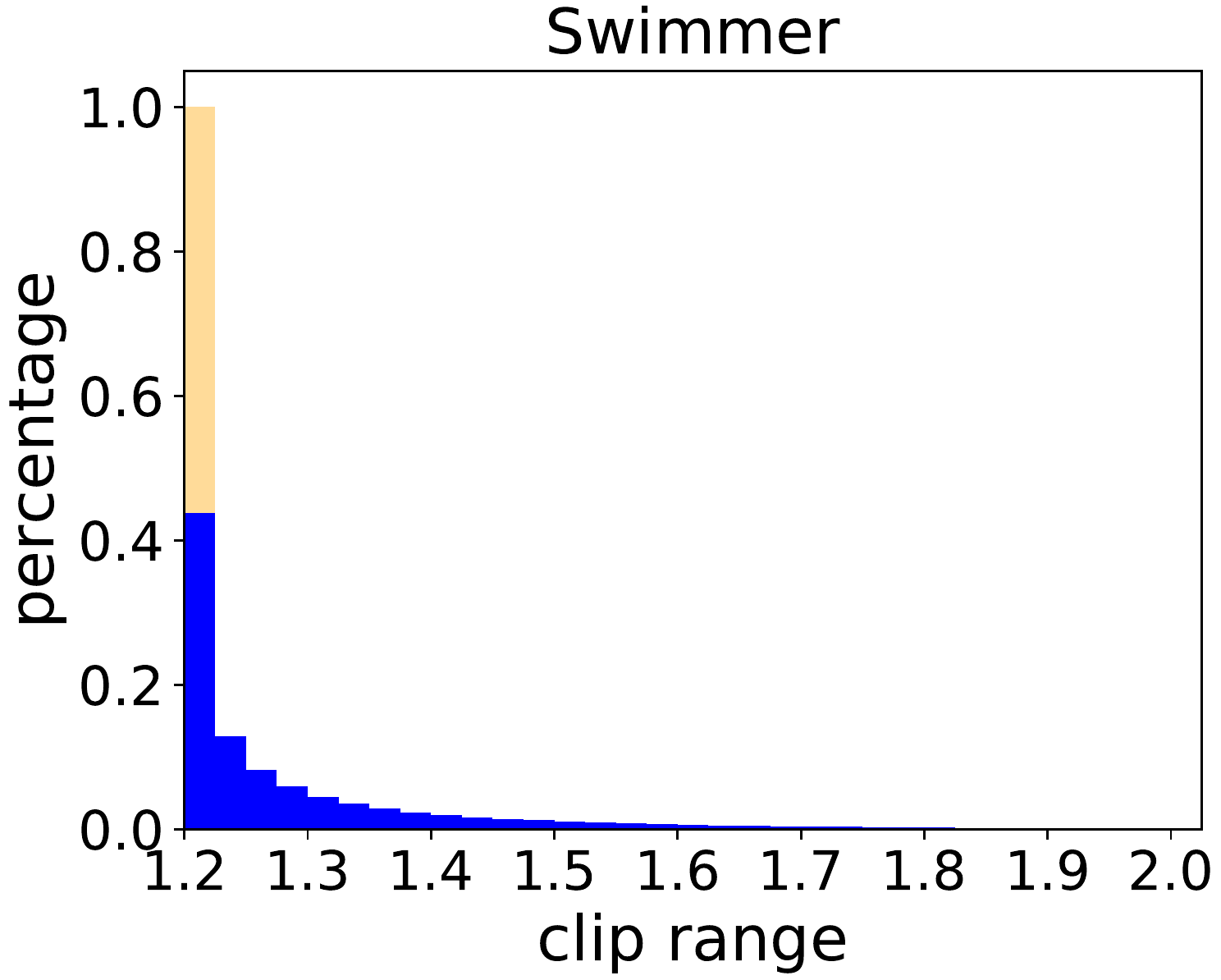}
			  		\includegraphics[width=0.5\linewidth]{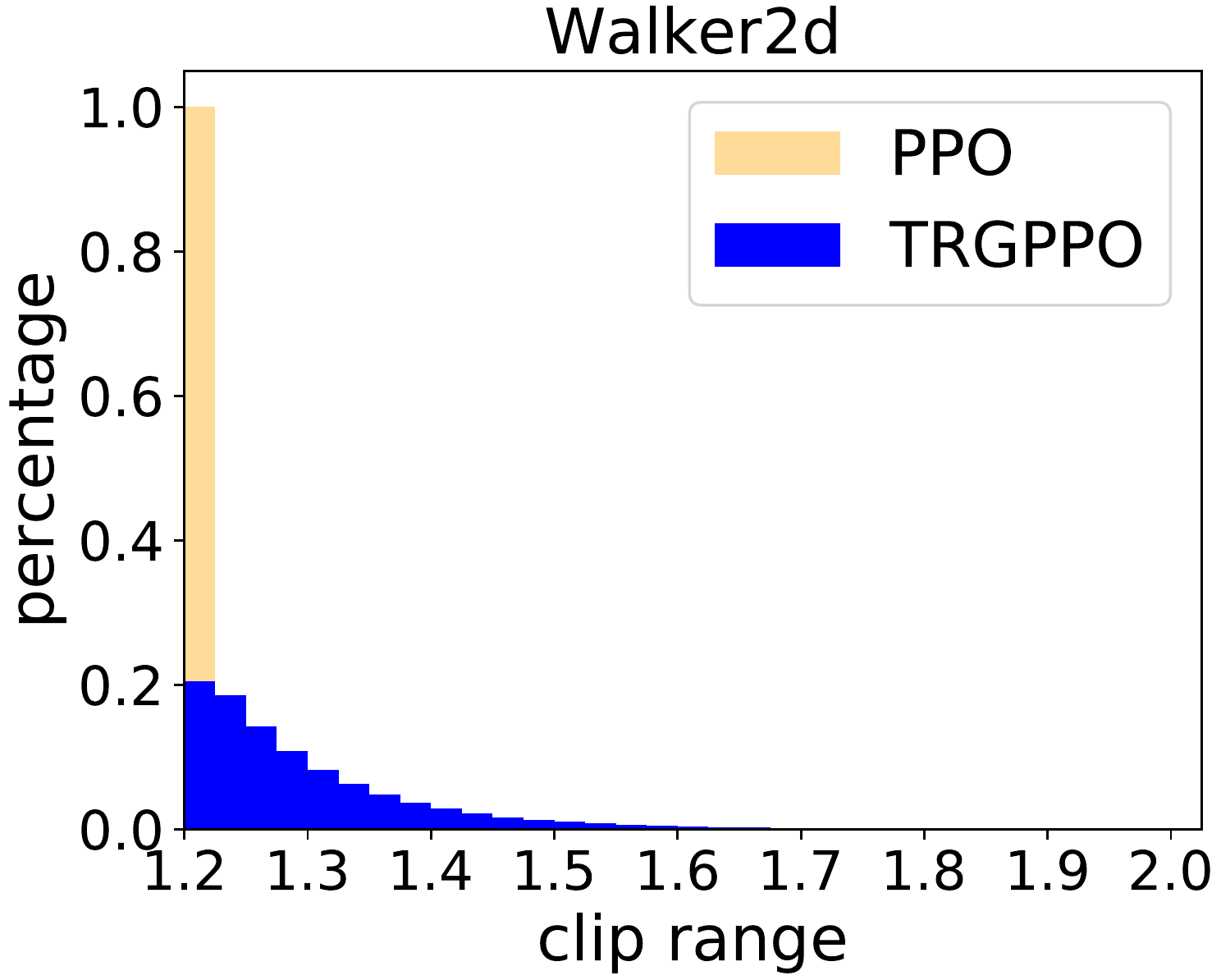}
		  		}
		  		\caption{Upper clipping range}\label{fig_clippingrange}
	  		\end{subfigure}
			\begin{subfigure}[]{0.33\linewidth}
				\centerline{
		  			\includegraphics[width=0.5\linewidth]{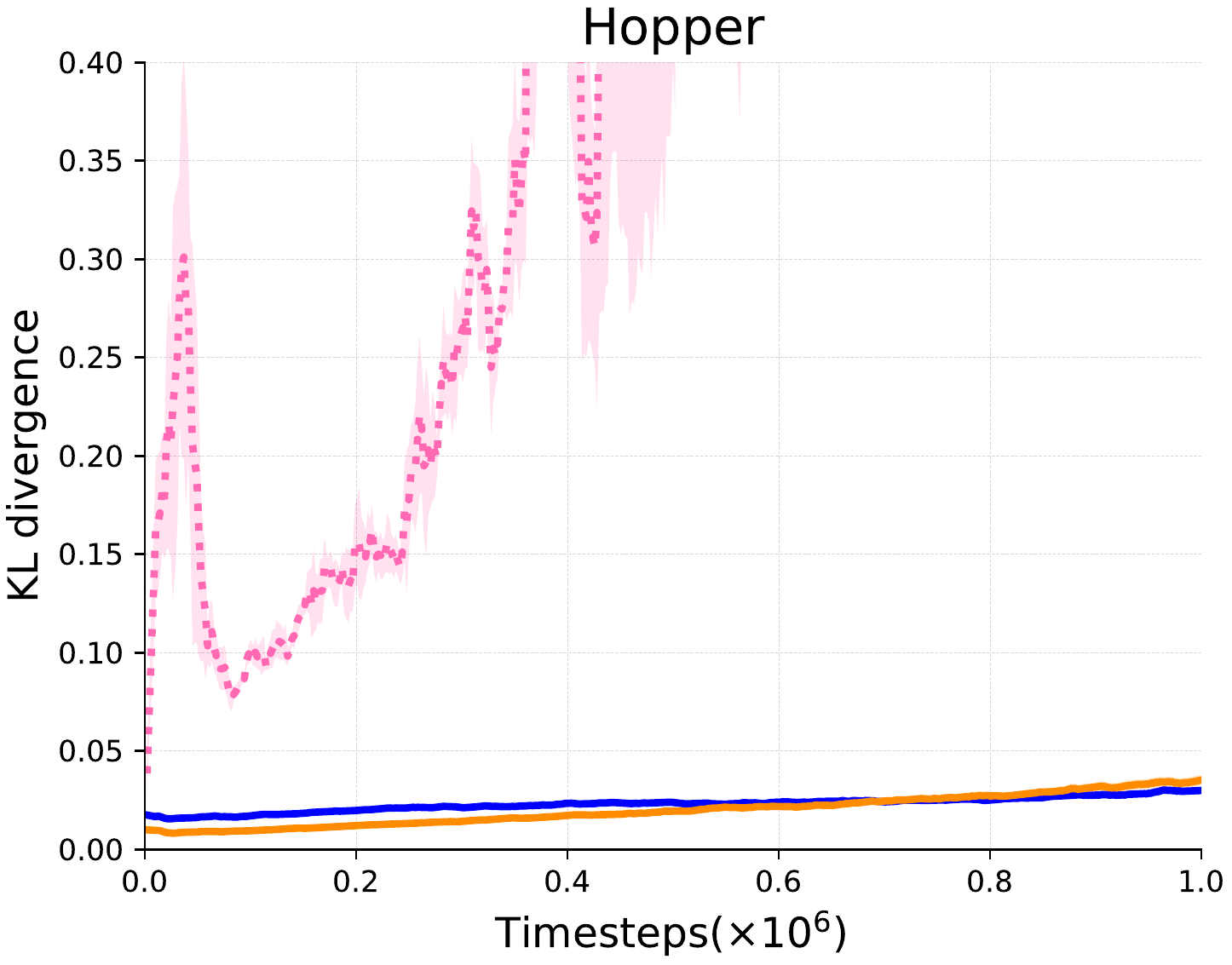}
		  			\includegraphics[width=0.5\linewidth]{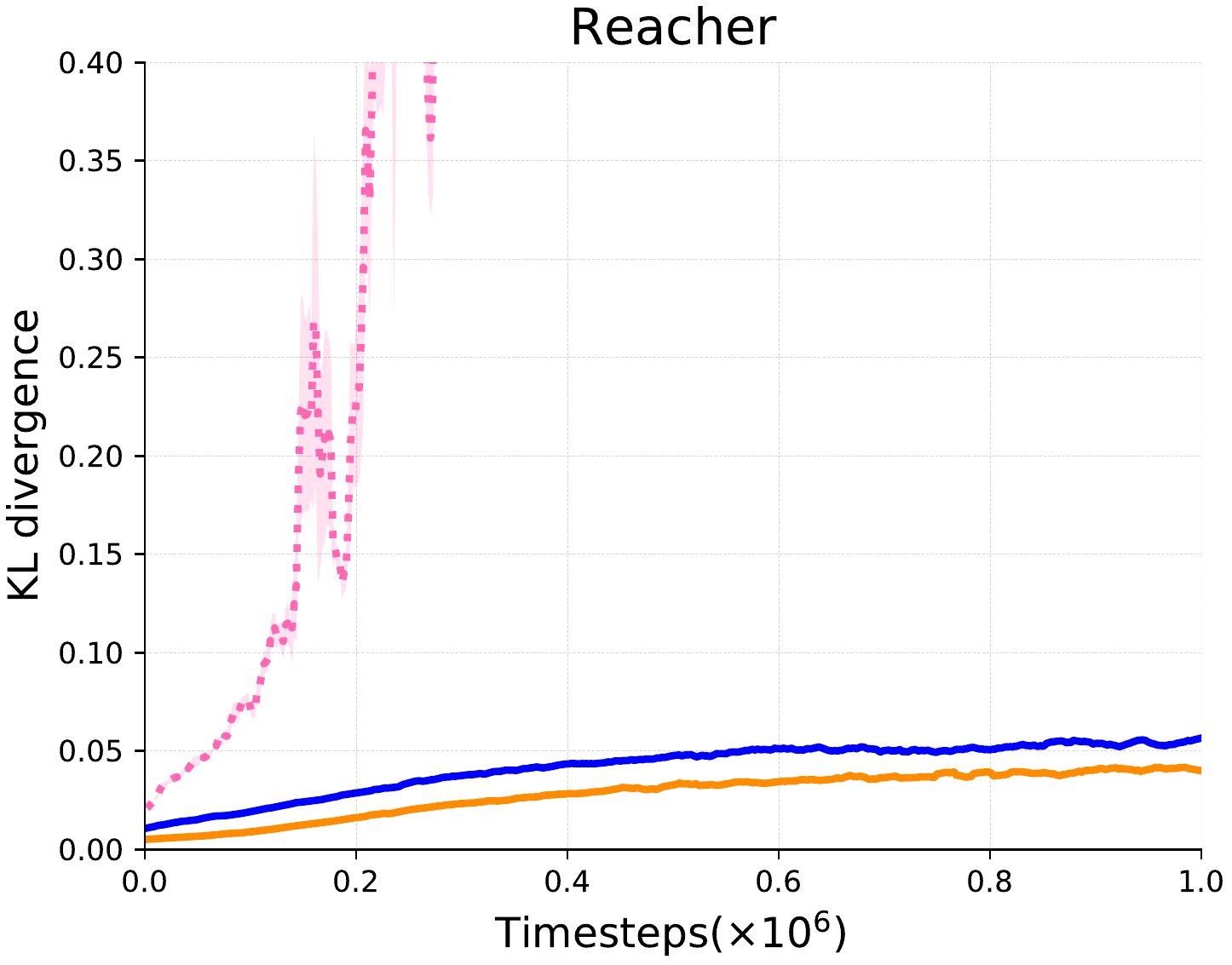}
		  		}
		  		\centerline{
			  		\includegraphics[width=0.5\linewidth]{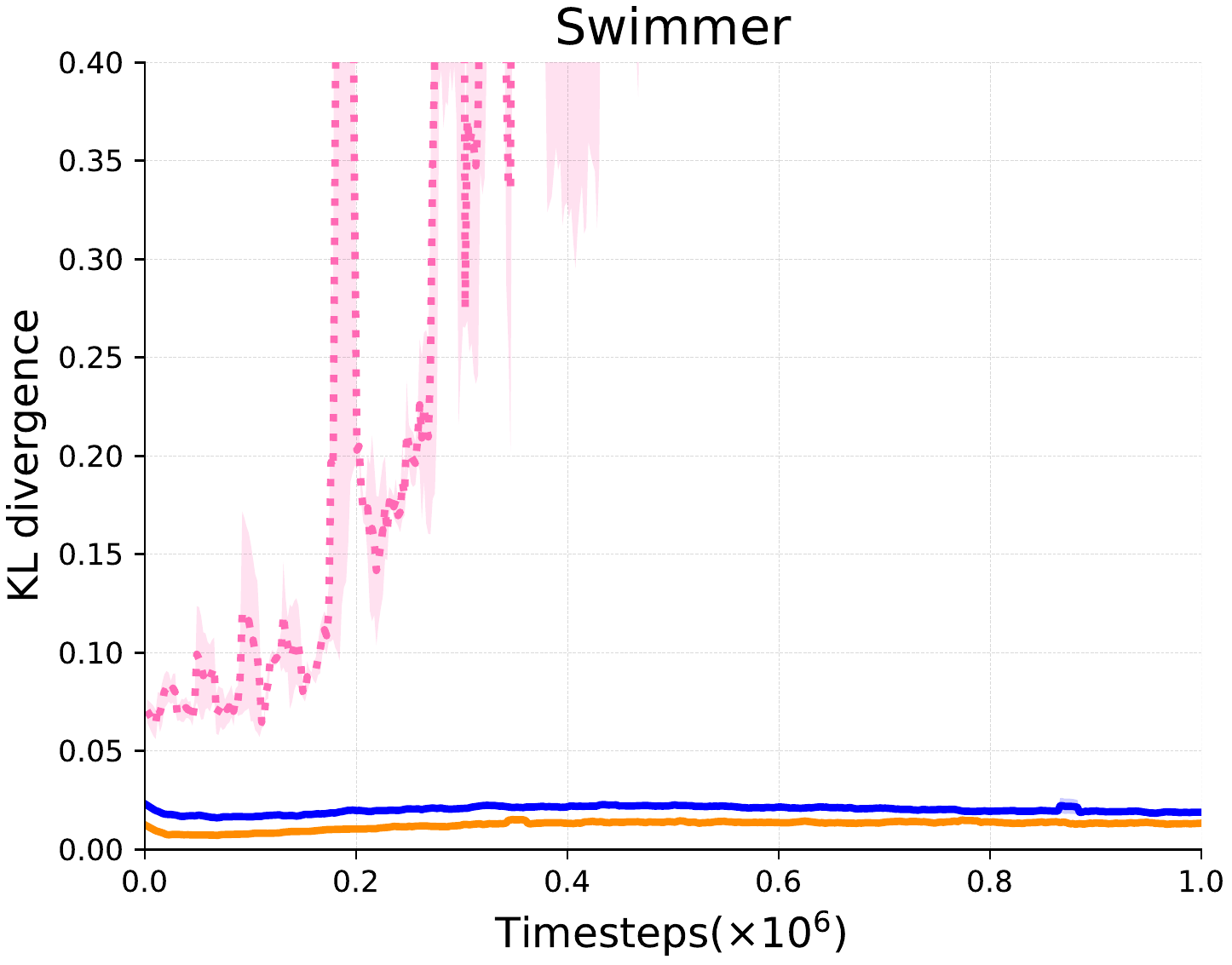}
			  		\includegraphics[width=0.5\linewidth]{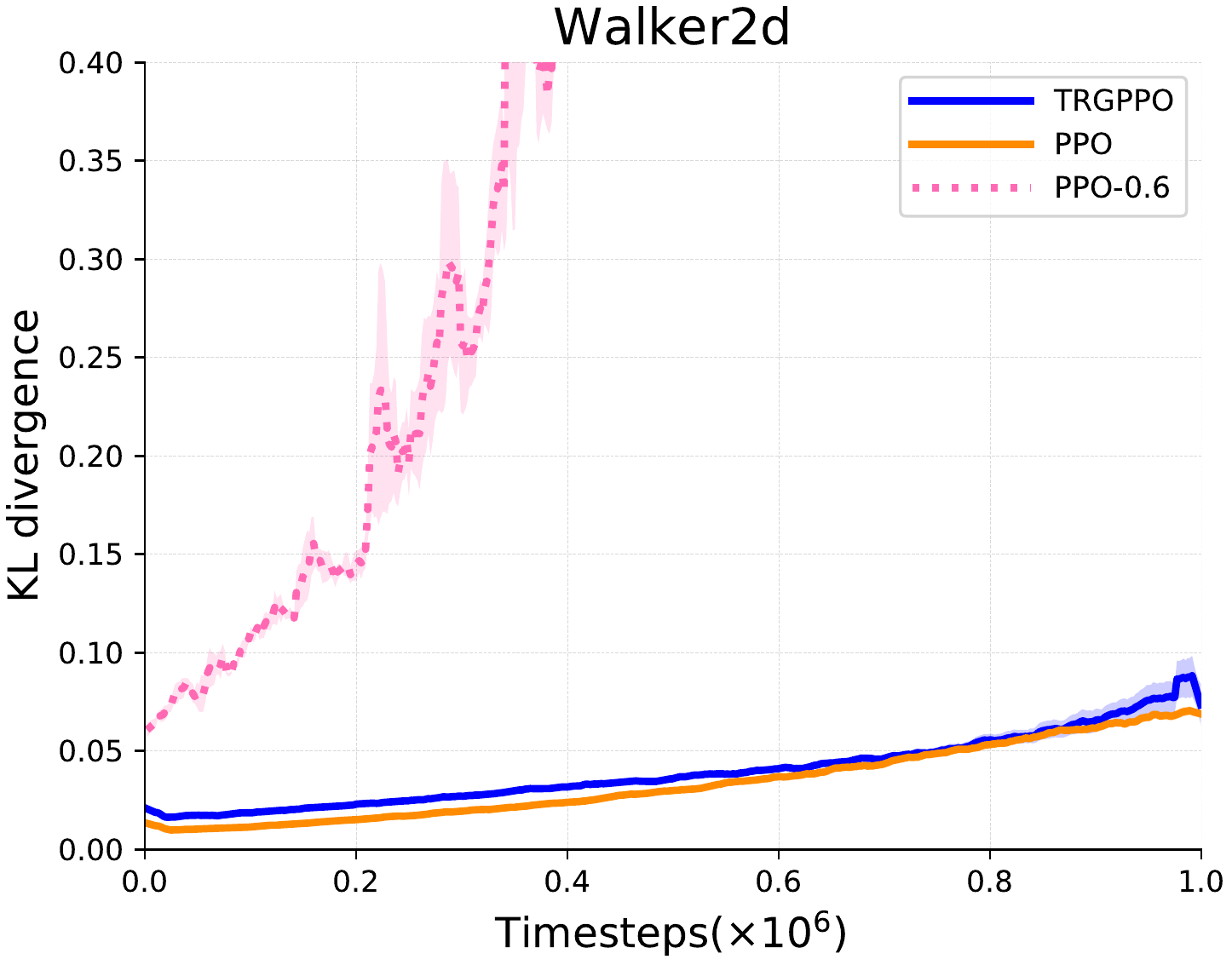}
		  		}
		  		\caption{KL divergence}\label{fig_kl}
	  		\end{subfigure}
	  	}
	%  		}
	%		    \subfigure{\includegraphics[width=0.3\linewidth]{figs/policy_entropy/Walker} }
	%		    \subfigure{\includegraphics[width=0.3\linewidth]{figs/policy_entropy/HalfCheetah}}
	%	\end{minipage}
	%	\subfigure[Policy Entropy]{\includegraphics[width=0.38\textwidth]{/media/d/yunwithhugo/entropy.pdf} \label{fig_entropy}}
	%	\subfigure[KL divergence]{\includegraphics[width=0.35\textwidth]{/media/d/yunwithhugo/maximumkl.pdf} \label{fig_kl}}
	%	\subfigure[Upper Clipping Ranges]{\includegraphics[width=0.24\textwidth]{/media/d/yunwithhugo/clippingrange.pdf} \label{fig_clippingrange}}
	%	}
		\setlength{\abovecaptionskip}{0mm}
		\caption{(a) shows the policy entropy during training process. (b) shows the statistics of the computed upper clipping ranges over all samples.
		(c) shows the KL divergence during the training process.
		}
		\end{figure}

	\textbf{Performance/Exploration:}
%	We evaluate the performance of policy and the policy entropy during the training process. The policy entropy could reflect the level of exploration. 
%	We additionally evaluate the performance of a deterministic version of the trained policy.
	Table \ref{tab_reward_hit} (b) lists the averaged rewards over last 40\% episodes during training process. 
	\pmethod/ outperforms the original PPO on almost all tasks except Reacher.
	\fig \ref{fig_entropy} shows the policy entropy during training process, the policy entropy of \pmethod/ is {obviously} higher than that of PPO. 
	These results implies that our \pmethod/ method could maintain a level of \ttt{entropy} learning and encourage the policy to explore more.

  	\textbf{The Clipping Ranges and Policy Divergence:}
  	\fig \ref{fig_clippingrange} shows the statistics of the upper clipping ranges of \pmethod/ and PPO. Most of the resulted adaptive clipping ranges of \pmethod/ are much larger that of PPO.
  	Nevertheless, our method has similar KL divergences with PPO (see \fig \ref{fig_kl}).
	However, the method of arbitrary enlarging clipping range (PPO-$0.6$) does not enjoy such property and fails on most of tasks.
%  	We evaluate the maximum KL divergence over all sampled states at each update iteration during the training process. 
  	
	\textbf{Training Time: }
%	We report the training wall-clock time of each algorithm.
	Within one million timesteps, the training wall-clock time for our \pmethod/ is 25 min; for PPO, 24 min; for SAC, 182 min (See Appendix \ref{app-sec_computaion_efficiency} for the detail of evaluation).
	\pmethod/ does not require much additional computation time than PPO does. 
	
	\textbf{Comparison with State-of-the-art Method: }
	\pmethod/ achieves higher reward than SAC on 5 tasks while is not as good as it on HalfCheetah. 
	And \pmethod/ is not sample efficient as SAC on HalfCheetah and Humanoid.
	This may due to that \pmethod/ is an on-policy algorithm while SAC is an off-policy one.
	However, \pmethod/ is much more computationally efficient (25 min vs. 182 min). 
	In addition, SAC tuned hyperparameters specifically for each task in the implementation of the original authors. 
	In contrast, our \pmethod/ uses the same hyperparameter across different tasks.%\pmethod/ requires relatively less effort on hyperparameter tuning as
%	 while SAC conducted a well-tuning on different tasks.
%	As can be seen, \pmethod/ requires much less training time than DVRL; this is mainly due to that the inference of \pmethod/ is made analytically while not requiring additional training to approximate the inference procedure.
%	Whereas compared to DRPPO and ADRPPO, \pmethod/ incorporates an additional imputation and filtering operation to perform robust inference, thus requires more computation.

	\section{Conclusion}
	In this paper, we improve the original PPO by an adaptive clipping mechanism with a trust region-guided criterion. Our \pmethod/ method improves PPO with more exploration and better sample efficiency and is competitive with several state-of-the-art methods,
	while maintains the stable learning property and simplicity of PPO.
%	Based on this observation, we proposed a novel policy optimization method, named \pmethod/, 
%	which adaptively adjusts the clipping range within the trust region.
%	We formally show that this method not only helps to make more exploration within the trust region but enjoys a better performance bound compared to the original PPO as well. Extensive experiments verify the advantage of the proposed method.

	To our knowledge, this is the first work to reveal the effect of the metric of policy constraint on the exploration behavior of the policy learning. 
	While recent works devoted to introducing inductive bias to guide the policy behavior, e.g., maximum entropy learning \citep{Ziebart2010ModelingIV, haarnoja2017reinforcement}, curiosity-driven method \citep{pathak2017curiosity}. In this sense, our adaptive clipping mechanism is a novel alternative approach to incorporate prior knowledge to achieve fast and stable policy learning. We hope it will inspire future work on investigating more well-defined policy metrics to guide efficient learning behavior.

	\section*{Acknowledgement}
	This work is partially supported by National Science Foundation of China (61976115,61672280, 61732006), AI+ Project of NUAA(56XZA18009), Postgraduate Research \& Practice Innovation Program of Jiangsu Province (KYCX19\_0195).
	We would also like to thank Yao Li, Weida Li, Xin Jin, as well as the anonymous reviewers, for offering thoughtful comments and helpful advice on earlier versions of this work.
	\bibliography{TRGPPO}

	\clearpage
	

\section*{Supplementary material}

# A Example: Continuous-Armed Bandit

In this section, we show how PPO and TRGPPO perform in continuous action space by a simple continuous-armed bandit problem. Fig. 2 (b) shows the plot of reward function (black dashed curve). Let $a_{\text{subopt}}$ denote any sub-optimal action which achieves second-highest reward; and let $a_{\text{opt}}$ denote the optimal one. The policy is a parametrized Gaussian (orange solid curve).

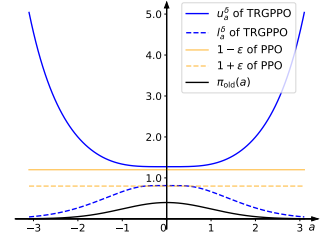

Fig. 1 shows the clipping ranges for different actions in continuous action space, where $dim(\mathcal{A}) = 1$. The old policy is $\pi_{\text{old}}(a) = \mathcal{N}(a|0, 1)$ (black curve). Note that the probability $\pi_{\text{old}}(a)$ goes smaller as $a$ is away from zero which the mode of the Gaussian distribution. As the figure shows, in continuous action space, our TRGPPO method sets larger clipping range for action which is less likely to be chosen, while PPO sets a constant clipping range under all actions.

Figure 1: Clipping Range of TRGPPO (blue curve) and PPO (orange curve) on different actions. For TRGPPO, $\delta = 0.05$, while for PPO, $\epsilon = 0.2$. The distribution of old policy is $\pi_{\text{old}}(a) = \mathcal{N}(a|0, 1)$.

Fig. 2 shows the training process of PPO on the continuous-armed bandit problem. As can be seen, the allowable improvement of $\pi(a_{\text{opt}})$ of PPO is quite limited. It will require quite a large number of steps for $\pi(a_{\text{opt}})$ to peak. On the other hand, the limited improvement at $\pi(a_{\text{opt}})$ will prevent the policy from allocating more probability at $a_{\text{opt}}$. In other words, it will explore less at $a_{\text{opt}}$. Whereas the allowable improvement at $a_{\text{subopt}}$ is relatively relax. This uneven restriction may result in a growing improvement of $\pi(a_{\text{subopt}})$ and a diminishment of $\pi(a_{\text{opt}})$. Consequently, the policy is trapped in local optima, as Fig. 2 (c) shows. Note that although we use Gaussian distribution as our policy in the example, these issues could also happen on multimodal distribution like Mixture Gaussian or heavy tailed distribution like Beta distribution.

Fig. 3 shows the training process of TRGPPO. When the policy enters into a locally optimal one, the corresponding feasible variation of $\pi(a_{\text{opt}})$ is close to that of $\pi(a_{\text{subopt}})$ in TRGPPO, as Fig. 3 (b) shows. Note that the advantage value at $a_{\text{opt}}$ is larger than that at $a_{\text{subopt}}$. This could result in a growing reinforcement of $\pi(a_{\text{opt}})$, which accordingly leads to a diminishment of $\pi(a_{\text{subopt}})$. In this way, the policy jumps out of the local optima and converges to the optimal policy, as Fig. 3 (c) shows.

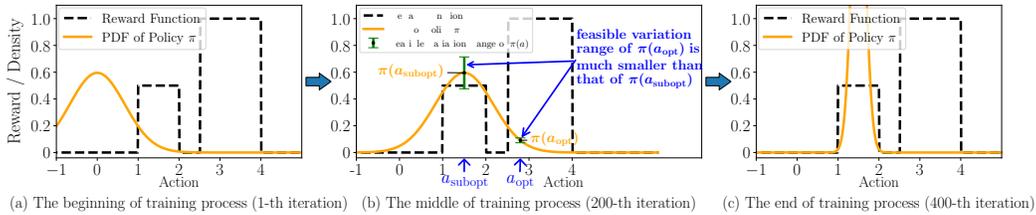

Figure 2: Training process of PPO. The orange solid curve plots the probability density function (PDF) of policy at the training process. The black dashed curve plots the reward function of the bandit problem.

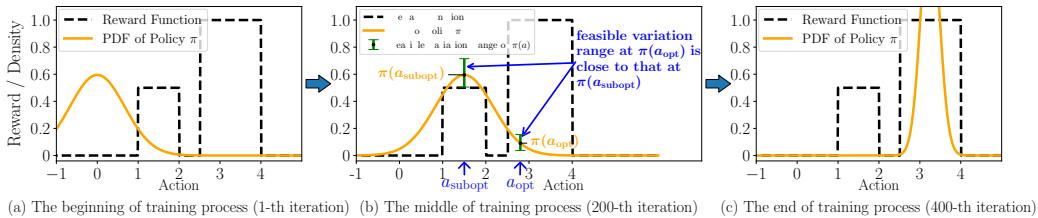

Figure 3: Training process of TRGPPO.

## B Computation of Adaptive Clipping Range

In this section, we detail the method for adaptive clipping range computation, which is formalized as the following problem.

$$l^{\delta}_{s_t,a_t} = \min_{\pi} \left\{ \frac{\pi(a_t|s_t)}{\pi_{\text{old}}(a_t|s_t)} : D^{s_t}_{KL}(\pi_{\text{old}}, \pi) \leq \delta \right\} \quad (1)$$

$$u^{\delta}_{s_t,a_t} = \max_{\pi} \left\{ \frac{\pi(a_t|s_t)}{\pi_{\text{old}}(a_t|s_t)} : D^{s_t}_{KL}(\pi_{\text{old}}, \pi) \leq \delta \right\} \quad (2)$$

To be abbreviated, we describe the approach for minimization case under discrete and continuous action space respectively, while that for the maximization case is similar.

### B.1 Discrete Action Space

For discrete action space tasks, it is standard to use a DNN with softmax output layer to represent the policy, i.e., $\pi(a|s) = [f^p_\theta(s)]_a$, where $f^p_\theta(s)$ is the parameter of categorical distribution on state $s$ and the subscript $a$ denote the $a$-th entry of the vector. The optimal clipping range should be computed should be independent of special parametrization of $f^p_\theta$. Thus the problem is formalized as an optimization problem of two Categorical distributions. Given $s_t$ and $a_t$, let $p' = f^p_{\theta'}(s_t)$, where $\theta'$ is the parameter of old policy $\pi_{\text{old}}$, the computation of $l^{\delta}_{s_t,a_t}$ in (1) is formalized as the following optimization problem:

$$\min_{p} \frac{p_{a_t}}{p'_{a_t}}$$

$$\text{s.t.} \sum_{a \in \mathcal{A}} p'_a \log \frac{p'_a}{p_a} \leq \delta, \quad \sum_{a \in \mathcal{A}} p_a = 1, \quad (3)$$

$$0 \leq p_a \leq 1 \text{ for } a \in \mathcal{A}$$

While the optimization problem for $u^{\delta}_{s_t,a_t}$ is the maximization case of problem (3). Let $\lambda$ and $\nu$ be the Lagrangian multipliers.

$$\begin{cases} \lambda \left( -\frac{p'_a}{p_a} \right) + \nu = 0, \quad \text{for } a \neq a_t & (4a) \\ \frac{1}{p'_{a_t}} + \lambda \left( -\frac{p'_{a_t}}{p_{a_t}} \right) + \nu = 0 & (4b) \\ \lambda \left( \sum_{a \in \mathcal{A}} p'_a \log \frac{p'_a}{p_a} - \delta \right) = 0 & (4c) \\ \sum_{a} p_a = 1 & (4d) \end{cases}$$

By (a)(b), we have $\lambda \neq 0$, since if $\lambda = 0$ then $\nu = 0$ (by (a)), which contradicts (b). Second, by (c) and $\lambda \neq 0$, we have $\sum_{a \in \mathcal{A}} p'_a \log(p'_a/p_a) = \delta$. Third, taking (a) into (d), we have $p'_a/p_a = \nu/\lambda = (1 - p'_{a_t})/(1 - p_{a_t})$ for $a \neq a_t$. Then, taking this equation into $\sum_{a \in \mathcal{A}} p'_a \log(p'_a/p_a) = \delta$, this problem is transformed into solving the following equation w.r.t. $p_{a_t}$.

$$(1 - p'_{a_t}) \log \frac{1 - p'_{a_t}}{1 - p_{a_t}} - p'_{a_t} \log \frac{p_{a_t}}{p'_{a_t}} = \delta \quad (5)$$

In fact, there are two groups of solution for (5), where $p_{a_t}/p'_{a_t} < 1$ is the one for the minimization case, while $p_{a_t}/p'_{a_t} > 1$ is the one for the maximization case.

### B.2 Continuous Action Space

For continuous action space tasks, it is standard to represent the stochastic policy by a parameterized conditional Gaussian distribution[4, 2], i.e., $\pi(a|s) = \mathcal{N}(a|f^\mu_\theta(s), f^\Sigma_\theta(s))$, where $f^\mu_\theta$ and $f^\Sigma_\theta$ are two

DNNs which output the mean vector and covariance matrix. Note that the optimal clipping range should be computed independent of special parametrization of $f_\theta^\mu$ and $f_\theta^\Sigma$. Thus the problem (1) is formalized as an optimization problem of two Gaussian distributions. Given $(s_t, a_t)$, let $\mu' = f_{\theta'}^\mu(s_t)$ and $\Sigma' = f_{\theta'}^\Sigma(s_t)$, where $\theta'$ is the parameter of old policy $\pi_{\text{old}}$, the computation of $l_{s_t,a_t}^\delta$ in (1) is formalized as the following optimization problem:

$$\min_{\mu,\Sigma} \frac{1}{2} \left( -\log|\Sigma| - (\mu - a_t)^\top \Sigma^{-1} (\mu - a_t) + \log|\Sigma'| + (\mu' - a_t)^\top \Sigma'^{-1} (\mu' - a_t) \right)$$
$$\text{s.t.} \frac{1}{2} \left( \log\left|\Sigma'\Sigma^{-1}\right| + tr\left\{\Sigma'^{-1}\Sigma\right\} + (\mu - \mu')^\top \Sigma'^{-1}(\mu - \mu') - D \right) - \delta \leq 0 \quad (6)$$

where $\mu \in \mathbb{R}^D$, $\Sigma \in \mathbb{R}^{D \times D}$ is a positive semi-definite matrix, $D = dim(\mathcal{A})$ is the dimension of action space. The objective function is log of the ratio $\pi$. The covariance matrix could be decomposed by $\Sigma' = \bar{\Sigma}'\bar{\Sigma}'^\top$, and we introduce a rotation matrix $R \in \mathbb{R}^{D \times D}$ (which has $R^\top R = R^{-1} R = I$).

Second, by replacing $\mu$ with $\bar{\Sigma}' R \mu + \mu'$, and $\Sigma$ with $\bar{\Sigma}' R \Sigma R^\top \bar{\Sigma}'^\top$, we could transform the problem to

$$\min_{\mu,\Sigma} \frac{1}{2} \left( -\log|\Sigma| - (\mu - \bar{a}_t)^\top \Sigma^{-1} (\mu - \bar{a}_t) + \bar{a}_t^\top \bar{a}_t \right)$$
$$\text{s.t.} \frac{1}{2} \left( \log\left|\Sigma^{-1}\right| + tr\left\{\Sigma\right\} + \mu^\top \mu - D \right) - \delta \leq 0 \quad (7)$$

where $\bar{a}_t = R^\top \bar{\Sigma}'^{-1} (\mu' - a_t)$.

Next, we constrain the covariance matrix $\Sigma$ to be diagonal. The final result is sub-optima compared to the original problem. However, we don't require accurate clipping bound when optimizing policy. Another reason is that in practice the diagonal Gaussian policy is widely used in RL realizations. Then (7) is equivalent to the following problem.

$$\min_{\mu,\sigma} \frac{1}{2} \sum_{d=1}^{D} \left( -\log \sigma_d - (\mu_d - \bar{a}_{t,d})^2 \sigma_d^{-1} + \bar{a}_{t,d}^2 \right)$$
$$\text{s.t.} \frac{1}{2} \sum_{d=1}^{D} \left[ -\log \sigma_d + \sigma_d + \mu_d^2 - D \right] - \delta \leq 0 \quad (8)$$

where $\mu \in \mathbb{R}^D$, $\sigma \in \mathbb{R}^{+D}$. We choose appropriate $R$ to make $\bar{a}_{t,d} = \dot{a}_t = \left\|(\mu' - a_t)\bar{\Sigma}'^{-1}\right\|/\sqrt{D}$ for $d = 1, 2, \cdots, D$, which means that all entries of $\bar{a}_t$ are equal. Let $\lambda$ be the Lagrangian multiplier, by appling the KKD condition,

$$\begin{cases} -\frac{(\mu_d - \dot{a}_t)}{\sigma_d} + \lambda \mu_d = 0 & d = 1, \cdots, D \\ -\frac{1}{\sigma_d} + \frac{(\mu_d - \dot{a}_t)^2}{\sigma_d^2} + \lambda \left[-\frac{1}{\sigma_d} + 1\right] = 0 & d = 1, \cdots, D \\ \lambda \left( \frac{1}{2} \sum_{d=1}^{D} \left[-\log \sigma_d + \sigma_d + \mu_d^2 - D\right] - \delta \right) = 0 \\ \lambda \geq 0 \end{cases} \quad (9)$$

By the equations above, we could easily know that $\mu_d$ and $\sigma_d$ are equal for all $d$. Thus the problem could collapse to the following problem,

$$\min_{\mu,\sigma} \frac{1}{2} D \left( -\log \sigma - (\mu - \dot{a}_t)^2 \sigma^{-1} + (0 - \dot{a}_t)^2 \right)$$
$$\text{s.t.} \frac{1}{2} \left( -\log \sigma + \sigma + \mu^2 - 1 \right) - D^{-1} \delta \leq 0 \quad (10)$$

Remember that $\dot{a}_t = \left\|(\mu' - a_t)\bar{\Sigma}'^{-1}\right\|/\sqrt{D}$, $\mu \in \mathbb{R}$, $\sigma \in \mathbb{R}^+$. Until now, the original $D$-dimensional optimization problem is transformed to a 1-dimensional optimization problem. By the KKT conditions above, we could obtain the following equations w.r.t. $\mu$, $\sigma$ and $\lambda$.

When $\dot{a}_t \neq 0$, the problem is transformed into solving the following equations w.r.t. $\mu, \sigma, \lambda$:

$$\begin{cases} -\log \frac{(\mu-\dot{a}_t)(\mu^2-\dot{a}_t\mu-1)}{\dot{a}_t} + \frac{(\mu-\dot{a}_t)(\mu^2-\dot{a}_t\mu-1)}{\dot{a}_t} + \mu^2 - 1 - 2\delta D^{-1} = 0 \\ \sigma = (\mu-\dot{a}_t)(\mu^2 - \dot{a}_t\mu - 1)/\dot{a}_t \\ \lambda = \frac{\mu-\dot{a}_t}{\sigma\mu} \end{cases} \quad (11)$$

When $\dot{a}_t = 0$, the problem is transformed into solving the following equations w.r.t. $\mu, \sigma, \lambda$:

$$\begin{cases} \mu = 0 \\ -\log \sigma + \sigma - 1 - 2\delta/D = 0 \\ \lambda = \frac{-\sigma+\dot{a}_t^2}{-\sigma+\sigma^2} \end{cases} \quad (12)$$

There are two groups of solution for both (11) and (12), where $\lambda > 0$ is the one for the minimization case, while $\lambda < 0$ is the one for the maximization case.

### B.3 Computation Acceleration

Note the solutions in (5) only depend on the one-dimensional constant $p'_a$ and $\delta$, while (11) and (12) only depend on one-dimensional constant $\dot{a}$ and $\delta/D$. We use MINPACK's HYBRD and HYBRJ routines [3] as the solver. To accelerate this computation procedure, we propose two additional approach. One is to train a DNN which input $\pi_{\text{old}}(a|s)$ and $\delta$ and approximately output the solutions, which serve as the initial solutions for the solver. The other is to discretize the space of the input and save all the solutions in advance. The experimental results in our main content are conducted with the discretization version.

Table 2 shows the wall-clock time required by variants of TRGPPO and PPO to finish benchmark tasks in a modern CPU. With our proposed acceleration tricks, the optimization time of calculating clipping range can be reduced significantly. The result is obtained with the same experiment setup as previous experiments. The experiments are applied on a computer with an Intel i5-7500 CPU, 16GB of memory and a GeForce GTX 1060 GPU.

Table 1: Input and output of the DNN for solving problems. For discrete action space, we sample 1000 $p'_a$. For continuous action space, we sample 1000 $\dot{a}$ and 1000 $\delta/D$ (note we take $\delta/D$ as an entity). We solve these problems and obtain the corresponding solutions, and these data are used to train our DNN.

|  | Input | Output |
|---|---|---|
| **Discrete Action Space** | $p'_a \sim U(0,1)$ | $p_a$ |
| **Continuous Action Space** | $\dot{a} \sim U(-5,5); D^{-1}\delta \sim U[0.0002, 0.01]$ | $\mu, \sigma$ |

Table 2: Comparison of computation cost for TRGPPO with different acceleration tricks.

|  | PPO | TRGPPO (discretization) | TRGPPO (solver) | TRGPPO (DNN) |
|---|---|---|---|---|
| **Mujoco($10^6$ timesteps)** | 24 min | 25 min | 52 min | 29 min |
| **Atari($10^7$ timesteps)** | 195 min | 198 min | 243 min | 213 min |

### B.4 Adaptively Setting $\delta$ with $\epsilon$

We detail method about how to adaptively set $\delta$ by $\epsilon$. Our goal is to make TRGPPO has theoretical maximum KL divergence over all sampled states.

For discrete action space, let $p^+ = \max_{t:A_t>0} \pi_{\text{old}}(a_t|s_t)$, $p^- = \max_{t:A_t<0} \pi_{\text{old}}(a_t|s_t)$. By eq. (5), we set $\delta$ by

$$\delta = max(\delta^+, \delta^-) \quad (13a)$$

$$\delta^+ = (1-p^+)\log \frac{1-p^+}{1-p^+(1+\epsilon)} - p^+ \log(1+\epsilon) \quad (13b)$$

$$\delta^- = (1-p^-)\log \frac{1-p^-}{1-p^-(1-\epsilon)} - p^- \log(1-\epsilon) \quad (13c)$$

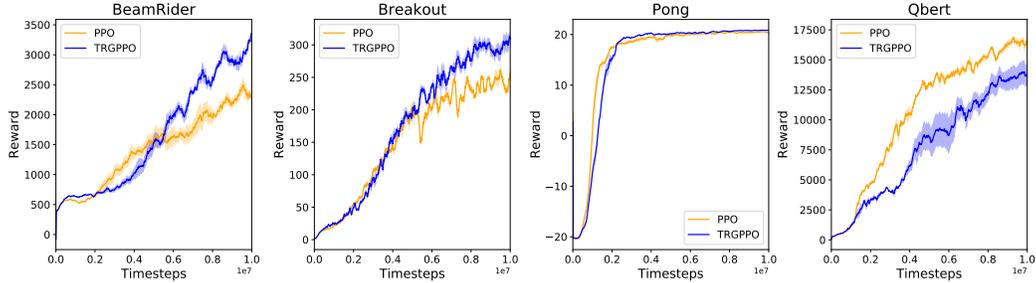

Figure 4: Episode rewards achieved by algorithm during the training process averaged over 4 random seeds. TRGPPO (blue line) achieves better performance than PPO (orange line).

For continuous action space, for PPO, theoretically, it always achieves maximum KL divergence at the sampled action which is the mode of the Gaussian. Our idea is to make the optimal clipping range at the mode of distribution equals the clipping range of PPO. By problem eq. (10) and (12), we set $\delta$ by

$$\delta = max(\delta^+, \delta^-) \tag{14a}$$

$$\delta^+ = \log(1+\epsilon) + \frac{1}{2} D \exp(\frac{-2\log(1+\epsilon)}{D}) - \frac{1}{2} D \tag{14b}$$

$$\delta^- = \log(1-\epsilon) + \frac{1}{2} D \exp(\frac{-2\log(1-\epsilon)}{D}) - \frac{1}{2} D \tag{14c}$$

## C   Additional Experiment

To evaluate the proposed TRGPPO on discrete tasks, we use Atari games as a testing environment, so the policies are learned with raw images. We present results on several atari games in Fig. 4, the blue and orange curves visualize the results using TRGPPO and PPO. We set $\delta = 0.001$ for all tasks.

Both TRGPPO and PPO adopt exactly same implementations and hyperparameters given in [1] for discrete tasks except that clipping range of TRGPPO is computed adaptively according to given $\delta$, and the policy entropy coefficient is 0 but not 0.01 used in PPO. This is because our TRGPPO has better *exploration* property than PPO, so it does not need to add extra entropy regularization.

## D   Implementation Details

Table 3: Hyperparameters for PPO and TRGPPO on Mujoco tasks.

| Hyperparameter | Value |
|---|---|
| learning rate | $3 \times 10^{-4}$ |
| number of parallel environments | 64 (Humanoid) <br> 2 (Other tasks) |
| timesteps per epoch | 1024 |
| initial logstd of policy | -1.34 (HalfCheetah,Humanoid) <br> 0 (Other tasks) |
| policy | Gaussian |
| $\lambda$ (GAE) | 0.95 |
| $\epsilon$ (clipping range) | 0.2 |

Table 4: Hyperparameters for PPO and TRGPPO on Atari tasks.

| Hyperparameter | Value |
|---|---|
| learning rate | $2.5 \times 10^{-4}$ |
| number of parallel environments | 8 |
| timesteps per epoch | 128 |
| policy | CNN |
| $\lambda$ (GAE) | 0.95 |
| $\epsilon$ (clipping range) | LinearAnneal(0.1,0) |

# E  Theorem Proof

In this section, we will give theorem proofs. To make it easier to read, we will mention the related notations again.

## E.1  Theorems in Section 4

**Lemma 1.** $\Delta_{\pi_0,t} \triangleq \mathbb{E}_{\pi_t}\left[\|\pi_t - \pi^*\|_\infty | \pi_0\right] = 1 - \mathbb{E}_{\pi_t}\left[\pi_t(a_{\text{opt}})|\pi_0\right]$.

**Proof:** For any $a \neq a_{\text{opt}}$, then $|\pi_t(a) - \pi^*(a)| = |\pi_t(a)| \leq |1 - \pi_t(a_{\text{opt}})|$.

Thus we have $\|\pi_t - \pi^*\|_\infty = \max_{a \in \mathcal{A}} \|\pi_t(a) - \pi^*(a)\| = 1 - \pi_t(a_{\text{opt}})$.

**Lemma 2.** $\mathbb{E}_{\pi_{t+1}}\left[\pi_{t+1}(a)|\pi_0\right] = \mathbb{E}_{\pi_t}\left[\mathbb{E}_{\pi_{t+1}}\left[\pi_{t+1}(a)|\pi_t\right]|\pi_0\right]$.

**Theorem 2.** *Given initial policy $\pi_0$, if $\pi_0^2(a_{\text{opt}}) \cdot |\mathcal{A}| < \sum_{a_{\text{subopt}} \in \mathcal{A}_{\text{subopt}}} \pi_0^2(a_{\text{subopt}}) - \sum_{a^- \in \mathcal{A}^-} \pi_0^2(a^-)$, then we have*

*(i)* $\sum_{a_{\text{subopt}} \in \mathcal{A}_{\text{subopt}}} \pi_0(a_{\text{subopt}}) < \sum_{a_{\text{subopt}} \in \mathcal{A}_{\text{subopt}}} \mathbb{E}_{\pi_1^{\text{PPO}}}\left[\pi_1^{\text{PPO}}(a_{\text{subopt}})|\pi_0\right] < \cdots < \sum_{a_{\text{subopt}} \in \mathcal{A}_{\text{subopt}}} \mathbb{E}_{\pi_t^{\text{PPO}}}\left[\pi_t^{\text{PPO}}(a_{\text{subopt}})|\pi_0\right]$;

*(ii)* $\pi_0(a_{\text{opt}}) > \mathbb{E}_{\pi_1^{\text{PPO}}}\left[\pi_1^{\text{PPO}}(a_{\text{opt}})|\pi_0\right] > \cdots > \mathbb{E}_{\pi_t^{\text{PPO}}}\left[\pi_t^{\text{PPO}}(a_{\text{opt}})|\pi_0\right]$;

*(iii)* $\Delta_{\pi_0,0} < \Delta_{\pi_0,1}^{\text{PPO}} < \cdots < \Delta_{\pi_0,t}^{\text{PPO}}$.

**Proof:** For PPO, if $c(a) > 0$, then we have

$$\mathbb{E}_{\pi_{t+1}^{\text{PPO}}}\left[\pi_{t+1}^{\text{PPO}}(a)|\pi_t\right] = \pi_t(a) + \left[\pi_t^2(a) - \sum_{a^+ \in \mathcal{A}^+/\{a\}} \frac{\pi_t^2(a^+)}{|\mathcal{A}|-1} + \sum_{a^- \in \mathcal{A}^-} \frac{\pi_t^2(a^-)}{|\mathcal{A}|-1}\right]\epsilon \quad (15)$$

Let $L(a) = \pi_t^2(a) - \sum_{a^+ \in \mathcal{A}^+/\{a\}} \frac{\pi_t^2(a^+)}{|\mathcal{A}|-1} + \sum_{a^- \in \mathcal{A}^-} \frac{\pi_t^2(a^-)}{|\mathcal{A}|-1}$.

If $\pi_0^2(a_{\text{opt}}) \cdot |\mathcal{A}| < \sum_{a_{\text{subopt}} \in \mathcal{A}_{\text{subopt}}} \pi_0^2(a_{\text{subopt}}) - \sum_{a^- \in \mathcal{A}^-} \pi_0^2(a^-)$, then we have $L(a_{\text{opt}}) < 0$.

Hence we obtain $\pi_0^{\text{PPO}}(a_{\text{opt}}) > \mathbb{E}_{\pi_1^{\text{PPO}}}\left[\pi_1^{\text{PPO}}(a_{\text{opt}})|\pi_t\right]$.

For $a_{\text{subopt}}$, we have

$$\sum_{a \in \mathcal{A}_{\text{subopt}}} L(a_{\text{subopt}})$$
$$= \sum_{a \in a_{\text{subopt}}} \pi_0^2(\text{subopt}) - \pi_0^2(a^+)\frac{|\mathcal{A}_{\text{subopt}}|}{|\mathcal{A}|-1} + \sum_{a^- \in \mathcal{A}^-} \frac{\pi_0^2(a^-)}{|\mathcal{A}|-1} > 0 \quad (16)$$

Thus we have

$$\sum_{a_{\text{subopt}} \in \mathcal{A}_{\text{subopt}}} \pi_0(a_{\text{subopt}}) < \sum_{a_{\text{subopt}} \in \mathcal{A}_{\text{subopt}}} \mathbb{E}_{\pi_1^{\text{PPO}}}\left[\pi_1^{\text{PPO}}(a_{\text{subopt}})|\pi_0\right] \quad (17)$$

Then by Lemma 2, we obtain (i) and (ii). Since (ii) holds, by Lemma 1, we get (iii).

$\square$

### E.2 Theorems in Section 5

**Lemma 3.** *For TRGPPO with hyperparameter $\delta$, we have $\frac{du_{s,a}^\delta}{d\pi_{\text{old}}(a|s)} < 0$, $\frac{dl_{s,a}^\delta}{d\pi_{\text{old}}(a|s)} > 0$.*

**Proof:** By solving (1) and (2) for discrete space, we have

$$(1 - \pi_{\text{old}}(a|s)) \log \frac{1 - \pi_{\text{old}}(a|s)}{1 - \pi_{\text{old}}(a|s)l_{s,a}^\delta} - \pi_{\text{old}}(a|s) \log l_{s,a}^\delta = \delta \tag{18}$$

$$(1 - \pi_{\text{old}}(a|s)) \log \frac{1 - \pi_{\text{old}}(a|s)}{1 - \pi_{\text{old}}(a|s)u_{s,a}^\delta} - \pi_{\text{old}}(a|s) \log u_{s,a}^\delta = \delta \tag{19}$$

To be abbreviated, let $l = l_{s_t,a_t}^\delta$, $u = u_{s_t,a_t}^\delta$, $p = \pi_{\text{old}}(a_t|s_t)$. By eq. (18), we have

$$\begin{aligned}
\frac{dl}{dp} &= \frac{l\left((1-pl)\log\frac{(1-p)l}{1-pl} + 1 - l\right)}{p(l-1)} \\
&= \frac{l(1-pl)}{p(l-1)}\left(\log\frac{(1-p)l}{1-pl} + \frac{1-l}{1-pl}\right) \\
&= \frac{l(1-pl)}{p(l-1)}\left(\log\left(1 + \frac{l-1}{1-pl}\right) - \frac{l-1}{1-pl}\right)
\end{aligned} \tag{20}$$

Note that $0 < l < 1$, $1 - pl > 0$, we have $\frac{l-1}{1-pl} = -\frac{1-l}{1-pl} > -1$. Indeed, $\log(1+x) - x < 0$ for any $x > -1$. Hence, $\log\left(1 + \frac{l-1}{1-pl}\right) - \frac{l-1}{1-pl} < 0$. We obtain $\frac{dl}{dp} > 0$.

Similarly, we can get that $\frac{du}{dp} > 0$.

$\square$

**Theorem 3.** *For TRGPPO with hyperparameter $(\delta, \epsilon)$ and PPO with same $\epsilon$. If $\delta \leq g(\max_{a \in \mathcal{A}_{\text{subopt}}} \pi_t(a), 1+\epsilon)$ for all $t$, then we have $\Delta_{\pi_0,t}^{\text{TRGPPO}} \leq \Delta_{\pi_0,t}^{\text{PPO}}$ for any $t$.*

**Proof:** If $\delta \leq g(\max_{a \in \mathcal{A}_{\text{subopt}}} \pi_t(a), 1+\epsilon)$, then by Lemma 3 and Lemma 5 we have $u_{a_{\text{subopt}}}^\delta \leq 1 + \epsilon$. Hence, $u_{a_{\text{subopt}}}^{\delta,\epsilon} = 1 + \epsilon$.

Meanwhile, $u_a^{\delta,\epsilon} \geq 1 + \epsilon$ and $l_a^{\delta,\epsilon} \leq 1 - \epsilon$ for any $a$.

If $c(a) \geq 0$, then we have

$$\mathbb{E}_{\pi_{\text{new}}}[\pi_{\text{new}}(a)|\pi_t] = \pi_t(a) + \left[\pi_t^2(a)(u_a - 1) - \sum_{a^+ \in \mathcal{A}^+/\{a\}} \frac{\pi_t^2(a^+)}{|\mathcal{A}|-1}(u_{a^+} - 1) + \sum_{a^- \in \mathcal{A}^-} \frac{\pi_t^2(a^-)}{|\mathcal{A}|-1}(1 - l_{a^-})\right] \tag{21}$$

Since $u_{a_{\text{opt}}}^{\delta,\epsilon} \geq 1 + \epsilon$ and $l_{a^-}^{\delta,\epsilon} \leq 1 - \epsilon$ while $u_{a_{\text{subopt}}}^{\delta,\epsilon} = 1 + \epsilon$, we can get $\mathbb{E}_{\pi_{t+1}^{\text{TRGPPO}}}\left[\pi_{t+1}^{\text{TRGPPO}}(a_{\text{opt}})|\pi_t\right] \geq \mathbb{E}_{\pi_{t+1}^{\text{PPO}}}\left[\pi_{t+1}^{\text{PPO}}(a_{\text{opt}})|\pi_t\right]$.

Then by Lemma 2, we have $\mathbb{E}_{\pi_{t+1}^{\text{TRGPPO}}}\left[\pi_{t+1}^{\text{TRGPPO}}(a_{\text{opt}})|\pi_0\right] \geq \mathbb{E}_{\pi_{t+1}^{\text{PPO}}}\left[\pi_{t+1}^{\text{PPO}}(a_{\text{opt}})|\pi_0\right]$.

Finally, by Lemma 1, we have $\Delta_{\pi_0,t}^{\text{TRGPPO}} \leq \Delta_{\pi_0,t}^{\text{PPO}}$.

$\square$

We now derive the form of the optimal solution which achieves minimum KL divergence over all optimal solutions. The general form of surrogate objective function of PPO is as follows:

$$\hat{L}_{\pi_{\text{old}}}^{\text{CLIP}}(\pi) = \frac{1}{T} \sum_{t=1}^{T} [\min(r_\pi(s_t, a_t) A_t, clip(r_\pi(s_t, a_t), l_{s_t,a_t}, u_{s_t,a_t}) A_t))] \quad (22)$$

Let $\Pi_{\text{new}}$ denote the set of all the optimal solutions of the empirical surrogate objective function of PPO, and let $\pi_{\text{new}} \in \Pi_{\text{new}}$ denote the optimal solution which achieves minimum KL divergence over all optimal solutions, i.e., $D_{\text{KL}}^{s_t}(\pi_{\text{old}}, \pi_{\text{new}}) \leq D_{\text{KL}}^{s_t}(\pi_{\text{old}}, \pi)$ for any $\pi \in \Pi_{\text{new}}$ under all $s_t$.

**We first give the form of $\Pi_{\text{new}}$.**

**Lemma 4.** $\Pi_{\text{new}} = \{\pi | \text{for all } t \text{ that } A_t < 0, \pi(a_t|s_t) \leq \pi_{\text{old}}(a_t|s_t) l_{s_t,a_t}; \text{for all } t \text{ that } A_t > 0, \pi(a_t|s_t) \geq \min(\pi_{\text{old}}(a_t|s_t) u_{s_t,a_t}, 1)\}$.

**Proof:**

We first prove that if a policy $\pi^*$ satisfies the conditions in $\Pi_{\text{new}}$, then $\pi^* \in \Pi_{\text{new}}$.

Let $\hat{L}_{\pi_{\text{old}}}^t(\pi) = \min(r_\pi(s_t, a_t) A_t, clip(r_\pi(s_t, a_t), l_{s_t,a_t}, u_{s_t,a_t}) A_t))$. To prove that $\hat{L}_{\pi_{\text{old}}}^{\text{CLIP}}(\pi^*) \geq \hat{L}_{\pi_{\text{old}}}^{\text{CLIP}}(\pi)$ for any $\pi$, we just need to prove that $\hat{L}_{\pi_{\text{old}}}^t(\pi^*) \geq \hat{L}_{\pi_{\text{old}}}^t(\pi)$ for any $\pi$ under all $t$.

If $A_t < 0$, $\hat{L}_{\pi_{\text{old}}}^t(\pi)$ could be rewritten as the following form:

$$\hat{L}_{\pi_{\text{old}}}^t(\pi) = \begin{cases} l_{s_t,a_t} A_t & r_\pi(s_t, a_t) \leq l_{s_t,a_t} \\ r_\pi(s_t, a_t) A_t & r_\pi(s_t, a_t) > l_{s_t,a_t} \end{cases} \quad (23)$$

Thus, we have $\hat{L}_{\pi_{\text{old}}}^t(\pi) \leq l_{s_t,a_t} A_t = \hat{L}_{\pi_{\text{old}}}^t(\pi^*)$ for any $\pi$.

Similarly, if $A_t > 0$, we also have $\hat{L}_{\pi_{\text{old}}}^t(\pi) \leq \hat{L}_{\pi_{\text{old}}}^t(\pi^*)$ for any $\pi$.

We then prove that if a policy $\pi_0$ does not satisfy the conditions in $\Pi_{\text{new}}$, then $\pi^*$ is not an optimal solution in maximization problem of eq. (22).

We can construct a policy $\pi^*$ that satisfy the conditions in the $\Pi_{\text{new}}$. We have $\hat{L}_{\pi_{\text{old}}}^t(\pi_0) < \hat{L}_{\pi_{\text{old}}}^t(\pi^*)$ on $t$ that does not satisfy the conditions. Hence, $\hat{L}_{\pi_{\text{old}}}^{\text{CLIP}}(\pi_0) < \hat{L}_{\pi_{\text{old}}}^{\text{CLIP}}(\pi^*)$.

$\square$

**We now derive the form of $\pi_{\text{new}}$.**

If $A_t < 0$, by Lemma 4, $\min_{\pi \in \Pi_{\text{new}}} D_{\text{KL}}^{s_t}(\pi_{\text{old}}, \pi)$ is formalized as the following problem:

$$\min_\pi \sum_a \pi_{\text{old}}(a|s_t) \log \frac{\pi_{\text{old}}(a_t|s_t)}{\pi(a_t|s_t)}$$
$$s.t. \pi(a_t|s_t) \leq \pi_{\text{old}}(a_t|s_t) l_{s_t,a_t}, \quad (24)$$
$$\sum_a \pi(a|s_t) = 1, \pi(a|s_t) > 0$$

By using the KKT conditions, we can get that

$$\pi_{\text{new}}(a|s_t) = \begin{cases} \frac{\pi_{\text{old}}(a|s_t)(1 - \pi_{\text{old}}(a_t|s_t) l_{s_t,a_t})}{1 - \pi_{\text{old}}(a_t|s_t)} & a \neq a_t \\ \pi_{\text{old}}(a_t|s_t) l_{s_t,a_t} & a = a_t \end{cases} \quad (25)$$

The according KL divergence is

$$D_{\text{KL}}^{s_t}(\pi_{\text{old}}, \pi_{\text{new}}) = (1 - \pi_{\text{old}}(a_t|s_t)) \log \frac{1 - \pi_{\text{old}}(a_t|s_t)}{1 - \pi_{\text{old}}(a_t|s_t) l_{s_t,a_t}} - \pi_{\text{old}}(a_t|s_t) \log l_{s_t,a_t} \quad (26)$$

Similarly, if $A_t > 0$, we can get

$$\pi_{\text{new}}(a|s_t) = \begin{cases} \frac{\pi_{\text{old}}(a|s_t)(1 - \min(\pi_{\text{old}}(a_t|s_t) u_{s_t,a_t}, 1))}{1 - \pi_{\text{old}}(a_t|s_t)} & a \neq a_t \\ \min(\pi_{\text{old}}(a_t|s_t) u_{s_t,a_t}, 1) & a = a_t \end{cases} \quad (27)$$

If $A_t > 0$ and $\pi_{\text{old}}(a_t|s_t)u_{s_t,a_t} \leq 1$, the according KL is

$$D_{\text{KL}}^{s_t}(\pi_{\text{old}}, \pi_{\text{new}}) = (1 - \pi_{\text{old}}(a_t|s_t)) \log \frac{1 - \pi_{\text{old}}(a_t|s_t)}{1 - \pi_{\text{old}}(a_t|s_t)u_{s_t,a_t}} - \pi_{\text{old}}(a_t|s_t) \log u_{s_t,a_t} \quad (28)$$

If $A_t > 0$ and $\pi_{\text{old}}(a_t|s_t)u_{s_t,a_t} > 1$, we have $D_{\text{KL}}^{s_t}(\pi_{\text{old}}, \pi_{\text{new}}) = +\infty$. Equation (26) and eq. (28) have just the same form w.r.t. $l_{s_t,a_t}$ and $u_{s_t,a_t}$ respectively. In fact, since $l_{s_t,a_t} \in (0,1)$ and $u_{s_t,a_t} \in (1, +\infty)$, the monotonicity w.r.t. $l_{s_t,a_t}$ and $u_{s_t,a_t}$ on these two intervals are different, and we obtain the correlation between clipping range and KL divergence.

**Lemma 5.** *(i) If $A_t < 0$, we have $dD_{\text{KL}}^{s_t}(\pi_{\text{old}}, \pi_{\text{new}})/dl_{s_t,a_t} < 0$, $dD_{\text{KL}}^{s_t}(\pi_{\text{old}}, \pi_{\text{new}})/d\pi_{\text{old}}(a_t|s_t) > 0$. (ii) If $A_t > 0$ and $\pi_{\text{old}}(a_t|s_t)u_{s_t,a_t} \leq 1$, we have $dD_{\text{KL}}^{s_t}(\pi_{\text{old}}, \pi_{\text{new}})/du_{s_t,a_t} > 0$, $dD_{\text{KL}}^{s_t}(\pi_{\text{old}}, \pi_{\text{new}})/d\pi_{\text{old}}(a_t|s_t) > 0$.*

**Proof:** To be abbreviated, let $D = D_{\text{KL}}^{s_t}(\pi_{\text{old}}, \pi_{\text{new}})$, $l = l_{s_t,a_t}^\delta$, $u = u_{s_t,a_t}^\delta$, $p = \pi_{\text{old}}(a_t|s_t)$.

If $A_t < 0$, by eq. (26), we have

$$\begin{aligned}\frac{dD}{dp} &= -\log \frac{(1-p)l}{1-pl} + \frac{l-1}{1-pl} \\ &= -\log\left(1 + \frac{l-1}{1-pl}\right) + \frac{l-1}{1-pl}\end{aligned} \quad (29)$$

If $A_t > 0$ and $\pi_{\text{old}}(a_t|s_t)u_{s_t,a_t} \leq 1$, by eq. (28), we have

$$\frac{dD}{dp} = -\log\left(1 + \frac{u-1}{1-pu}\right) + \frac{u-1}{1-pu} \quad (30)$$

We have $\frac{l-1}{1-pl} > -1$ and $\frac{u-1}{1-pu} > 0$. Indeed, $-\log(1+x) + x > 0$ for any $x > -1$. Thus, we have $\frac{dD}{dp} > 0$.

If $A_t < 0$, by eq. (26), we have

$$\frac{dD}{dl} = \frac{p(l-1)}{l(1-pl)} < 0 \quad (31)$$

If $A_t > 0$ and $\pi_{\text{old}}(a_t|s_t)u_{s_t,a_t} \leq 1$, by eq. (28), we have

$$\frac{dD}{du} = \frac{p(u-1)}{u(1-pu)} > 0 \quad (32)$$

$\square$

We introduce an empirical version of lower performance bound.

$$\hat{M}_{\pi_{\text{old}}}(\pi) = \hat{L}_{\pi_{\text{old}}}(\pi) - C \max_t D_{\text{KL}}^{s_t}(\pi_{\text{old}}, \pi). \quad (33)$$

where $\hat{L}_{\pi_{\text{old}}}(\pi) = \frac{1}{T} \sum_{t=1}^T [r_\pi(s_t, a_t)A_t] + \hat{\eta}^{\pi_{\text{old}}}$, $\hat{\eta}^{\pi_{\text{old}}}$ is the estimated performance of $\pi_{\text{old}}$.

**Lemma 6.** *(i) For PPO, assume that $\max_t D_{\text{KL}}^{s_t}(\pi_{\text{old}}, \pi_{\text{new}}^{\text{PPO}}) < +\infty$, if a given $(s_t, a_t)$ satisfies $\pi_{\text{old}}(a_t|s_t) \neq \max_{\hat{t}:A_{\hat{t}}<0} \pi_{\text{old}}(a_{\hat{t}}|s_{\hat{t}})$ and $\pi_{\text{old}}(a_t|s_t) \neq \max_{\hat{t}:A_{\hat{t}}>0} \pi_{\text{old}}(a_{\hat{t}}|s_{\hat{t}})$, then $D_{\text{KL}}^{s_t}(\pi_{\text{old}}, \pi_{\text{new}}^{\text{PPO}}) < \max_{\hat{t}} D_{\text{KL}}^{s_{\hat{t}}}(\pi_{\text{old}}, \pi_{\text{new}}^{\text{PPO}})$. (ii) For TRGPPO, we have $D_{\text{KL}}^{s_t}(\pi_{\text{old}}, \pi_{\text{new}}^{\text{TRGPPO}}) = \max_{\hat{t}} D_{\text{KL}}^{s_{\hat{t}}}(\pi_{\text{old}}, \pi_{\text{new}}^{\text{TRGPPO}})$ for any $(s_t, a_t)$.*

**Proof:** We first prove (i).

If $A_t < 0$, if $\pi_{\text{old}}(a_t|s_t) \neq \max_{\hat{t}:A_{\hat{t}}<0} \pi_{\text{old}}(a_{\hat{t}}|s_{\hat{t}})$, then $\pi_{\text{old}}(a_t|s_t) < \max_{\hat{t}:A_{\hat{t}}<0} \pi_{\text{old}}(a_{\hat{t}}|s_{\hat{t}})$. By Theorem 5, we have $D_{\text{KL}}^{s_t}(\pi_{\text{old}}, \pi_{\text{new}}^{\text{PPO}}) < \max_{\hat{t}:A_{\hat{t}}<0} D_{\text{KL}}^{s_{\hat{t}}}(\pi_{\text{old}}, \pi_{\text{new}}^{\text{PPO}}) \leq \max_{\hat{t}} D_{\text{KL}}^{s_{\hat{t}}}(\pi_{\text{old}}, \pi_{\text{new}}^{\text{PPO}})$.

Similarly, if $A_t > 0$, we also have $D_{\text{KL}}^{s_t}(\pi_{\text{old}}, \pi_{\text{new}}^{\text{PPO}}) < \max_{\hat{t}} D_{\text{KL}}^{s_{\hat{t}}}(\pi_{\text{old}}, \pi_{\text{new}}^{\text{PPO}})$.

We then prove (ii). If $A_t < 0$, by eq. (18) and eq. (26), we have

$$D_{\text{KL}}^{s_t}(\pi_{\text{old}}, \pi_{\text{new}}) = (1 - \pi_{\text{old}}(a_t|s_t)) \log \frac{1 - \pi_{\text{old}}(a_t|s_t)}{1 - \pi_{\text{old}}(a_t|s_t)l_{s_t,a_t}^{\delta}} - \pi_{\text{old}}(a_t|s_t) \log l_{s_t,a_t}^{\delta} = \delta \quad (34)$$

Similarly, if $A_t > 0$, we also have $D_{\text{KL}}^{s_t}(\pi_{\text{old}}, \pi_{\text{new}}) = \delta$.

□

**Theorem 4.** *Assume that* $\max_t D_{\text{KL}}^{s_t}(\pi_{\text{old}}, \pi_{\text{new}}^{\text{PPO}}) < +\infty$ *for all t. If TRGPPO and PPO have the same hyperparameter $\epsilon$, we have:*

*(i) $u_{s_t,a_t}^{\delta} \geq 1 + \epsilon$ and $l_{s_t,a_t}^{\delta} \leq 1 - \epsilon$ for all $(s_t, a_t)$;*

*(ii) $\max_t D_{\text{KL}}^{s_t}(\pi_{\text{old}}, \pi_{\text{new}}^{\text{TRGPPO}}) = \max_t D_{\text{KL}}^{s_t}(\pi_{\text{old}}, \pi_{\text{new}}^{\text{PPO}})$;*

*(iii) $\hat{M}_{\pi_{\text{old}}}(\pi_{\text{new}}^{\text{TRGPPO}}) \geq \hat{M}_{\pi_{\text{old}}}(\pi_{\text{new}}^{\text{PPO}})$. Particularly, if there exists at least one $(s_t, a_t)$ such that $\pi_{\text{old}}(a_t|s_t) \neq \max_{\hat{t}:A_{\hat{t}}<0} \pi_{\text{old}}(a_{\hat{t}}|s_{\hat{t}})$ and $\pi_{\text{old}}(a_t|s_t) \neq \max_{\hat{t}:A_{\hat{t}}>0} \pi_{\text{old}}(a_{\hat{t}}|s_{\hat{t}})$, then $\hat{M}_{\pi_{\text{old}}}(\pi_{\text{new}}^{\text{TRGPPO}}) > \hat{M}_{\pi_{\text{old}}}(\pi_{\text{new}}^{\text{PPO}})$.*

**Proof:**

We first prove (ii). By Equation (13) and Lemma 6, we have $\max_t D_{\text{KL}}^{s_t}(\pi_{\text{old}}, \pi_{\text{new}}^{\text{TRGPPO}}) = \delta = \max_t D_{\text{KL}}^{s_t}(\pi_{\text{old}}, \pi_{\text{new}}^{\text{PPO}})$.

We then prove (i). By (ii) we have $D_{\text{KL}}^{s_t}(\pi_{\text{old}}, \pi_{\text{new}}^{\text{PPO}}) \leq \max_{\hat{t}} D_{\text{KL}}^{s_{\hat{t}}}(\pi_{\text{old}}, \pi_{\text{new}}^{\text{PPO}}) = \delta = D_{\text{KL}}^{s_t}(\pi_{\text{old}}, \pi_{\text{new}}^{\text{TRGPPO}})$ for all $(s_t, a_t)$. Thus, we have $D_{\text{KL}}^{s_t}(\pi_{\text{old}}, \pi_{\text{new}}^{\text{PPO}}) \leq D_{\text{KL}}^{s_t}(\pi_{\text{old}}, \pi_{\text{new}}^{\text{TRGPPO}})$. Indeed, by Lemma 5, we have $dl_{s_t,a_t}/dD_{\text{KL}}^{s_t}(\pi_{\text{old}}, \pi_{\text{new}}) < 0$, $du_{s_t,a_t}/dD_{\text{KL}}^{s_t}(\pi_{\text{old}}, \pi_{\text{new}}) > 0$. Thus, we obtain $l_{s_t,a_t}^{\delta} \leq 1 - \epsilon$ and $u_{s_t,a_t}^{\delta} \geq 1 + \epsilon$.

Particularly, by Lemma 6, if a given $(s_t, a_t)$ satisfies $\pi_{\text{old}}(a_t|s_t) \neq \max_{\hat{t}:A_{\hat{t}}<0} \pi_{\text{old}}(a_{\hat{t}}|s_{\hat{t}})$ and $\pi_{\text{old}}(a_t|s_t) \neq \max_{\hat{t}:A_{\hat{t}}>0} \pi_{\text{old}}(a_{\hat{t}}|s_{\hat{t}})$, then $D_{\text{KL}}^{s_t}(\pi_{\text{old}}, \pi_{\text{new}}^{\text{PPO}}) < D_{\text{KL}}^{s_t}(\pi_{\text{old}}, \pi_{\text{new}}^{\text{TRGPPO}})$. Hence, we have $l_{s_t,a_t}^{\delta} < 1 - \epsilon$ and $u_{s_t,a_t}^{\delta} > 1 + \epsilon$.

We finally prove (iii). By eq. (25) and eq. (27), we can get that

$$r_{\pi_{\text{new}}}(s_t, a_t)A_t = \begin{cases} l_{s_t,a_t} A_t & A_t < 0 \\ \min(u_{s_t,a_t}, 1)A_t & A_t > 0 \end{cases} \quad (35)$$

By (i) we have $r_{\pi_{\text{new}}^{\text{TRGPPO}}}(s_t, a_t)A_t \geq r_{\pi_{\text{new}}^{\text{PPO}}}(s_t, a_t)A_t$ on all $s_t, a_t$. Thus, $\frac{1}{T}\sum_{t=1}^{T}\left[r_{\pi_{\text{new}}^{\text{TRGPPO}}}(s_t, a_t)A_t\right] \geq \frac{1}{T}\sum_{t=1}^{T}\left[r_{\pi_{\text{new}}^{\text{PPO}}}(s_t, a_t)A_t\right]$. By (ii) and the definition of $\hat{M}_{\pi_{\text{old}}}$, we obtain $\hat{M}_{\pi_{\text{old}}}(\pi_{\text{new}}^{\text{TRGPPO}}) \geq \hat{M}_{\pi_{\text{old}}}(\pi_{\text{new}}^{\text{PPO}})$.

Particularly, if there exists one $(s_t, a_t)$ that satisfies $\pi_{\text{old}}(a_t|s_t) \neq \max_{\hat{t}:A_{\hat{t}}<0} \pi_{\text{old}}(a_{\hat{t}}|s_{\hat{t}})$ and $\pi_{\text{old}}(a_t|s_t) \neq \max_{\hat{t}:A_{\hat{t}}>0} \pi_{\text{old}}(a_{\hat{t}}|s_{\hat{t}})$, then we have $r_{\pi_{\text{new}}^{\text{TRGPPO}}}(s_t, a_t)A_t > r_{\pi_{\text{new}}^{\text{PPO}}}(s_t, a_t)A_t$. Hence, we obtain $\hat{M}_{\pi_{\text{old}}}(\pi_{\text{new}}^{\text{TRGPPO}}) > \hat{M}_{\pi_{\text{old}}}(\pi_{\text{new}}^{\text{PPO}})$.

□

                                              
\end{document}